\documentclass{article}

    \PassOptionsToPackage{numbers, compress}{natbib}

 \usepackage[main, final]{neurips_2026}

\usepackage[utf8]{inputenc} 
\usepackage[T1]{fontenc}    
\usepackage{hyperref}       
\usepackage{url}            
\usepackage{booktabs}       
\usepackage{amsfonts}       
\usepackage{nicefrac}       
\usepackage{microtype}      
\usepackage{xcolor}         

\usepackage{amsmath}

\usepackage{graphicx}
\usepackage{subcaption}
\usepackage{adjustbox}
\usepackage{multirow}
\usepackage{diagbox}
\usepackage{wrapfig}
\usepackage{enumitem}
\usepackage[table]{xcolor}
\usepackage{xspace}

\usepackage[toc]{appendix}
\usepackage{etoc}
\usepackage{minitoc}
\etocsettocstyle{\section*{Appendix}}{}

\definecolor{cadmiumgreen}{rgb}{0.0, 0.42, 0.24}
\definecolor{cornellred}{rgb}{0.7, 0.11, 0.11}
\definecolor{citecolor}{HTML}{0071bc}

\hypersetup{
  urlcolor = cadmiumgreen,
    citecolor = cadmiumgreen,
  colorlinks = true,
}

\newcommand{\colorup}[1]{~{\color{cadmiumgreen}(+#1)}}
\newcommand{\up}[1]{~{\color{gray! 30}(+#1)}}
\newcommand{\down}[1]{~{\color{gray! 30}(\,-#1)}}

\newcommand{\blankup}{\phantom{\up{x.x}}}
\newcommand{\stdv}[1]{{\tiny$\pm$#1}}
\newcommand{\blankstdv}{\phantom{\stdv{x.x}}}

\newcommand{\lname}{Topology-Aware Multimodal Representation Alignment\xspace}
\newcommand{\sname}{ToMA\xspace}
\newcommand{\snameall}{ToMA\xspace}
\newcommand{\snamedom}{ToMA-domain\xspace}

\renewcommand{\eqref}[1]{Eq.~\ref{#1}}

\title{%
Topology-Aware Representation Alignment
for Semi-Supervised Vision-Language Learning
}

\author{%
 Junwon You$^1$ \quad
  Mihyun Jang$^2$ \quad
  Sangwoo Mo$^{2,\dagger}$ \quad
  Jae-Hun Jung$^{2,\dagger}$ \\
  $^1$KAIST\quad$^2$POSTECH
}

\begin{document}

\maketitle

\etocdepthtag.toc{mtchapter}
\etocsettagdepth{mtchapter}{subsection}
\etocsettagdepth{mtappendix}{none}
\faketableofcontents

\renewcommand*{\thefootnote}{$\dagger$}
\footnotetext{Equal advising. Code: \url{https://github.com/junwon0/ToMA}}

\setcounter{footnote}{0}
\renewcommand{\thefootnote}{\arabic{footnote}}

\begin{abstract}
Vision-language models have shown strong performance, but they often generalize poorly to specialized domains. 
While semi-supervised vision-language learning mitigates this limitation by leveraging a small set of labeled image-text pairs together with abundant unlabeled images, existing methods remain fundamentally pairwise and fail to model the global structure of multimodal representation manifolds.
Most existing topology-based alignment methods rely on persistence diagram matching, which neither guarantees geometric alignment nor uses the image-text pairing information central to vision-language learning.
We propose \lname (\sname), a framework that uses persistent homology to
decompose pairwise relations into $H_0$-death and $H_1$-birth relations and
aligns them across modalities through available cross-modal correspondences.
This lightweight decomposition captures both connectivity and cycle-closing
structure without constructing 2-simplices.
Experiments show clear improvements on remote sensing, while fashion results reveal strong gains in classification and particularly strong retrieval improvements for difficult queries.
Additional analysis shows that \sname provides more consistent gains than alternative topology-based objectives and that lightweight $H_1$-birth edges provide useful higher-order structural signals.

\end{abstract}
\section{Introduction}

Vision-language models (VLMs) such as CLIP~\citep{radford2021learning} have shown strong performance in learning joint image-text representations. However, they often generalize poorly to specialized domains, including remote sensing~\citep{arutiunian2021fine}, medical imaging~\citep{kim2023bias}, and fashion, where the visual distribution deviates from that of the pretraining data. Adapting VLMs to these domains typically requires domain-specific image-text pairs, which are expensive to collect, while unlabeled images are readily available. This motivates semi-supervised vision-language learning, where adaptation is performed using a small set of labeled image-text pairs and abundant unlabeled images~\citep{mo2023s, gan2025semi}.

\begin{figure}[!t]
\centering\small
\includegraphics[width=0.98\linewidth]{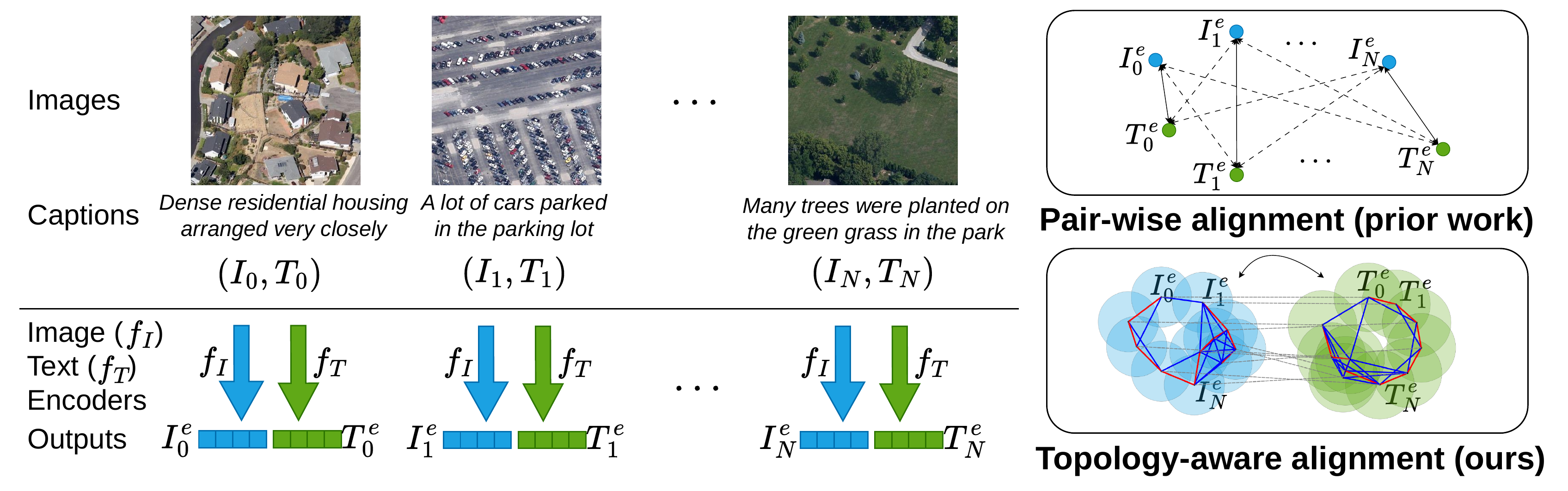} 
\vspace{-6pt}
\caption{
\textbf{Comparison between pair-wise and topology-aware alignment.}
Given image-text pairs and their encoder outputs, prior approaches align corresponding embeddings independently.
In contrast, our method aligns the global structural organization of image and text representation spaces by preserving topologically meaningful relationships among embeddings.
This topology-aware perspective promotes cross-modal consistency at the level of the representation manifold.
}
\label{fig:motiv}
\vspace{-17pt}
\end{figure}

Several recent works have extended CLIP to semi-supervised learning. S-CLIP~\citep{mo2023s} introduces a semi-supervised contrastive objective with caption-level and keyword-level pseudo-labeling. SemiCLIP~\citep{gan2025semi} builds on this idea by mining semantic concepts from labeled data and generating pseudo-labels for unlabeled samples. It further enforces cross-modal geometric consistency via trapezoidal consistency regularization over image-text embedding pairs. Despite these advances, existing objectives remain fundamentally pairwise. They align matched image-text pairs and impose geometric constraints only between pairs of embeddings (Figure~\ref{fig:motiv}, Pair-wise alignment).

Deep neural networks often learn representation spaces with manifold structure~\citep{chen2017deep, cohen2020separability}. These spaces encode semantic relationships through geometric organization, including clusters, connectivity, and higher-order structure. 
Preserving these geometric organizations is important for maintaining semantic consistency in learned representations~\citep{bronstein2017geometric, moor2020topological, kim2024topological, yanfan2024persistence, zhang2024homology, you2026topological}. 
In multimodal settings, however, semantically related samples are often separated by the modality gap, which induces modality-specific geometric biases~\citep{liang2022mind, gideoni2025misalignment}.
The Platonic Representation Hypothesis suggests that, despite these differences, representations from distinct modalities may converge toward a shared abstract organization of the underlying manifold~\citep{huh2024position, maniparambil2024vision, groger2025with, roschmann2026sotalign}. 
Under this view, cross-modal alignment should go beyond instance-level matching and instead target the shared global structure of the underlying manifold.

Topological data analysis (TDA)~\citep{carlsson2009topology} provides a framework for characterizing the global structure of data manifolds. Rather than focusing only on pairwise distances, TDA captures structural properties that are preserved under continuous geometric transformations. In particular, it summarizes topological features of representation spaces, such as connected components and loop structures.

Most existing studies align representation spaces by matching their topological structures via persistence diagrams (PDs), which summarize the birth and death  time of topological features~\citep{moor2020topological, kim2024topological, yanfan2024persistence, you2026topological}.
However, PD matching alone has two limitations in multimodal learning.
First, PDs are not injective summaries of geometry: distinct representation spaces can yield identical or highly similar PDs. 
Thus, even if PD distances are minimized, this does not provide an inverse guarantee that the underlying representation geometries are aligned.
Second, PD-based alignment does not incorporate the image-text pairing information that is central to VLMs.
Moreover, most previous studies~\citep{moor2020topological, kim2024topological, yanfan2024persistence, zhang2024homology, you2026topological} focus on 0-dimensional persistent homology (PH), $H_0$.
In particular, HC~\citep{zhang2024homology} replaces PD matching with persistence track coincidence defined over birth and death features, but their approach remains limited to $H_0$ because higher-dimensional PH significantly increases computational cost but brings almost no additional performance benefits.

\textbf{Contribution.}
We propose \lname (\sname) for semi-supervised vision-language learning. 
\sname uses topology to identify topology-induced relations and aligns them using the available image-text pairing information (Figure~\ref{fig:motiv}, Topology-aware alignment). 
This design addresses the two main limitations of PD-based alignment: it avoids using PD as the alignment target, and it explicitly incorporates cross-modal pairing information.

To identify such edges, we use PH to capture topological structure. In particular, $H_0$ characterizes the merging of connected components and thus reflects semantic cluster structure. Beyond $H_0$, \sname also leverages $1$-dimensional PH ($H_1$) to capture higher-order relations. To keep  computation lightweight, we use a filtration defined only over vertices and edges, without constructing $2$-simplices (Figure~\ref{fig:method}\textcolor{red}{a}). Under this construction, the birth times of $H_1$ classes can be obtained directly from the filtration defined by \eqref{eq:filtration}, allowing us to exploit informative loop-level signals. Empirically, even this birth-only $H_1$ information provides consistent performance gains (Figure~\ref{fig:tda_ablation-2}).

Notably, the death simplices of $H_0$ recover the minimum spanning tree (MST) of the embeddings~\citep{you2026topological}, which induces a unique path between any pair of points. 
By emphasizing these topologically significant edges and aligning their directions rather than their Euclidean distances, \sname preserves meaningful structural relations while remaining well matched to CLIP, whose geometry is governed by cosine similarity. 
Together, topology-induced edge decomposition and directional alignment promote global consistency while respecting modality-specific geometry.

Across remote sensing~\citep{lu2017exploring, yang2010bag, zhang2014saliency} and fashion datasets~\citep{han2017automatic, rostamzadeh2018fashion, vasileva2018learning}, \sname generally improves over the baseline.
The gains are most pronounced on remote sensing, while fashion experiments reveal a complementary pattern: ToMA generally improves zero-shot classification and preferentially benefits difficult retrieval queries, accompanied by stronger cross-modal structural agreement and more semantically coherent retrieved neighborhoods.
Additional analysis further shows that \sname is more consistent than alternative topology-based objectives, and that lightweight $H_1$-birth edges provide effective higher-order structural signals beyond conventional $H_0$ alignment.

The contributions of this work are as follows:
\begin{itemize}[leftmargin=3.5mm]
    
    \item We propose \sname, a topology-aware alignment framework for semi-supervised vision-language learning that aligns image and text representations at the level of global structure.

    \item We design a topology-aware objective $\mathcal{L}_{\text{\sname}}$ tailored to semi-supervised vision-language learning, which leverages available image-text correspondences and respects the cosine-based geometry of CLIP representations.

    \item We show that topology can be effectively leveraged by decomposing pairwise relations into $H_0$-death and lightweight $H_1$-birth relations and aligning their directions, going beyond PD matching.

    \item Experiments on remote sensing and fashion datasets show that \sname provides strong overall gains in zero-shot classification and provides task-dependent retrieval benefits.
\end{itemize}

\section{Related Work}

\noindent\textbf{Vision-language pre-training (VLP)}
has become a core paradigm for diverse downstream vision-language tasks~\citep{radford2021learning, jia2021scaling, pham2023combined, huang2023nlip}, with many approaches proposed to learn aligned visual and textual representations~\citep{li2019visualbert, chen2020uniter, kim2021vilt, wang2022simvlm, wang2023image, zhai2022lit, li2023scaling}. 
Among them, CLIP~\citep{radford2021learning} is a representative contrastive framework that learns a joint image-text embedding space and performs strongly on zero-shot classification, image-text retrieval, object detection, and few-shot learning~\citep{li2023rs, luo2022clip4clip, gu2022openvocabulary, xu2022simple, liu2022few}.

\noindent\textbf{Cross-modal alignment with semi-supervised learning.}
In semi-supervised vision-language learning, cross-modal alignment has become an effective strategy for exploiting unlabeled data~\citep{wang2022medclip, wang2023learning, wang2022simvlm}. 
Representative methods such as S-CLIP~\citep{mo2023s} and SemiCLIP~\citep{gan2025semi} improve downstream performance by aligning image and text representations. 
Relatedly, recent studies have explored unimodal alignment based on the Platonic representation hypothesis~\citep{huh2024position}, suggesting a broader alignment perspective beyond direct cross-modal matching~\citep{maniparambil2024vision, groger2025with, roschmann2026sotalign}. 
Our method is instantiated in the former setting, while extending it to the latter direction is left for future work.

\noindent\textbf{Persistent homology (PH) for preserving representation space.}
PH has been increasingly used to characterize and preserve the geometric structure~\citep{madhu2023toposrl, chen2024topogcl, kizaric2024principle, zhang2024deep, papamarkou2024position, eitan2025topological, tung2025towards, YOU2026132147, carriere2026persistence}. 
Early work applies PH-based regularization: Topological Autoencoders~\citep{moor2020topological} preserve the PH of the input during dimensionality reduction, while \cite{trofimov2023learning} enforce topology preservation through Representation Topology Divergence~\citep{barannikov2021representation}. 
PH has also been adopted for knowledge transfer, including knowledge distillation~\citep{kim2024topological} and semi-supervised continual learning~\citep{yanfan2024persistence}. 
More recently, topology-aware constraints have been introduced in multimodal learning, where HC~\citep{zhang2024homology} and ToMCLIP~\citep{you2026topological} regularize representations by encouraging topological consistency across modalities. 
Despite these advances, most prior work focuses on matching topological summaries~\citep{moor2020topological, kim2024topological, yanfan2024persistence, you2026topological, trofimov2023learning}. 
Our method leverages the structural information carried by birth and death simplices to align topologically meaningful edges.

\section{%
\sname: \lname
}


\subsection{Preliminaries}


\paragraph{Contrastive language-image pre-training (CLIP)}
learns a shared embedding space for images and texts by contrasting matched and mismatched image-text pairs~\citep{radford2021learning}. 
Given a batch $\mathcal{D}^l = \{(I_i, T_i)\}_{i=1}^N$, image and text encoders, $f_I(\cdot)$ and $f_T(\cdot)$, produce normalized embeddings $I_i^e$ and $T_i^e$. The model is trained with a symmetric contrastive loss that increases the cosine similarity of matched pairs and decreases that of unmatched pairs:
\[
\mathcal{L}_{\text{CLIP}} = -\frac{1}{2N} \sum_{i=1}^N \left( \log \frac{\exp(\langle I^e_i, T^e_i \rangle / \tau)}{\sum_{t=1}^N \exp(\langle I^e_i, T^e_t \rangle / \tau)} + \log \frac{\exp(\langle I^e_i, T^e_i \rangle / \tau)}{\sum_{t=1}^N \exp(\langle I^e_t, T^e_i \rangle / \tau)} \right)
\]
Here, $\langle \cdot, \cdot \rangle$ denotes cosine similarity and $\tau$ is a temperature. While the original CLIP learns $\tau$ during pre-training, we fix $\tau=1$ in our experiments following~\citep{gan2025semi}.

\paragraph{Semi-supervised vision-language alignment}
addresses the setting where paired image-text data $\mathcal{D}^l = \{(I_i, T_i)\}_{i=1}^N$ are scarce, while unlabeled images $\mathcal{D}^u = \{I_j\}_{j=1}^M$ are abundant with $M \gg N$. Its goal is to leverage unlabeled images to improve the alignment of the shared embedding space. SemiCLIP~\citep{gan2025semi} approaches this problem with a two-stage framework.
In Stage 1, it mines semantic concepts from labeled captions and trains a concept classifier together with the CLIP objective. In Stage 2, it assigns pseudo-labels to unlabeled images and fine-tunes the model using concept-level semantic consistency and caption-level trapezoidal consistency. We denote the corresponding objectives as $\mathcal{L}^{(1)}_{\mathrm{Semi}}$ and $\mathcal{L}^{(2)}_{\mathrm{Semi}}$, respectively, and build ToMA on top of these two objectives.
A detailed explanation is provided in Appendix~\ref{app:preliminary_semisup}.

\paragraph{Persistent homology (PH)} 
studies the global structure of data beyond local pairwise relations. Given a finite set $X = \{x_i\}_{i=1}^N$ in a metric space $(\mathcal{X}, d)$, PH characterizes topology by constructing a nested family of simplicial complexes over increasing scales $\alpha$. It records the appearance and disappearance of homological features, such as connected components and loops~\citep{dey2022computational}.

A simplicial complex $K$ over a vertex set $X$ is a collection of subsets of $X$ closed under inclusion, i.e., if $\sigma \in K$ and $\tau \subseteq \sigma$, then $\tau \in K$. Each $\sigma \in K$ is a simplex with dimension $\dim(\sigma)=|\sigma|-1$. We define a function $f : K \rightarrow \mathbb{R}_{\geq 0} \cup \{\infty\}$ as
\begin{equation}
\label{eq:filtration}
    f(\sigma) =
    \begin{cases}
    0, & |\sigma| = 1, \\
    d(x,y), & \sigma = \{x,y\}, \\
    \infty, & |\sigma| > 2
    \end{cases}    
\end{equation}
which induces the filtration
$
\{ K_{\alpha_i} \hookrightarrow K_{\alpha_j} \}_{\alpha_i \leq \alpha_j}
$,
with
$
K_\alpha = \{ \sigma \in K \mid f(\sigma) \le \alpha \}.
$
As $\alpha$ grows, edges are added to form a nested sequence of complexes. 
This is a Vietoris-Rips filtration truncated at the 1-skeleton, retaining only vertices and edges. 
Under this construction, PH can be computed using a minimum spanning tree, while capturing the birth events of $H_1$~\citep{you2026topological}.
Moreover, these $H_1$ birth events can be incorporated at no extra cost relative to $H_0$ alone. 
Despite this, most prior work~\citep{moor2020topological, kim2024topological, yanfan2024persistence, zhang2024homology, you2026topological} considers only $H_0$.
In contrast, our method benefits from $H_1$ information (Figure~\ref{fig:tda_ablation-2}).

Fix a homology dimension $p \in \{0,1\}$. 
The filtration induces homomorphisms
$
H_p(K_{\alpha_1}) \to H_p(K_{\alpha_2})
$
for $\alpha_1 \leq \alpha_2$, enabling the tracking of $p$-dimensional topological features. Each PH class $\mu$ is characterized by a birth scale $b_\mu$ and a death scale $d_\mu$, together with its corresponding birth $p$-simplex $\sigma^{b}_{\mu}$ and death $(p+1)$-simplex $\sigma^{d}_{\mu}$. In particular, for $p=0$, births occur at vertices and deaths at edges, whereas for $p=1$, births occur at edges and deaths at 2-simplices. 
Because our filtration is restricted to the 1-skeleton, no 2-simplices enter at finite filtration values. Consequently, the resulting $H_1$ classes have no finite death times, and ToMA uses their birth edges rather than finite-persistence intervals of a full Vietoris-Rips filtration. Thus, our construction yields a homologically defined decomposition of pairwise relations into $H_0$-death edges and $H_1$-birth edges.

\subsection{Our \sname Framework}
\label{subsec:tomar}


\begin{figure}[!t]
  \centering\small
  \includegraphics[width=\linewidth]{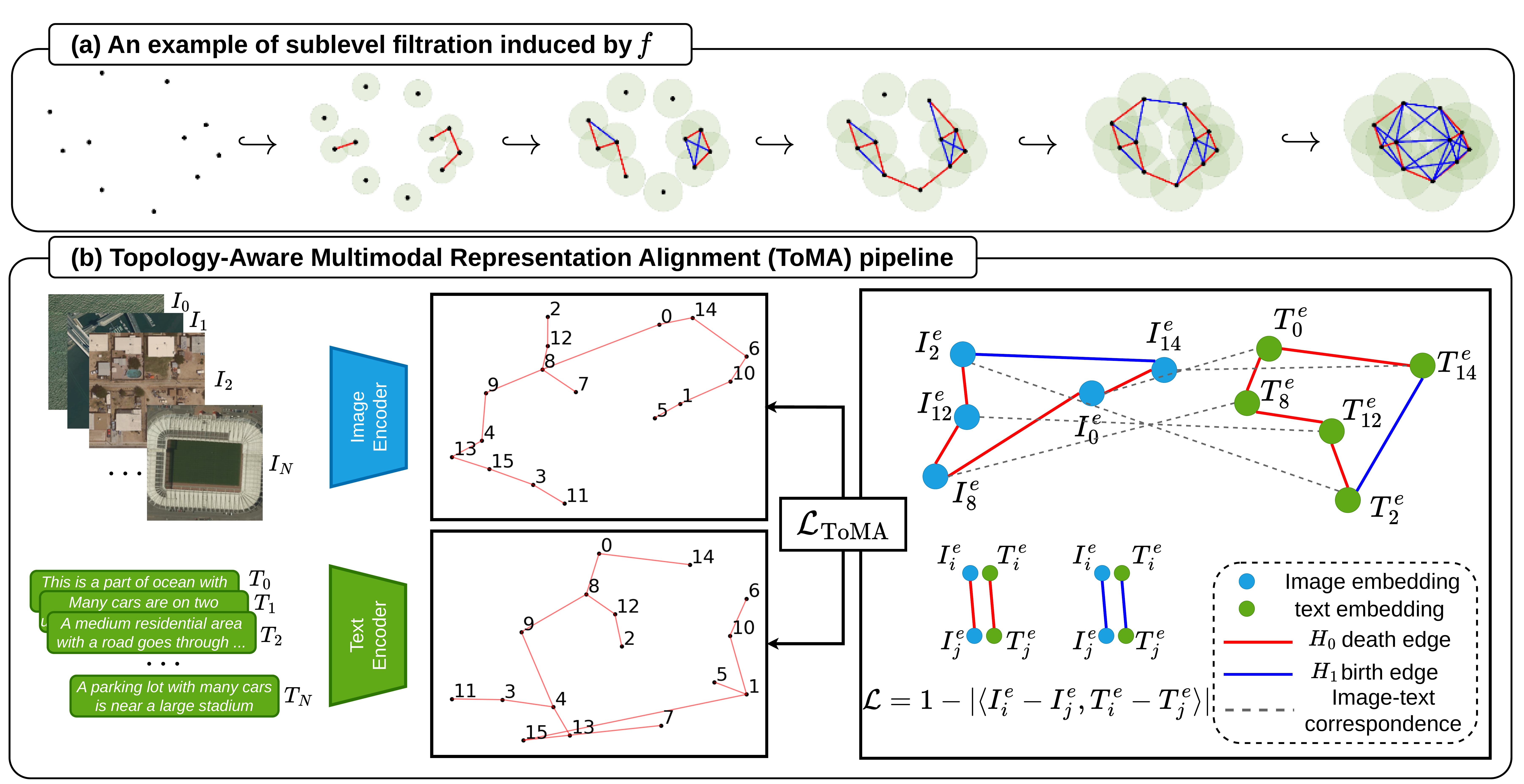}
  \caption{\textbf{\lname (\sname).} 
    \textbf{(a)} Topological decomposition of edges under the filtration induced by \(f\). 
    The death edges of \(H_0\) form the minimum spanning tree (MST; red), capturing connectivity structure, while the birth edges of \(H_1\) are cycle-closing non-MST edges (blue), capturing higher-order structure. 
    \textbf{(b)} \sname aligns the directions of topology-induced $H_0$-death and $H_1$-birth relations across image and text representations using cross-modal correspondences.    
    For visual clarity, the two intermediate embedding plots in panel (b) show only the \(H_0\)-death edges.}
    \label{fig:method}
    \vspace{-6pt}
\end{figure}

Our \sname framework integrates two stages of SemiCLIP~\citep{gan2025semi} and aligns image and text representation spaces through a topology-induced
decomposition of pairwise relations rather than through persistence diagram (PD)-level matching.
As illustrated in Figure~\ref{fig:method}, \sname consists of two steps. 
First, under the 1-skeleton filtration by \eqref{eq:filtration}, the edge set is decomposed into two topologically distinct types, shown in Figure~\ref{fig:method}\textcolor{red}{a}. 
Second, these edges are aligned across image and text using cross-modal correspondences, as summarized in Figure~\ref{fig:method}\textcolor{red}{b}. 
Because the filtration is truncated at the 1-skeleton, \sname can exploit not only the death edges of $H_0$ but also the birth edges of $H_1$ without constructing 2-simplices.

As shown in Figure~\ref{fig:method}\textcolor{red}{a}, the death simplices of $H_0$ form the edges of the MST, whereas the birth simplices of $H_1$ 
consist of all non-MST edges, each of which closes a cycle.
This decomposition gives two complementary classes of topologically salient edges: $H_0$-death edges, which capture connectivity structure through the MST, and $H_1$-birth edges, which capture cycle-level relational structure through cycle-closing edges. 
In the context of vision-language representations, the former primarily reflects local neighborhood relations, while the latter captures higher-order organization among multiple nearby samples. 
This structural distinction, summarized schematically in Figure~\ref{fig:method}, is further illustrated with a concrete example based on real image-text embeddings in Appendix~\ref{app:tomar_example}.

We now formalize the alignment process illustrated in Figure~\ref{fig:method}\textcolor{red}{b}.
Let \(X \subset \mathbb{R}^d\) be a point cloud, and let \(\Gamma_p^X\) denote the set of \(p\)-dimensional PH classes of \(X\).
For each \(\mu \in \Gamma_p^X\), let \(\sigma_{\mu}^{b}\) and \(\sigma_{\mu}^{d}\) denote its birth and death simplices, respectively.
Let \(M_i\) and \(M_j\) denote the point clouds of two modalities.
In our setting, \(M_i\) and \(M_j\) correspond to the image and text embeddings, respectively.
We define a cross-modal pairing map \(\pi_{i \to j} : M_i \to M_j\) according to the supervision available at each training stage:
in stage~1, \(\pi_{i \to j}\) is induced by ground-truth image-text pairs, whereas in stage~2 it is induced by ground-truth pairs for labeled samples and surrogate text embeddings for unlabeled samples.

For \(\mu \in \Gamma_0^{M_i}\), let
$
\sigma_{\mu}^{d} = \{x_0^{(\mu_d)}, x_1^{(\mu_d)}\}
$
be its death simplex in modality \(M_i\).
We denote the corresponding points in modality \(M_j\) by
$
\{\pi_{i \to j}(x_0^{(\mu_d)}), \pi_{i \to j}(x_1^{(\mu_d)})\}.
$
We define the directional consistency of this \(H_0\)-death edge as
\begin{equation}
T_{\mu}^{\text{0-death}}(M_i \to M_j)
=
\left\langle
x_0^{(\mu_d)} - x_1^{(\mu_d)},\;
\pi_{i \to j}(x_0^{(\mu_d)}) - \pi_{i \to j}(x_1^{(\mu_d)})
\right\rangle ,
\label{eq:tomar_h0}
\end{equation}
where \(\langle u,v \rangle := u^\top v / (\|u\|_2 \|v\|_2)\) denotes the cosine similarity between the two edge-direction vectors.
Similarly, for a \(1\)-dimensional PH class \(\mu \in \Gamma_1^{M_i}\), let
$
\sigma_{\mu}^{b} = \{x_0^{(\mu_b)}, x_1^{(\mu_b)}\}
$
be its birth simplex.
The directional consistency of an \(H_1\)-birth edge is defined analogously as
\begin{equation}
T_{\mu}^{\text{1-birth}}(M_i \to M_j)
=
\left\langle
x_0^{(\mu_b)} - x_1^{(\mu_b)},\pi_{i \to j}(x_0^{(\mu_b)}) - \pi_{i \to j}(x_1^{(\mu_b)})
\right\rangle .
\label{eq:tomar_h1}
\end{equation}
We align \emph{edge directions} rather than raw Euclidean distances because CLIP operates on normalized features and measures similarity through cosine similarity.
A direction-based topological alignment therefore better matches the angular geometry.
Moreover, the PH-selected edge sets in the two modalities need not coincide exactly.
For this reason, we align both directions, \(M_i \to M_j\) and \(M_j \to M_i\).
For \( * \in \{\text{0-death}, \text{1-birth}\}\), we define the bidirectional \sname loss as
\begin{equation}
\mathcal{L}_{\text{\sname}}^{*}(M_i, M_j)
=
\frac{1}{\left| \Gamma_{*}^{M_i} \right|}
\sum_{\mu \in \Gamma_{*}^{M_i}}
\left( 1 - \left| T_{\mu}^{*}(M_i \to M_j) \right| \right)
+
\frac{1}{\left| \Gamma_{*}^{M_j} \right|}
\sum_{\mu \in \Gamma_{*}^{M_j}}
\left( 1 - \left| T_{\mu}^{*}(M_j \to M_i) \right| \right),
\label{eq:tomar_loss}
\end{equation}
where, by a slight abuse of notation, \(\Gamma_{*}^{M_i}\) denotes \(\Gamma_{0}^{M_i}\) for \( * = \text{0-death}\) and \(\Gamma_{1}^{M_i}\) for \( * = \text{1-birth}\).
The absolute value makes the objective invariant to whether the mapped edge direction is parallel or anti-parallel to the source edge, thereby aligning the underlying one-dimensional relational subspace rather than enforcing a fixed global orientation (Appendix~\ref{app:direction}).
For each stage \(s \in \{1,2\}\), \sname is applied to the corresponding SemiCLIP objective:
$
\mathcal{L}^{(s)}
=
\mathcal{L}_{\text{Semi}}^{(s)}
+
\frac{c}{2}
\left(
\mathcal{L}_{\text{\sname}}^{\text{0-death}}
+
\mathcal{L}_{\text{\sname}}^{\text{1-birth}}
\right),
$
where \(c\) controls the strength of \sname.


\section{Experiments}


\subsection{Experiment Setup}
\label{sec:setup}


\noindent\textbf{Datasets.}
Following S-CLIP~\citep{mo2023s} and SemiCLIP~\citep{gan2025semi}, we evaluate our method on specialized-domain datasets, including remote sensing datasets (Section~\ref{sec:rs}) and fashion datasets (Section~\ref{sec:fashion}). 
We additionally report retrieval results on science-figure and comics datasets in Appendix~\ref{app:more_datasets} to assess \sname on further specialist domains.
Dataset details are provided in Appendix~\ref{sec:data_details}.
In the semi-supervised setting, we randomly sample 10\% of the training image-text pairs as labeled data and treat the remaining 90\% as unlabeled data.

\noindent\textbf{Implementation details.}
We adopt CLIP~\citep{radford2021learning} as the pretrained backbone for experiments, using ViT-B/16 by default, with additional vision encoder results in Appendix~\ref{app:model_variants}. 
The coefficient $c$ in $\mathcal{L}^{(s)}$ is set to 0.5 for remote sensing and 0.1 for fashion based on the hyperparameter search in Appendix~\ref{app:loss_co}. 
We consider two variants: \snameall, which applies $\mathcal{L}_{\text{\sname}}$ to each batch without domain information, and \snamedom, which applies it separately to each domain. 
To isolate the effect of domain partitioning from that of topology-aware alignment, we additionally evaluate a domain-partitioned SemiCLIP control without the \sname objective in Appendix~\ref{app:domain_partition_control}. Further details are in Appendix~\ref{app:implementation_details}.

\noindent\textbf{Baselines.}
We compare \sname with CLIP (original/fine-tuned/oracle), Hard-PL~\citep{lee2013pseudo}, Soft-PL~\citep{assran2021semi}, S-CLIP~\citep{mo2023s}, and SemiCLIP~\citep{gan2025semi}. 
The scores of CLIP, Hard-PL, Soft-PL, and S-CLIP are imported from~\citep{gan2025semi}, while we retrain SemiCLIP with the same random seeds as our method for fair comparison and report originally published scores in Appendix~\ref{app:semiclip}. 
We additionally conduct controlled objective-level comparisons corresponding to ToMCLIP~\citep{you2026topological} and HC~\citep{zhang2024homology} in Figure~\ref{fig:tda_ablation}.
We evaluate zero-shot classification and image-text retrieval using Top-1 accuracy (\%) and Recall@K (R@K), respectively, and report the mean and standard deviation over three runs with fixed random seeds $\{0,1,2\}$; further implementation details are in Appendix~\ref{app:implementation_details}.

\subsection{Remote Sensing Datasets}
\label{sec:rs}

For remote sensing, we follow the fine-tuning setup of~\citep{arutiunian2021fine} and use RSICD~\citep{lu2017exploring}, UCM~\citep{yang2010bag}, and Sydney~\citep{zhang2014saliency}, collectively denoted as RS-ALL. 
We randomly use 10\% of RS-ALL as labeled data L and the remaining 90\% as unlabeled data U. 
When the unlabeled data are drawn from RS-ALL, we denote the setting as L$=$U; when they are drawn from RESISC45~\citep{cheng2017remote}, we denote it as L$\neq$U.

\begin{table*}[t]
\centering\small
\caption{%
Zero-shot classification results on remote sensing datasets. We compare original CLIP, supervised CLIP fine-tuned on labeled data (L), and semi-supervised methods using unlabeled data from the same (L=U) or different (L$\neq$U) distribution. Parentheses denote performance gains over supervised CLIP, and {\color{cadmiumgreen}green} values indicate gains larger than one. Bold denotes the best semi-supervised result in each setting.
}\label{tab:rs-zeroshot}
\vspace{-0.08in}
\begin{adjustbox}{width=\linewidth}
\begin{tabular}{@{}lccccccc@{}}
\toprule
Method & Data & RSICD-CLS & UCM-CLS & WHU-RS19 & RSSCN7 & AID\\
\midrule
CLIP (original) & - &
59.2\blankstdv\blankup & 60.2\blankstdv\blankup & 81.2\blankstdv\blankup & 69.0\blankstdv\blankup & 59.6\blankstdv\blankup\\
\midrule
CLIP (fine-tuned) & L &
75.7\stdv{2.2}\blankup & 80.1\stdv{6.8}\blankup & 92.2\stdv{1.5}\blankup & 70.7\stdv{5.5}\blankup & 79.8\stdv{3.8}\blankup\\
CLIP (oracle) & L + U &
87.5\stdv{1.8}\blankup & 67.9\stdv{2.3}\blankup & 94.2\stdv{1.5}\blankup & 75.7\stdv{2.9}\blankup & 89.3\stdv{1.2}\blankup\\
\midrule
Hard-PL~\citep{lee2013pseudo} & \multirow{6}{*}{L$=$U} &
78.0\stdv{2.0}\colorup{2.3} & 72.4\stdv{5.2}\down{7.7} & 92.3\stdv{2.6}\up{0.1} & 71.7\stdv{4.1}\colorup{1.0} & 81.4\stdv{2.2}\colorup{1.6}\\
Soft-PL~\citep{assran2021semi} & &
81.7\stdv{0.6}\colorup{6.0} & 78.9\stdv{5.2}\down{1.2} & 95.3\stdv{0.8}\colorup{3.1} & 72.0\stdv{2.5}\colorup{1.3} & 85.8\stdv{1.6}\colorup{6.0}\\
S-CLIP~\citep{mo2023s} & &
81.4\stdv{1.8}\colorup{5.7} & 81.3\stdv{3.4}\colorup{1.2} & 95.9\stdv{1.8}\colorup{3.7} & 75.1\stdv{2.0}\colorup{4.4} & 86.4\stdv{1.7}\colorup{6.6}\\
SemiCLIP~\citep{gan2025semi} & &
81.4\stdv{1.1}\colorup{5.7} & 84.3\stdv{3.2}\colorup{4.2} & 95.6\stdv{0.3}\colorup{3.4} & \textbf{76.2}\stdv{2.7}\colorup{5.5} & 84.0\stdv{1.2}\colorup{4.2}\\
\rowcolor{gray!15}
\snameall (ours) & &  
\textbf{84.8}\stdv{1.3}\colorup{9.1} & 86.1\stdv{3.4}\colorup{6.0} & \textbf{97.9}\stdv{0.4}\colorup{5.7} & 74.2\stdv{1.7}\colorup{3.5} & \textbf{87.3}\stdv{2.2}\colorup{7.5}\\
\rowcolor{gray!15}
\snamedom (ours) & &  
83.8\stdv{3.2}\colorup{8.1} & \textbf{87.5}\stdv{4.4}\colorup{7.4} & 97.8\stdv{0.6}\colorup{5.6} & 73.9\stdv{1.0}\colorup{3.2} & 86.8\stdv{2.8}\colorup{7.0}\\
\midrule
Hard-PL~\citep{lee2013pseudo} & \multirow{6}{*}{L$\neq$U} &
79.9\stdv{2.5}\colorup{4.2} & 76.2\stdv{1.9}\down{3.9} & 91.0\stdv{2.3}\down{1.2} & 71.4\stdv{2.2}\up{0.7} & 82.4\stdv{2.0}\colorup{2.6}\\
Soft-PL~\citep{assran2021semi} & &
78.3\stdv{3.8}\colorup{2.6} & 78.3\stdv{3.4}\down{1.8} & 95.3\stdv{1.3}\colorup{3.1} & 73.8\stdv{2.7}\colorup{3.1} & 82.7\stdv{3.7}\colorup{2.9}\\
S-CLIP~\citep{mo2023s} & &
78.3\stdv{3.2}\colorup{2.6} & 79.5\stdv{4.9}\down{0.6} & 93.8\stdv{1.5}\colorup{1.6} & {73.9}\stdv{5.0}\colorup{3.2} & 84.9\stdv{3.3}\colorup{5.1}\\
SemiCLIP~\citep{gan2025semi} & &
80.1\stdv{1.4}\colorup{4.4} & 82.8\stdv{1.0}\colorup{2.7} & 95.1\stdv{0.1}\colorup{2.9} & 75.0\stdv{0.8}\colorup{4.3} & 82.5\stdv{1.9}\colorup{2.7}\\
\rowcolor{gray!15}
\snameall (ours) & &  
\textbf{83.1}\stdv{1.4}\colorup{7.4} & 85.7\stdv{2.5}\colorup{5.6} & \textbf{96.7}\stdv{0.5}\colorup{4.5} & \textbf{75.1}\stdv{3.0}\colorup{4.4} & 85.6\stdv{3.0}\colorup{5.8}\\
\rowcolor{gray!15}
\snamedom (ours) & &  
82.8\stdv{1.3}\colorup{7.0} & \textbf{87.5}\stdv{1.9}\colorup{7.4} & 96.4\stdv{0.3}\colorup{4.2} & 74.5\stdv{2.0}\colorup{3.8} & \textbf{86.5}\stdv{1.8}\colorup{6.7}\\
\bottomrule
\end{tabular}
\end{adjustbox}
\end{table*}

\noindent\textbf{Zero-shot classification.} 
We evaluate zero-shot classification on the validation splits of the RSICD and UCM classification benchmarks, denoted as RSICD-CLS and UCM-CLS, respectively, and further assess generalization on WHU-RS19~\citep{xia2010structural}, RSSCN7~\citep{zou2015deep}, and AID~\citep{xia2017aid}. 
Table~\ref{tab:rs-zeroshot} summarizes the results.
Compared with SemiCLIP, both \snameall and \snamedom improve zero-shot performance on most datasets under both L$=$U and L$\neq$U settings.
Under L$=$U, \snameall achieves the best results on RSICD-CLS, WHU-RS19, and AID, while \snamedom performs best on UCM-CLS; both methods underperform SemiCLIP only on RSSCN7.
Under L$\neq$U, \snameall consistently improves SemiCLIP on all datasets, and \snamedom achieves the best results on UCM-CLS and AID, with the largest gain of 4.7 p.p. on UCM-CLS over SemiCLIP.
These results suggest that topology-aware alignment provides a complementary training signal that improves transferability.

\begin{table*}[t]
\centering
\caption{%
Image-text retrieval results on remote sensing datasets using the same setup as in Table~\ref{tab:rs-zeroshot}.
}\label{tab:rs-retrieval@r5}
\vspace{-0.08in}
\begin{adjustbox}{width=0.75\linewidth}
\begin{tabular}{@{}lccccccc@{}}
\toprule
& & \multicolumn{3}{c}{Image$\to$Text R@5} & \multicolumn{3}{c}{Text$\to$Image R@5} \\
\cmidrule(lr){3-5}\cmidrule(lr){6-8}
Method & Data & RSICD & UCM & Sydney & RSICD & UCM & Sydney \\
\midrule
CLIP (original) & - &
\phantom{0}12.6\blankstdv & 46.7\blankstdv & 44.8\blankstdv & 13.9\blankstdv & 39.5\blankstdv & 44.8\blankstdv \\
\midrule
CLIP (fine-tuned) & L &
25.7\stdv{2.3} & 55.2\stdv{0.8} & 46.6\stdv{3.0} & 24.9\stdv{0.6} & 56.3\stdv{1.6} & 49.4\stdv{2.0} \\
CLIP (oracle) & L + U &
34.3\stdv{1.3} & 63.5\stdv{1.3} & 63.6\stdv{2.4} & 32.0\stdv{1.3} & 66.6\stdv{1.1} & 65.4\stdv{2.5} \\
\midrule
Hard-PL~\citep{lee2013pseudo} & \multirow{6}{*}{L$=$U} &
27.3\stdv{1.1} & 54.8\stdv{1.6} & 52.3\stdv{3.6} & 24.2\stdv{1.3} & 54.6\stdv{1.0} & 55.2\stdv{6.9} \\
Soft-PL~\citep{assran2021semi} & &
27.3\stdv{0.9} & 54.8\stdv{2.2} & 52.3\stdv{2.0} & 26.1\stdv{1.7} & 55.6\stdv{4.0} & 56.3\stdv{5.3} \\
S-CLIP~\citep{mo2023s} & &
27.5\stdv{1.1} & 57.0\stdv{1.5} & 51.1\stdv{4.3} & 25.6\stdv{0.6} & 57.3\stdv{4.1} & 50.6\stdv{2.6} \\
SemiCLIP~\citep{gan2025semi} & &
27.9\stdv{0.7} & 57.3\stdv{2.6} & 52.3\stdv{4.3} & 25.8\stdv{0.2} & 59.4\stdv{1.2} & 58.6\stdv{2.4}\\
\rowcolor{gray!15}
\snameall (ours) & &  
\textbf{29.0}\stdv{1.5} & \textbf{61.4}\stdv{4.6} & 48.9\stdv{2.1} & \textbf{27.6}\stdv{1.2} & 60.8\stdv{3.0} & 56.3\stdv{3.5}\\
\rowcolor{gray!15}
\snamedom (ours) & &  
28.7\stdv{1.2} & 59.2\stdv{3.9} & \textbf{54.0}\stdv{2.9} & 27.0\stdv{0.9} & \textbf{63.0}\stdv{2.5} & \textbf{62.6}\stdv{2.9}\\
\midrule
Hard-PL~\citep{lee2013pseudo} & \multirow{6}{*}{L$\neq$U} &
25.9\stdv{1.6} & 52.7\stdv{2.2} & 50.6\stdv{7.0} & 22.4\stdv{1.5} & 54.3\stdv{1.4} & 51.7\stdv{5.2} \\
Soft-PL~\citep{assran2021semi} & &
27.4\stdv{1.6} & 54.1\stdv{2.4} & 51.1\stdv{1.0} & 24.4\stdv{0.7} & 54.4\stdv{2.4} & 52.3\stdv{5.3} \\
S-CLIP~\citep{mo2023s} & &
27.2\stdv{0.6} & 55.7\stdv{3.0} & 50.0\stdv{4.6} & 26.1\stdv{0.7} & 55.7\stdv{2.9} & 52.9\stdv{5.5} \\
SemiCLIP~\citep{gan2025semi} & &
26.1\stdv{0.9} & 57.9\stdv{1.0} & 54.6\stdv{0.8} & 25.1\stdv{0.2} & 59.4\stdv{1.5} & 59.2\stdv{1.6}\\
\rowcolor{gray!15}
\snameall (ours) & &  
28.0\stdv{0.3} & \textbf{60.6}\stdv{1.9} & 54.6\stdv{0.8} & \textbf{26.8}\stdv{0.3} & 60.5\stdv{1.7} & 56.9\stdv{1.4}\\
\rowcolor{gray!15}
\snamedom (ours) & &  
\textbf{28.2}\stdv{0.7} & 58.7\stdv{1.0} & \textbf{55.2}\stdv{0.0} & 26.4\stdv{1.0} & \textbf{61.3}\stdv{1.2} & \textbf{60.9}\stdv{0.8}\\
\bottomrule
\end{tabular}
\end{adjustbox}
\end{table*}

\noindent\textbf{Image-text retrieval.}
We evaluate image-text retrieval on the validation sets of RSICD, UCM, and Sydney, with R@5 results reported in Table~\ref{tab:rs-retrieval@r5}. 
Overall, both \snameall and \snamedom outperform SemiCLIP on most settings.
Under L$=$U, \snameall performs best on RSICD and UCM for image-to-text retrieval and on RSICD for text-to-image retrieval, while \snamedom performs best on Sydney in both directions and on UCM for text-to-image retrieval.
Under L$\neq$U, \snamedom improves SemiCLIP on all six retrieval metrics and achieves the best results on RSICD and Sydney for image-to-text retrieval and on UCM and Sydney for text-to-image retrieval, whereas \snameall performs best on UCM for image-to-text retrieval and RSICD for text-to-image retrieval.
This shows that $\mathcal{L}_{\text{\sname}}$ also benefits cross-modal retrieval.

\subsection{Fashion Datasets}
\label{sec:fashion}

\begin{table*}[t]
\centering
\caption{%
Zero-shot classification results on fashion datasets. Parentheses indicate the performance gap from the supervised CLIP, where values highlighted in {\color{cadmiumgreen}green} indicate gaps larger than one.
}\label{tab:fashion-zeroshot}
\vspace{-0.08in}
\begin{adjustbox}{width=\linewidth}
\begin{tabular}{@{}lccccc@{}}
\toprule
& \multicolumn{2}{c}{Fashion200k} & \multicolumn{2}{c}{FashionGen} & Polyvore \\
\cmidrule(lr){2-3}\cmidrule(lr){4-5}\cmidrule(lr){6-6}
Method & Super-class & Sub-class & Super-class & Sub-class & Class \\
\midrule
CLIP (original) &
73.2\blankstdv & 27.7\blankstdv & 34.8\blankstdv & 26.4\blankstdv & 70.2\blankstdv \\
\midrule
CLIP (fine-tuned) &
79.0\stdv{3.5}\blankup & 35.1\stdv{0.7}\blankup & 35.4\stdv{8.1}\blankup & 24.5\stdv{2.4}\blankup & 60.4\stdv{2.3}\blankup \\
CLIP (oracle) &
88.6\stdv{1.3}\blankup & 45.3\stdv{4.8}\blankup & 53.3\stdv{2.5}\blankup & 45.5\stdv{7.3}\blankup & 76.3\stdv{8.8}\blankup \\
\midrule
Hard-PL~\citep{lee2013pseudo} &
\,\>54.9\stdv{6.4}\down{24.1} & \,\>23.9\stdv{2.3}\down{11.2} & \,\>24.2\stdv{3.9}\down{11.2} & 18.3\stdv{2.4}\down{6.2} & 34.3\stdv{6.8}\down{26.1} \\
Soft-PL~\citep{assran2021semi} &
82.5\stdv{2.8}\colorup{3.5} & 36.6\stdv{1.4}\colorup{1.5} & 44.8\stdv{3.5}\colorup{9.4} & 33.6\stdv{1.4}\colorup{9.1} & 73.6\stdv{1.7}\colorup{13.2} \\
S-CLIP~\citep{mo2023s} &
85.1\stdv{0.9}\colorup{6.1} & 38.4\stdv{0.7}\colorup{3.3} & 44.0\stdv{4.6}\colorup{8.6} & 29.6\stdv{4.0}\colorup{5.1} & 73.9\stdv{2.4}\colorup{13.5} \\
SemiCLIP~\citep{gan2025semi} &
84.5\stdv{1.8}\colorup{5.5} & 43.7\stdv{1.2}\colorup{8.6} & 43.4\stdv{3.9}\colorup{8.0} & \,\>43.0\stdv{2.6}\colorup{18.5} & 73.0\stdv{2.1}\colorup{12.6}\\
\rowcolor{gray!15}
\snameall (ours) & 
\textbf{87.3}\stdv{0.1}\colorup{8.3} & \textbf{45.0}\stdv{0.6}\colorup{9.9} & \,\>48.0\stdv{1.9}\colorup{12.6} & \,\>40.9\stdv{3.9}\colorup{16.4} & \textbf{74.0}\stdv{2.8}\colorup{13.6}\\
\rowcolor{gray!15}
\snamedom (ours) & 
86.2\stdv{1.4}\colorup{7.2} & 44.7\stdv{0.4}\colorup{9.6} & \,\>\textbf{49.1}\stdv{1.8}\colorup{13.7} & \,\>\textbf{46.5}\stdv{2.5}\colorup{22.0} & 72.0\stdv{2.9}\colorup{11.6}\\
\bottomrule
\end{tabular}
\end{adjustbox}
\end{table*}

We conduct experiments on three fashion datasets: Fashion200k~\citep{han2017automatic}, FashionGen~\citep{rostamzadeh2018fashion}, and Polyvore Outfits~\citep{vasileva2018learning}.
We train the models on the union of these datasets and evaluate them separately on each dataset.

\noindent\textbf{Zero-shot classification.}
Table~\ref{tab:fashion-zeroshot} reports the zero-shot classification results.
Compared with SemiCLIP, \sname improves performance in most settings.
The gains are particularly notable on FashionGen super-class classification, 5.7 p.p. over SemiCLIP.
The domain-wise variant improves the results on FashionGen, achieving 49.1\% and 46.5\% on super-class and sub-class classification, respectively, and surpassing the CLIP oracle on sub-class.
In contrast, the batch-wise variant performs best on Fashion200k and Polyvore, obtaining 87.3\%, 45.0\%, and 74.0\%, respectively.

\noindent\textbf{Image-text retrieval.}
On the standard fashion retrieval benchmarks, \snamedom improves SemiCLIP on five of six R@5 metrics, including all three image-to-text settings, although the aggregate gains remain modest (Appendix~\ref{app:fashion_retrival}).
To better understand this behavior, we conduct a controlled one-pair-per-product analysis on 7,519 FashionGen validation products (Appendix~\ref{app:fashion_retrieval_analysis}).
We divide queries into equally sized groups according to the bidirectional positive-negative retrieval margin of SemiCLIP.
For the lowest-margin third, \sname improves R@5 from 8.14 to 19.94 for I$\rightarrow$T and from 7.66 to 20.10 for T$\rightarrow$I, and improves the ground-truth rank for approximately 61\% of queries.
In contrast, performance decreases for some high-margin queries whose matched pairs are already well separated, partially offsetting these gains in aggregate Recall.
Meanwhile, \sname improves $H_0$ cross-modal neighbor overlap from 22.57\% to 24.45\% and edge-direction agreement from 33.21\% to 42.77\%, together with consistent improvements in semantic neighborhood coherence.
These results indicate that topology-aware alignment improves cross-modal structural agreement and poorly resolved queries, while changes in the fine-grained ordering of already well-separated pairs limit the aggregate exact-pair retrieval gain.


\subsection{%
Ablation Study and Analysis
}


\begin{figure}[!ht]
\centering\small
\begin{subfigure}{0.57\textwidth}
\centering\small
\includegraphics[width=\linewidth]{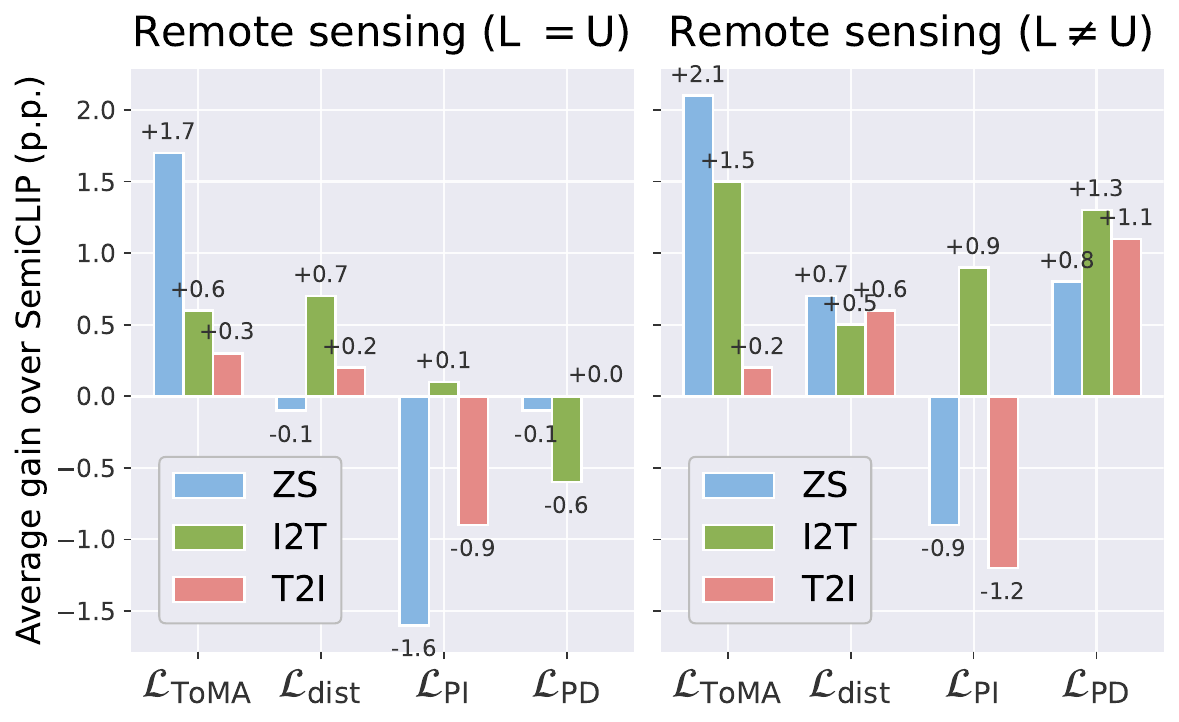}
\caption{Comparison with alternative topology-aware objectives}\label{fig:tda_ablation-1}
\end{subfigure}~
\begin{subfigure}{0.42\textwidth}
\centering\small
\includegraphics[width=\linewidth]{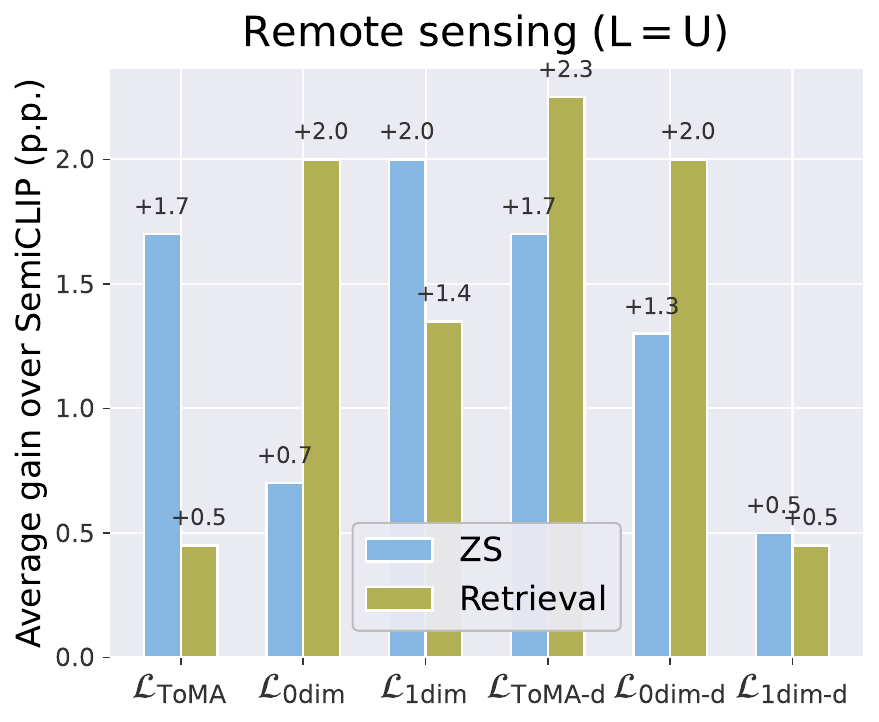}
\caption{Homology dimension analysis}\label{fig:tda_ablation-2}
\end{subfigure}
\caption{Comparison with prior topology-aware objectives and ablation on homology dimensions. 
In (a) and (b), bars denote average gains over SemiCLIP for zero-shot classification (ZS), image-to-text retrieval (I2T), text-to-image retrieval (T2I) and average of I2T and T2I (Retrieval).
Here, $\mathcal{L}_{\mathrm{PD}}$ corresponds to the objective used in ToMCLIP, while $\mathcal{L}_{\mathrm{0dim}}$ is equivalent to
HC.}
\label{fig:tda_ablation}
\end{figure}

\noindent\textbf{Comparison with alternative topology-aware objectives.}
We compare $\mathcal{L}_{\text{\sname}}$ with several alternative topology-aware objectives while keeping the underlying SemiCLIP framework and training protocol unchanged. These objectives differ in what topological information is used as the alignment target.
Topological Autoencoders~\citep{moor2020topological} preserve persistent homology (PH)-selected pairwise distances.
Following this idea, we define $\mathcal{L}_{\text{dist}}$ using the same PH-induced relations as ToMA, but replace the directional term $1-\left\vert T^*_\mu(M_i\to M_j)\right\vert$ in
\eqref{eq:tomar_loss} with
$
\left\vert
\left\lVert x_0^{(\mu_*)}-x_1^{(\mu_*)}\right\rVert
-
\left\lVert
\pi_{i\to j}(x_0^{(\mu_*)})
-
\pi_{i\to j}(x_1^{(\mu_*)})
\right\rVert
\right\vert .
$
Thus, $\mathcal{L}_{\text{dist}}$ aligns the lengths of PH-induced
relations, whereas $\mathcal{L}_{\text{\sname}}$ aligns their
directions.

We also compare against objectives that align topological summaries.
TopKD~\citep{kim2024topological} represents persistence diagrams (PDs)
as persistence images and matches these vectorized summaries, which we
denote by $\mathcal{L}_{\text{PI}}$.
PsHD~\citep{yanfan2024persistence} and
ToMCLIP~\citep{you2026topological} instead align PDs using the
Wasserstein distance, denoted by $\mathcal{L}_{\text{PD}}$.
Unlike these summary-level objectives, ToMA uses the available
cross-modal correspondence map $\pi_{i\to j}$ to directly align the
relations associated with $H_0$-death and $H_1$-birth events.

For a controlled comparison, we replace only
$\mathcal{L}_{\text{\sname}}$ with each alternative objective while
keeping the remaining training components fixed.
As shown in Fig.~\ref{fig:tda_ablation-1},
$\mathcal{L}_{\text{\sname}}$ provides the most consistent improvement,
particularly for zero-shot classification, whereas
$\mathcal{L}_{\text{dist}}$, $\mathcal{L}_{\text{PI}}$, and
$\mathcal{L}_{\text{PD}}$ show more variable behavior across
classification and retrieval settings.

\noindent\textbf{Contribution of $H_1$.}
We compare $\mathcal{L}_{\text{\sname}}$ with $H_0$-only
($\mathcal{L}_{0\text{dim}}$, equivalent to HC~\citep{zhang2024homology})
and $H_1$-only ($\mathcal{L}_{1\text{dim}}$) variants, together with their
domain-wise counterparts. As shown in Fig.~\ref{fig:tda_ablation-2}, the
$H_1$-only variant already improves over SemiCLIP, demonstrating that $H_1$
provides useful structural information beyond $H_0$. Combining both dimensions
yields the strongest overall results, particularly in the domain-wise setting,
suggesting that cycle-closing $H_1$ relations provide complementary structural
signals for cross-modal alignment.


\begin{figure}[t]
\centering\small
\includegraphics[width=\linewidth]{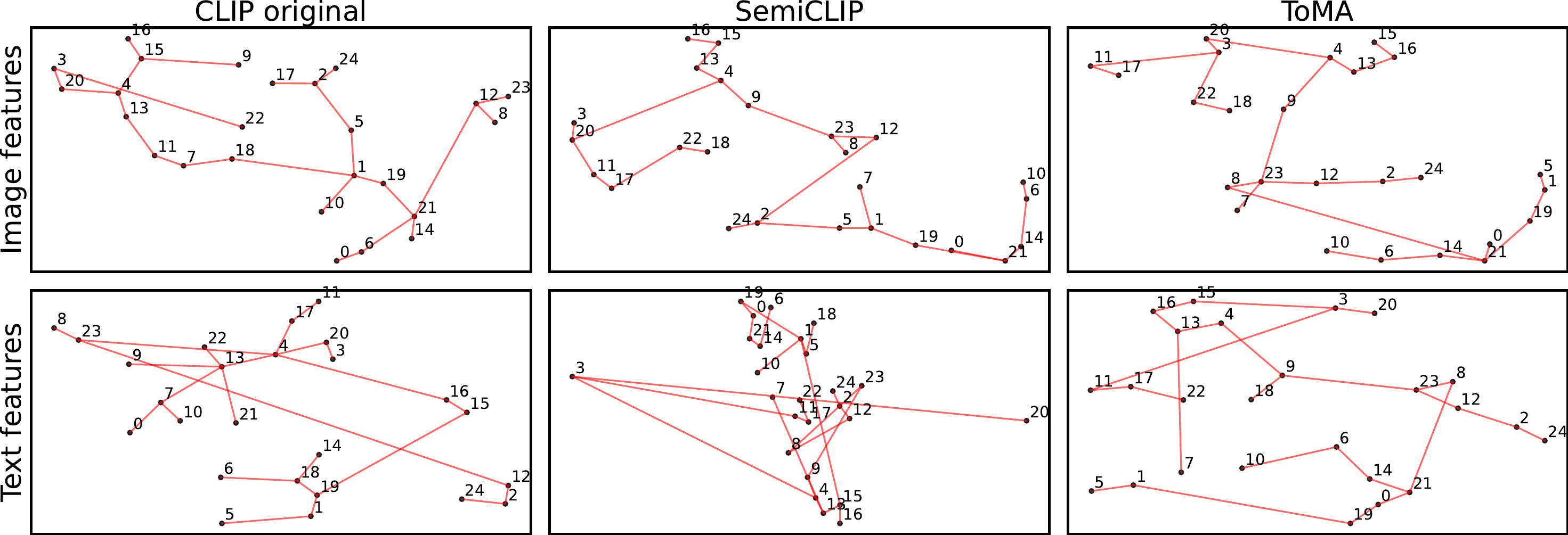}
\caption{Topological structure analysis of image and text embeddings across methods.}
\label{fig:tda_ablation-3_main}
\end{figure}

\begin{wraptable}{r}{0.36\columnwidth}
    \centering
    \vspace{-8pt}
    \caption{Cross-modal MST overlap under the L$=$U setting.}
    \label{tab:mst_overlap_m}
    \setlength{\tabcolsep}{3pt}
    \begin{tabular}{@{}lc@{}}
        \toprule
        Method & Overlap \\
        \midrule
        CLIP           & 22.11\% \\
        SemiCLIP       & 31.74\% \\
        \sname & \textbf{34.26}\% \\
        \snamedom & \textbf{33.57}\% \\
        \bottomrule
    \end{tabular}
    \vspace{-6pt}
\end{wraptable}

\noindent\textbf{Cross-modal structural alignment.}
To quantify whether corresponding samples induce consistent connectivity structure across modalities, we construct MSTs separately from the image and text embeddings and measure their normalized edge overlap. Table~\ref{tab:mst_overlap_m} shows that \sname consistently increases this cross-modal MST agreement over SemiCLIP in both settings and across seeds (Appendix~\ref{app:mst_overlap}).
Figure~\ref{fig:tda_ablation-3_main} provides a complementary qualitative view: while CLIP and SemiCLIP exhibit substantial cross-modal connectivity mismatch,
\sname produces a more consistently organized connectivity backbone.
In particular, the chain
$15\!\rightarrow\!16\!\rightarrow\!13\!\rightarrow\!4\!\rightarrow\!9
\!\rightarrow\!23\!\rightarrow\!12\!\rightarrow\!2\!\rightarrow\!24$
is more clearly preserved across both modalities, linking stadium-related and
residential/green-building branches. Together, these results provide evidence
that \sname reduces cross-modal structural mismatch 
(Appendices~\ref{app:toma_qual} and~\ref{app:mst_examples}).

\noindent\textbf{Comparison with topology-free graph baselines.}
To separate the effect of the additional alignment objective from the effect of using topology to construct the aligned relations, we replace the PH-based edge construction with random and $k$-NN graphs while retaining the same encoders, directional alignment objective, and training protocol. 
The random baseline uniformly samples pairwise relations, whereas the $k$-NN baseline uses nearest-neighbor relations in the current representation space; their edge counts are approximately matched to the $H_0$-death relations. The suffix ``-d'' denotes the corresponding variants in which graph construction is performed separately within each source-domain partition. Further implementation details are provided in Appendix~\ref{app:graph_baselines}.

Table~\ref{tab:graph_baselines} summarizes the average gains over SemiCLIP. Under L$=$U, ToMA yields a larger zero-shot gain (+1.8 p.p.) than the random (+0.7) and $k$-NN ($-0.3$) controls, while the standard $k$-NN baseline is more competitive for retrieval (+0.8 vs.\ +0.1 for ToMA). In the domain-partitioned setting, ToMA-d achieves the largest retrieval gain (+1.9). Under L$\neq$U, ToMA and ToMA
\begin{wraptable}{r}{0.42\columnwidth}
\centering
\caption{\textbf{Comparison with topology-free graph baselines.}
Average gains over SemiCLIP (p.p.). 
Ret. denotes the average I2T/T2I R@5.}
\label{tab:graph_baselines}
\vspace{-0.4\baselineskip}
\begin{tabular}{@{}lrrrr@{}}
\toprule
& \multicolumn{2}{c}{L$=$U}
& \multicolumn{2}{c}{L$\neq$U} \\
\cmidrule(lr){2-3}\cmidrule(l){4-5}
Method & ZS & Ret. & ZS & Ret. \\
\midrule
ToMA
& \textbf{+1.8} & +0.1
& +2.1 & +0.8 \\

Random
& +0.7 & -0.3
& +1.3 & +1.1 \\

$k$-NN
& -0.3 & +0.8
& +1.3 & +1.1 \\

ToMA-d
& +1.7 & \textbf{+1.9}
& \textbf{+2.4} & \textbf{+1.4} \\

Random-d
& -0.3 & +0.6
& +1.3 & +1.1 \\

$k$-NN-d
& -0.3 & +0.6
& +1.3 & +1.1 \\
\bottomrule
\end{tabular}

\end{wraptable}
-d again provide larger zero-shot gains (+2.1 and +2.4) than the topology-free controls (+1.3), whereas random and $k$-NN remain competitive with standard ToMA for retrieval. Overall, these results show that generic graph regularization can itself be beneficial, but the topology-induced construction provides additional advantages, particularly for zero-shot classification and domain-partitioned retrieval.

\noindent\textbf{Effect of domain partitioning.}
We further examine whether the performance of \snamedom can be explained by domain partitioning itself. To this end, we evaluate SemiCLIP-domain, which applies the SemiCLIP consistency objectives separately within each source-domain subset without using the ToMA objective. As reported in Appendix~\ref{app:domain_partition_control}, domain partitioning alone provides only a marginal change in averaged zero-shot accuracy, from 84.30 for SemiCLIP to 84.48 for SemiCLIP-domain. In comparison, ToMA and ToMA-domain achieve 86.06 and 85.96, respectively. These results suggest that the main improvement is associated with the ToMA objective rather than domain partitioning itself, while domain-wise topology construction serves as an optional, dataset-dependent extension.

\noindent\textbf{Robustness.}
Additional experiments show that \sname remains effective across
batch sizes of 32-256 and that the extracted connectivity structure
is stable under moderate batch perturbations
(Appendix~\ref{app:bs64}, \ref{app:bs}, and~\ref{app:mst_stability}).
It is also robust to moderate choices of the loss coefficient and
relative $H_0$/$H_1$ weights (Appendix~\ref{app:loss_co}).
Across labeled-pair ratios of 10\%, 20\%, and 30\%, \sname improves
the five-dataset average over SemiCLIP by approximately 1.8, 2.3,
and 0.6 p.p., respectively, outperforming SemiCLIP on four of the
five datasets at each ratio; the gain is therefore maintained across
label regimes but is not monotonic with the labeled ratio
(Appendix~\ref{app:ssl_ratio}).
Moreover, \snamedom matches or improves upon SemiCLIP in 25 of 27
seed-level aggregate comparisons (Appendix~\ref{app:exp_seed}).

\noindent\textbf{Computational efficiency.}
Under a controlled comparison on a single NVIDIA A100 GPU with a batch size of 256, ToMA increases the total training time of SemiCLIP only from 799.5 to 802.0 seconds, corresponding to a 0.3\% overhead. The Stage-1 run time increases from 19.2 to 19.3 seconds per epoch, while the Stage-2 run time remains unchanged at 21.3 seconds per epoch.
This small overhead results from our lightweight PH computation, which extracts $H_0$-death and $H_1$-birth relations from the 1-skeleton without constructing 2-simplices (Appendix~\ref{app:run_time}).




\section{Conclusion}

We presented \sname, a topology-aware alignment framework for semi-supervised vision-language learning that uses persistent homology to identify $H_0$-death and lightweight $H_1$-birth relations and align their directions across modalities. Across specialized-domain benchmarks, \sname yields strong zero-shot classification gains and task-dependent retrieval benefits, with clear improvements on remote sensing and stronger effects on low-margin fashion queries and semantic neighborhood organization. These results show that topology-aware structural alignment provides a complementary signal beyond pairwise cross-modal alignment. 

\noindent\textbf{Limitations.}
Our evaluation is limited to specialized-domain semi-supervised vision-language adaptation, focusing primarily on remote sensing and fashion datasets. 
The additional retrieval experiments on science figures and comics provide only limited evidence beyond these domains and do not establish broader generalization to other multimodal or unimodal alignment settings.
Further limitations are presented in Appendix~\ref{app:limitations}.

\noindent\textbf{Broader impact.}
\sname may help improve domain-specific retrieval and classification when labeled image-text pairs are scarce. At the same time, stronger multimodal alignment may be used in sensitive applications, especially in remote sensing, and may inherit biases from pretrained models and datasets, requiring careful evaluation before deployment in high-stakes settings.
Further discussion of broader impacts is presented in Appendix~\ref{app:impacts}.





\begin{ack}

This research was supported by Basic Science Research Program through the National Research Foundation of Korea (NRF), funded by the Ministry of Education (No. RS-2025-25397599); the Basic Science Research Institute Fund through the NRF (No. RS-2021-NR060139); the NRF grant funded by the Korea government (MSIT) (No. RS-2023-00219980); and the National Institute for Mathematical Sciences (NIMS) as a subproject of Project No. S26120000. This research was also supported in part by grants from the Institute of Information \& Communications Technology Planning \& Evaluation (IITP), funded by the Korea government (MSIT): No. RS-2026-25524173 (UltraLong-Term Hierarchical Memory and Reasoning Architecture for Next-Generation Omnimodal Agents) and the Global Data-X Leader HRD Program (No. IITP-2026-RS-2024-00441244).

\end{ack}

\bibliographystyle{unsrtnat}
\bibliography{reference}


\clearpage
\hypersetup{linkcolor=black}
\etocdepthtag.toc{mtappendix}
\etocsettagdepth{mtchapter}{none}
\etocsettagdepth{mtappendix}{subsection}
\tableofcontents
\hypersetup{linkcolor=red}

\appendix
\newpage
\section{Qualitative and Quantitative Analysis}
\label{app:tomar_qualitative_analysis}

\subsection{Cross-Modal Structural Alignment: Qualitative Analysis}
\label{app:toma_qual}

\begin{figure}[!ht]
\centering\small
\includegraphics[width=\linewidth]{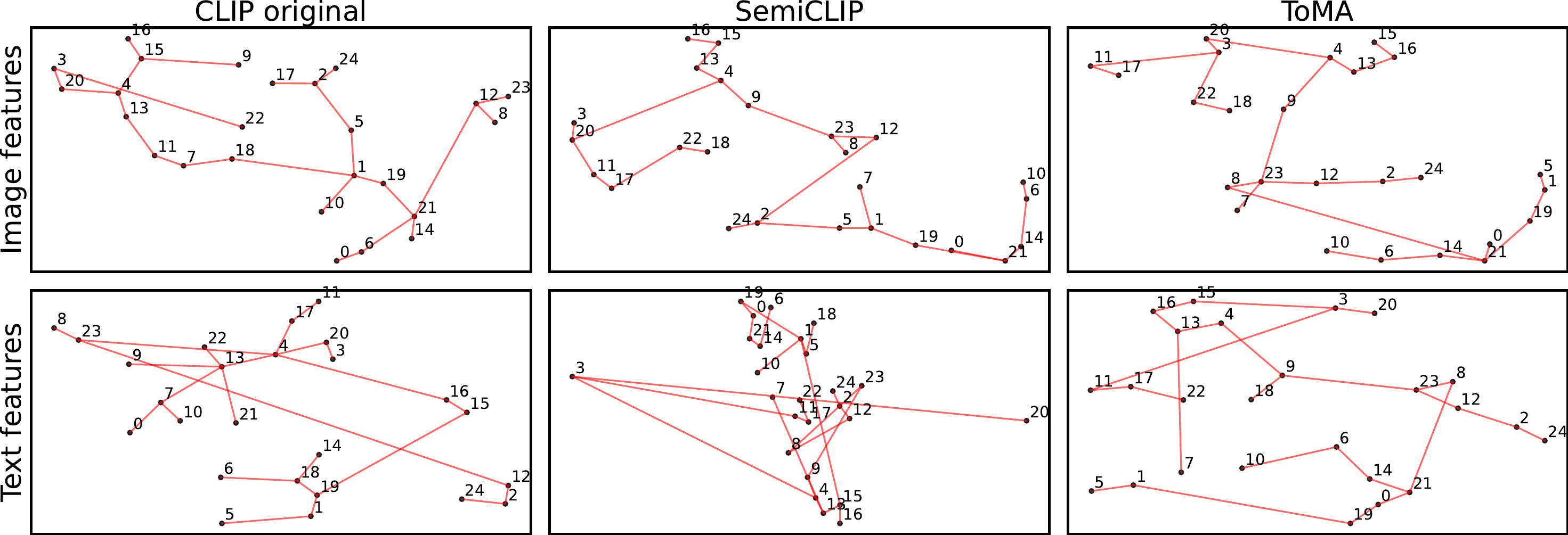}
\caption{Topological structure analysis of image and text embeddings across methods.}
\label{fig:tda_ablation-3}
\end{figure}

Figure~\ref{fig:tda_ablation-3} compares the minimum spanning trees of the embeddings in Figure~\ref{fig:figs_example}. 
While the original CLIP shows different connectivity patterns across image and text and SemiCLIP only partially recovers shared local structure, \sname yields much stronger cross-modal agreement. 
In particular, \sname preserves a larger connected backbone that is consistently organized across image and text, including the chain $15 \!\rightarrow\! 16 \!\rightarrow\! 13 \!\rightarrow\! 4 \!\rightarrow\! 9 \!\rightarrow\! 23 \!\rightarrow\! 12 \!\rightarrow\! 2 \!\rightarrow\! 24$. 
This connected structure links a stadium-related branch to a
residential/green-building branch and is much less clearly aligned in SemiCLIP, especially in the text embedding space. These observations suggest that ToMA better preserves relational topology across modalities than SemiCLIP. 

\subsection{Illustrative Example of \sname with the Embeddings in Figure~2b }
\label{app:tomar_example}

\begin{figure}[!t]
  \centering\small
  \includegraphics[width=\linewidth]{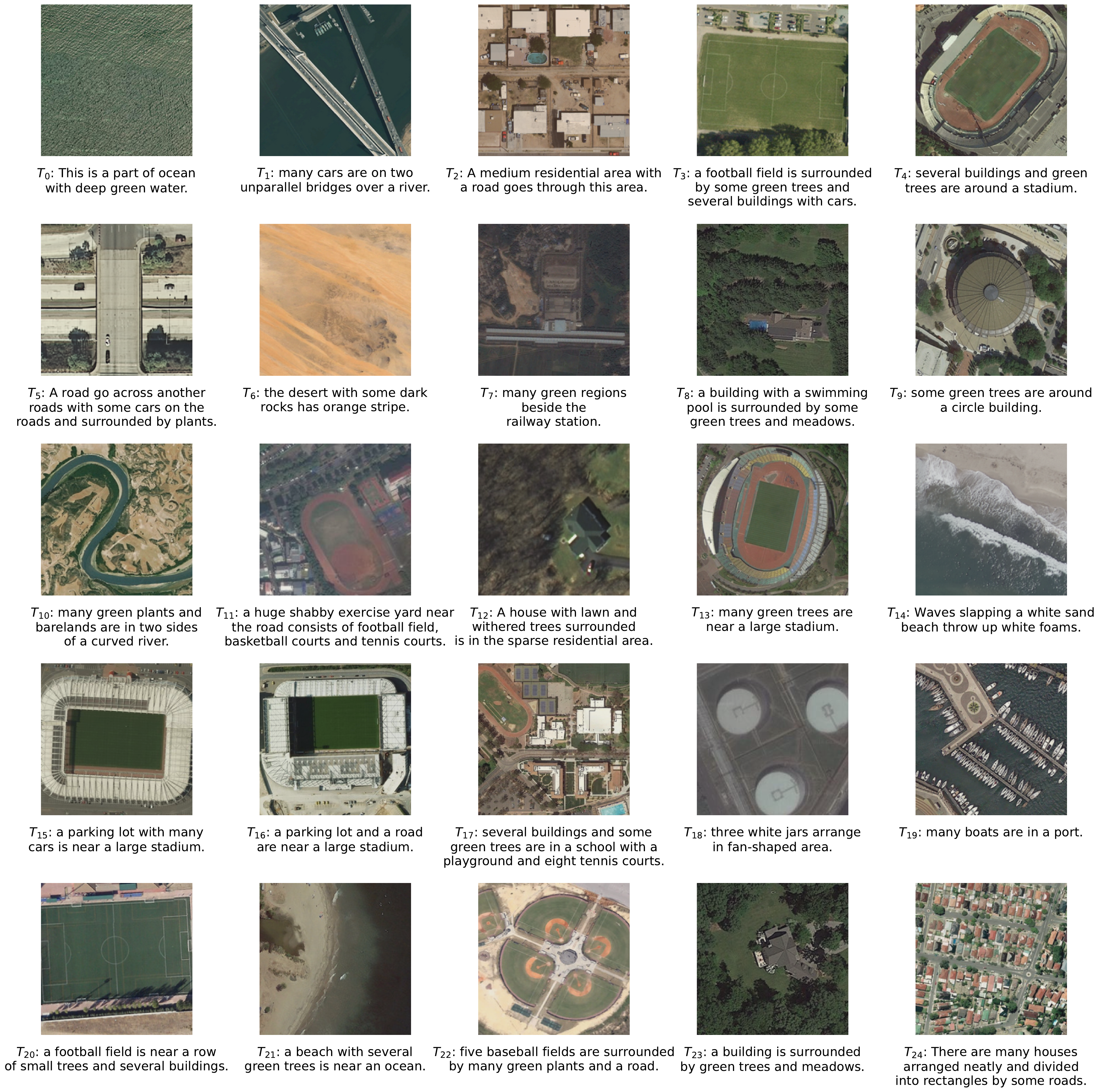} 
  \caption{The real image-text data used to generate the embeddings in Figure~\ref{fig:method}\textcolor{red}{b} (0--15) and Figure~\ref{fig:tda_ablation-3}.}
  \label{fig:figs_example}
\end{figure}

To provide a more intuitive understanding of \sname, we illustrate how the loss operates on the real image-text pairs shown in Figure~\ref{fig:method}\textcolor{red}{b} and Appendix Figure~\ref{fig:figs_example} ($T_0$--$T_{15}$). 
In this example, a mini-batch consists of 16 paired samples indexed from \(0\) to \(15\), where each image embedding \(I_i\) is matched with its corresponding text embedding \(T_i\).

These samples exhibit several semantically related groups.
For example, samples \(0\) and \(14\) both describe water scenes (ocean and beach), while samples \(4\), \(13\), and \(15\) are all related to stadium-like scenes.
Other samples form additional local semantic neighborhoods, such as residential scenes (\(2,12\)) or vegetation-rich scenes (\(8,9\)).
Thus, the batch contains multiple semantically meaningful local clusters together with higher-level relations between clusters.

Under the 1-skeleton filtration, the edge set of each modality decomposes into two types of topologically salient edges.
The first type consists of the \emph{\(H_0\)-death edges}, which form the minimum spanning tree (MST).
These edges connect points that merge previously disconnected components, and therefore mainly capture direct neighborhood relations.
In the example of Figure~\ref{fig:method}\textcolor{red}{b}, such edges connect samples within semantically coherent local groups, e.g., water-related samples or stadium-related samples.
From a representation-learning perspective, they encode \emph{local semantic proximity}.

The second type consists of the \emph{\(H_1\)-birth edges}, which are non-MST edges that close cycles.
Unlike \(H_0\)-death edges, an \(H_1\)-birth edge does not simply indicate direct proximity between two samples.
Rather, it closes a loop over an existing path of \(H_0\)-death edges and thereby reflects how multiple local neighborhoods are integrated into a larger relational pattern.
For example, in Figure~\ref{fig:method}\textcolor{red}{b}, the path \(2 \rightarrow 12 \rightarrow 8 \rightarrow 0 \rightarrow 14\) is first established through \(H_0\)-death edges.
This path begins with residential scenes (\(2,12\)), passes through an intermediate scene of a house-like structure surrounded by greenery (\(8\)), and then reaches water scenes (\(0,14\)).
Accordingly, the path does not merely connect the closest samples, but organizes several semantically adjacent neighborhoods into a longer chain.
An \(H_1\)-birth edge that closes a cycle over this chain therefore captures higher-order semantic organization beyond immediate pairwise similarity.
In this sense, \(H_0\)-death edges preserve local semantic structure, whereas \(H_1\)-birth edges preserve higher-order semantic organization induced by paths formed by \(H_0\)-death edges.

\sname aligns these topological relations across modalities by comparing corresponding edges in the image and text embedding spaces.
For an edge \((i,j)\) selected as a topological simplex in one modality, \sname considers the paired relation between the corresponding samples \((I_i,I_j)\) and \((T_i,T_j)\).
The loss encourages these two edge directions to be aligned, yielding the objective
\[
\mathcal{L}(i,j)
=
1-|\left\langle I_i-I_j,\; T_i-T_j \right\rangle|,
\]
where the inner product is computed after normalization.

This example clarifies the complementary roles of $H_0$ and $H_1$. The $H_0$-death edges constrain the connectivity backbone of the representation, while the $H_1$-birth edges provide additional cycle-closing relations among already connected regions. Their combination therefore captures complementary connectivity and higher-order relational structure across modalities.
As a result, \sname regularizes image and text encoders not only to match individual pairs, but also to preserve the topology of the underlying data manifolds at both local and higher-order levels.

\subsection{Quantitative Evidence of Cross-Modal Topology Mismatch}
\label{app:mst_overlap}

Building on the illustrative example in Appendix~\ref{app:tomar_qualitative_analysis},
Figure~\ref{fig:tda_ablation-3} qualitatively compares the image and
text MSTs constructed from the same paired samples.
The CLIP and SemiCLIP representations exhibit many
modality-specific connectivity relations. This indicates that SemiCLIP improves pairwise cross-modal consistency without fully aligning the batch-level connectivity backbones of the two modalities. In contrast, \sname produces visibly more consistent image-text MST structures.

To verify that this observation is systematic rather than specific to a single batch, we measure the normalized MST overlap
\begin{equation}
    A_{\mathrm{MST}}
    =
    \frac{\left|M_I \cap M_T\right|}{N-1},
    \label{eq:mst_overlap}
\end{equation}
where $M_I$ and $M_T$ denote the image and text MST edge sets constructed from the same batch of $N$ paired samples. Since the $H_0$-death edges under our 1-skeleton filtration form the MST, this metric directly measures the agreement between the cross-modal $H_0$ connectivity backbones.

We evaluate 500 batches, each containing 128 image-text pairs sampled from RSICD, UCM, and Sydney according to their training-set proportions. We use the same batch composition and stochastic caption-selection procedure as
during training.

\begin{table}[!ht]
    \centering
    \caption{Normalized overlap between the image and text MST edge sets under the L$=$U setting.
    Higher values indicate stronger agreement between the cross-modal
    $H_0$ connectivity backbones.}
    \label{tab:mst_overlap}
    \begin{tabular}{lc}
        \toprule
        Method & MST overlap (\%) \\
        \midrule
        CLIP (original)       & 22.11 \\
        SemiCLIP              & 31.74 \\
        \sname          & \textbf{34.26} \\
        \snamedom      & \textbf{33.57} \\
        \bottomrule
    \end{tabular}
\end{table}

As shown in Table~\ref{tab:mst_overlap}, SemiCLIP substantially increases MST agreement over the pretrained CLIP model, from 22.11\% to 31.74\%. Nevertheless, considerable cross-modal topology mismatch remains. \sname and \snamedom further increases the overlap to 34.26\% and 33.57\%, corresponding to improvements of 2.51 and 1.83 percentage points over SemiCLIP, respectively. The improvement is observed for every random seed in both ToMA variants.

Together, Figure~\ref{fig:tda_ablation-3} and the quantitative analysis in Table~\ref{tab:mst_overlap} show that cross-modal topology mismatch is a systematic residual limitation of SemiCLIP rather than an artifact of a particular example. \sname consistently reduces this mismatch, and the increased MST agreement is accompanied by improved average zero-shot classification and image-text retrieval performance. These
results support the role of topology-aware alignment as a complementary structural objective beyond pairwise cross-modal consistency.

\subsection{Additional Examples of 0 and 1-Dimensional Edges}
\label{app:mst_examples}

\begin{figure}[!ht]
  \centering\small
  \includegraphics[width=\linewidth]{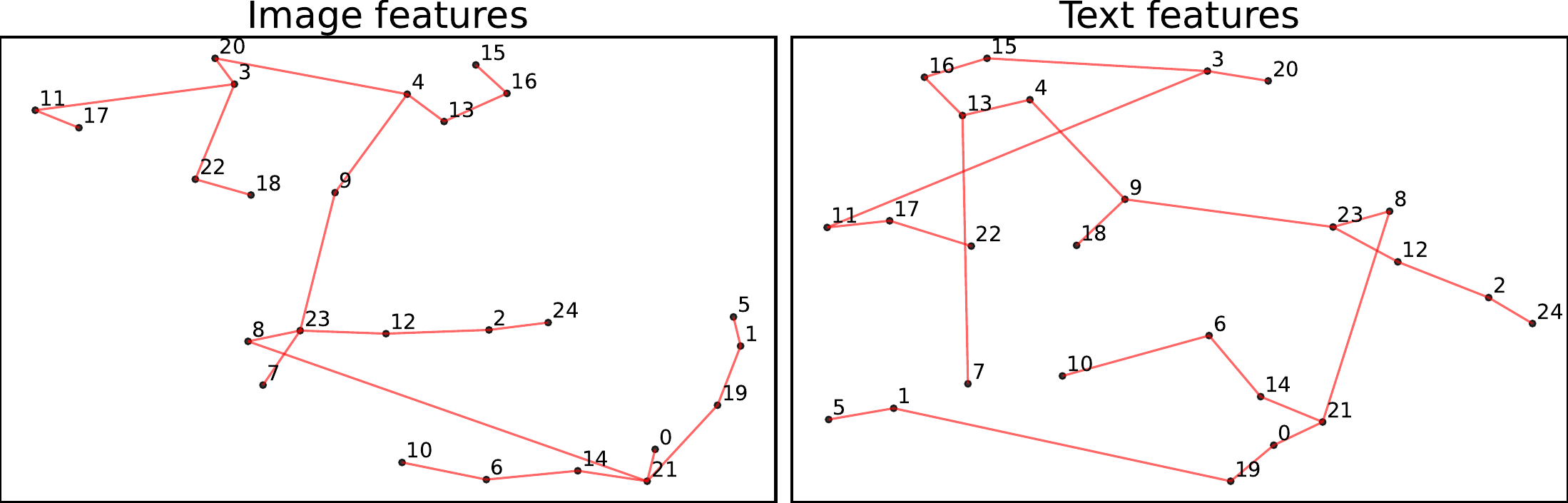} 
  \caption{Minimum spanning trees of the image and text embeddings from the image-text pairs shown in Figure~\ref{fig:figs_example}.}
  \label{fig:mst}
\end{figure}

\begin{figure}[!ht]
  \centering\small
  \includegraphics[width=\linewidth]{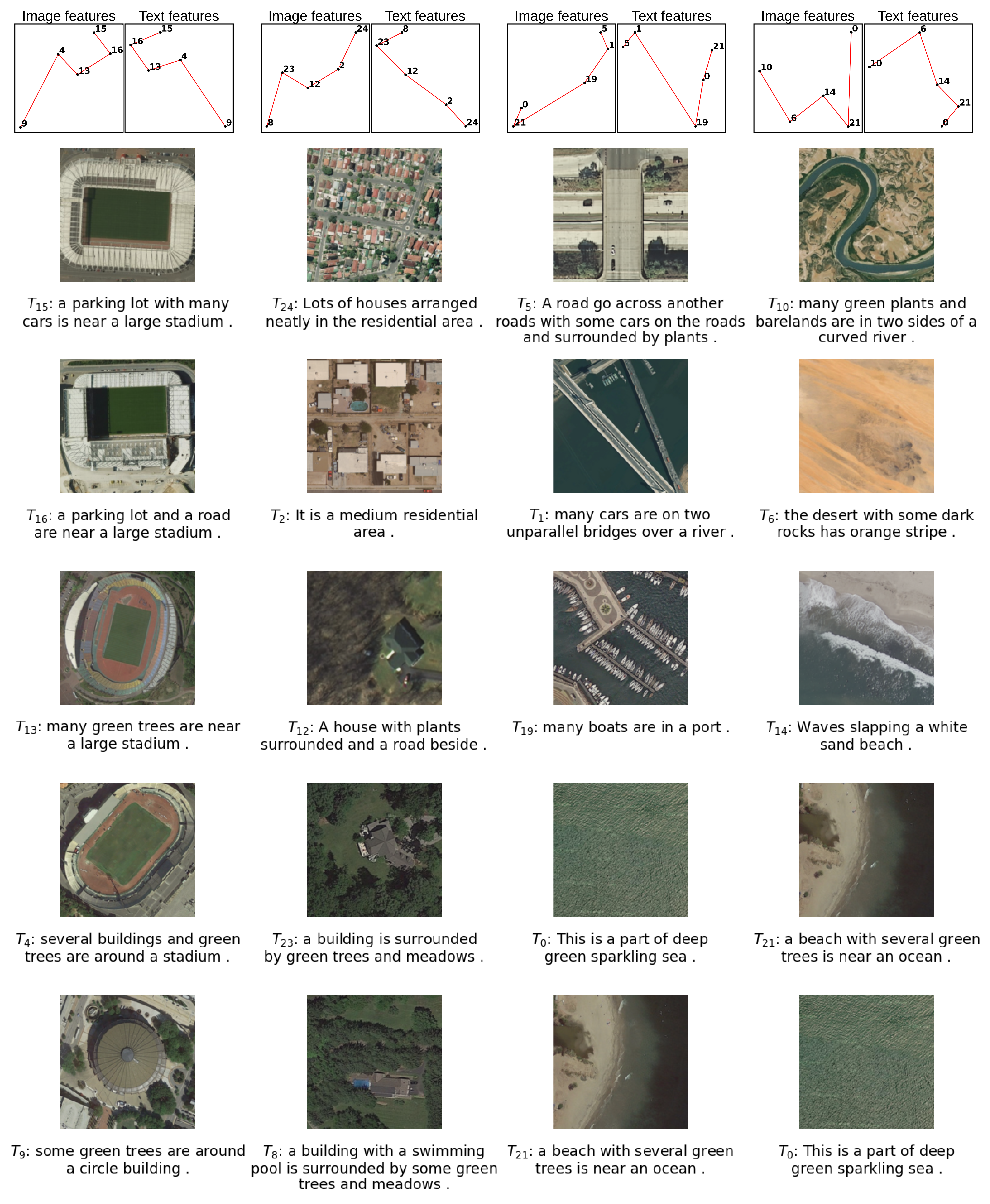} 
  \caption{Visual and semantic examples of topologically salient relations identified by \sname.}
  \label{fig:mst_part}
\end{figure}

To further understand the structural differences highlighted in Figure~\ref{fig:tda_ablation-3}, we provide a qualitative analysis of the $H_0$- and $H_1$-based edges extracted by \sname on the remote sensing dataset. 
As shown in the main text, the original CLIP fails to align the overall MST skeletons across image and text, while SemiCLIP recovers some local structure but still leaves mismatched long-range connections and a less coherent global organization, especially in the text space. 
The examples below illustrate how \sname addresses these limitations by preserving both semantically meaningful local clusters and longer-range topological relations across modalities.
Figure~\ref{fig:mst} visualizes the MST structure formed by 25 samples in Figure~\ref{fig:figs_example}, while Figure~\ref{fig:mst_part} displays the actual images and captions for representative edges selected from this structure.

As shown in the individual edge configurations in Figure~\ref{fig:mst_part}, the $H_0$-death edges do not merely connect points based on proximity; rather, they cluster samples by leveraging visual and semantic similarities. In the first column of Figure~\ref{fig:mst_part}, the edges sequentially connect samples that share distinctive visual features of stadiums. In the second column, the edges group samples under the common theme of residential areas. Notably, the semantic flow from high-density residential areas ($24$) to vegetation-mixed low-density residential areas ($12, 23, 8$) is well-preserved along the $H_0$-death edges.

Furthermore, $H_1$-birth edges complete the logical connections between different semantic clusters, thereby forming global cycles. The connection path $10 \to 6 \to 14 \to 21 \to 0$ in the last column of Figure~\ref{fig:mst_part} illustrates a significant geographical transition spanning from a river ($10$) and desert ($6$) to  a beach ($14$) and the ocean ($21, 0$). By aligning the $H_1$-birth edges that close these long-range paths into cycles, \sname regularizes the model to maintain the topological characteristics of the global data manifold, consistently preserving transitions that start from a river, pass through the sea, and return to terrestrial structures across both modalities.

Comparing the image and text features in Figure~\ref{fig:mst}, several corresponding
connectivity patterns are preserved across the two modalities. In particular,
the stadium- and residential-related branches observed in the image space are
also reflected in the text space. This suggests that \sname preserves
batch-level relational organization beyond individual image-text pairs.
Together, these examples help explain the increased cross-modal MST agreement
observed quantitatively in Table~\ref{tab:mst_overlap}.


\section{Limitations and Broader Impacts}
\label{app:limit_impact}

\subsection{Limitations and Future Work}
\label{app:limitations}

Our work has several limitations that also point to promising directions for future research. First, we evaluate \sname only in specialized-domain semi-supervised vision-language adaptation settings, specifically remote sensing, fashion, science figures and comics, and on two downstream tasks: zero-shot classification and image-text retrieval. Although the results are encouraging in these settings, they do not yet establish that the method generalizes to broader multimodal adaptation scenarios. Extending the evaluation to more diverse domains and tasks would help clarify the scope and robustness of the proposed approach.
Second, our study is restricted to cross-modal semi-supervised alignment. While recent work has explored broader unimodal alignment perspectives, we do not examine whether the proposed topology-aware edge alignment transfers effectively to those settings. Investigating such extensions could reveal whether the structural bias introduced by \sname is useful beyond the current paired multimodal formulation.
Third, \sname relies on a lightweight topological construction defined on the 1-skeleton, using $H_0$-death and $H_1$-birth edge-level signatures without constructing higher-order simplices. This design keeps the method computationally practical, but it may miss richer higher-order topological structure in the representation space. A natural direction for future work is therefore to explore whether incorporating higher-order simplicial constructions can provide additional benefits while maintaining reasonable computational cost.
Finally, the benefits of combining $H_0$- and $H_1$-based signals are not uniform across tasks. While the joint formulation provides a favorable trade-off for zero-shot classification, its advantages are less consistent across retrieval settings. This suggests that more adaptive ways of weighting, selecting, or scheduling topological signals according to the task or data characteristics may further improve performance.

\subsection{Broader Impacts}
\label{app:impacts}

This work studies topology-aware representation alignment for semi-supervised vision-language
learning in specialized domains, with experiments on remote sensing and fashion datasets.
Potential positive impacts include improving adaptation of vision-language models when labeled
image-text pairs are scarce, which may support applications in domain-specific retrieval and
classification. At the same time, methods for stronger representation alignment may also be used
in sensitive settings, especially in remote sensing, where improved modeling could contribute to
surveillance-related or other privacy-sensitive downstream applications. More broadly, the method
may inherit biases or limitations present in the underlying pretrained model and benchmark datasets.
We therefore view this work as a methodological contribution and encourage careful evaluation and
domain-specific oversight before deployment in high-stakes settings.

\section{Experimental Details}
\label{app:exp_details}

\subsection{Dataset Details}
\label{sec:data_details}

\begin{figure}[!ht]
    \centering
    \includegraphics[width=0.8\linewidth]{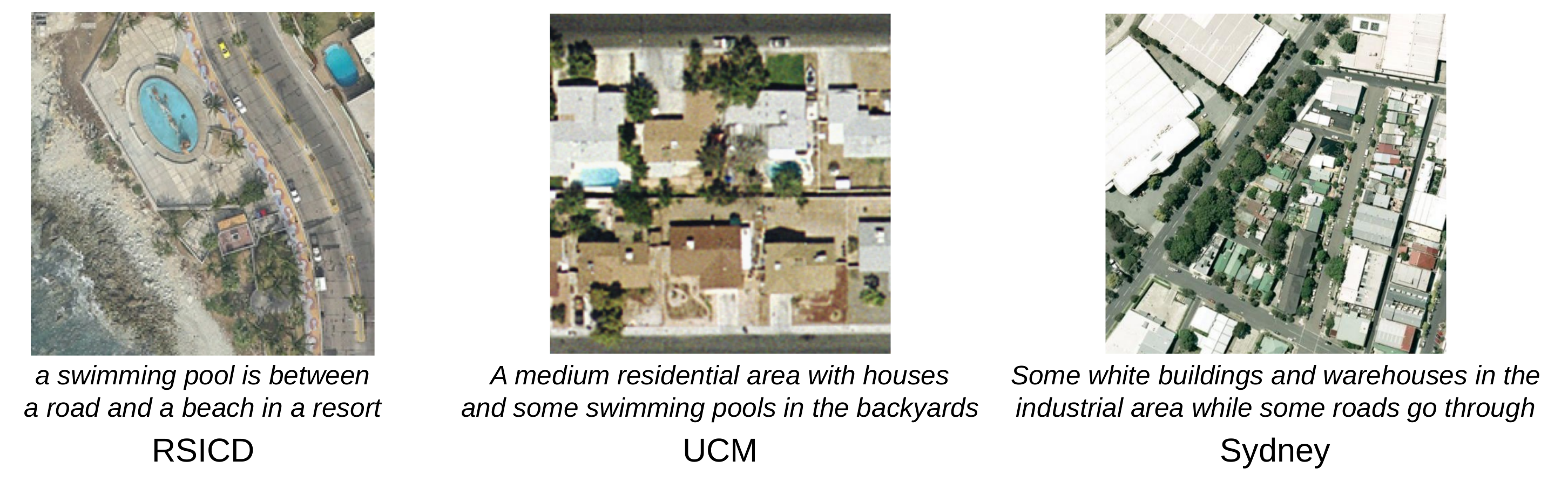}
    \caption{Examples of image-text pairs from the Remote sensing datasets.}
    \label{fig:dataex_rs}
\end{figure}

\paragraph{Remote sensing.}
Our training pipeline utilizes a combined collection of RSICD~\citep{lu2017exploring}, UCM~\citep{yang2010bag}, and Sydney~\citep{zhang2014saliency}, referred to as the RS-ALL suite. To implement our semi-supervised framework, a random 10\% subset of these image-text pairs is designated as labeled data ($L$), while the remaining 90\% serves as the unlabeled pool ($U$). To evaluate the model's robustness against distribution shifts, we introduce the RESISC45~\citep{cheng2017remote} dataset, which consists of 31,500 images across 45 scene categories, as an out-of-distribution unlabeled pool ($L \neq U$).
The numerical distribution of RS-ALL is detailed in Table~\ref{tab:stat_rs}, and visual examples are illustrated in Figure~\ref{fig:dataex_rs}. During training, if an image is associated with multiple textual descriptions, a single caption is selected stochastically at each epoch to ensure linguistic variety.

\begin{table}[!ht]
\centering\small
\caption{Statistics of image-text paired train datasets in the remote sensing domain.}
\label{tab:stat_rs}
\setlength{\tabcolsep}{12pt}
\begin{tabular}{lccc}
\toprule
Dataset & RSICD & UCM & Sydney \\
\cmidrule(lr){2-4}
\# of Pairs & 8,734 & 1,680 & 497 \\
\bottomrule
\end{tabular}
\end{table}

For zero-shot classification, we employ the validation sets of RSICD-CLS and UCM-CLS, which are the classification-oriented versions of RSICD~\citep{lu2017exploring} and UCM~\citep{yang2010bag}, respectively. To further examine the cross-domain generalization capability of our model, we incorporate three additional external benchmarks: WHU-RS19~\citep{xia2010structural}, RSSCN7~\citep{zou2015deep}, and AID~\citep{xia2017aid}. The numerical and categorical statistics for these evaluation sets are summarized in Table~\ref{tab:stat_rs_zeroshot}.

\begin{table}[!ht]
\centering\small
\caption{Statistics of zero-shot classification benchmarks in the remote sensing domain.}
\label{tab:stat_rs_zeroshot}
\setlength{\tabcolsep}{6pt}
\begin{tabular}{lccccc}
\toprule
Dataset & RSICD-CLS & UCM-CLS & WHU-RS19 & RSSCN7 & AID \\
\cmidrule(lr){2-6}
\# of Images & 1,094 & 2,100 & 1,005 & 2,800 & 10,000 \\
\# of Classes & 31 & 21 & 19 & 7 & 30 \\
\bottomrule
\end{tabular}
\end{table}

\begin{figure}[!ht]
    \centering
    \includegraphics[width=0.8\linewidth]{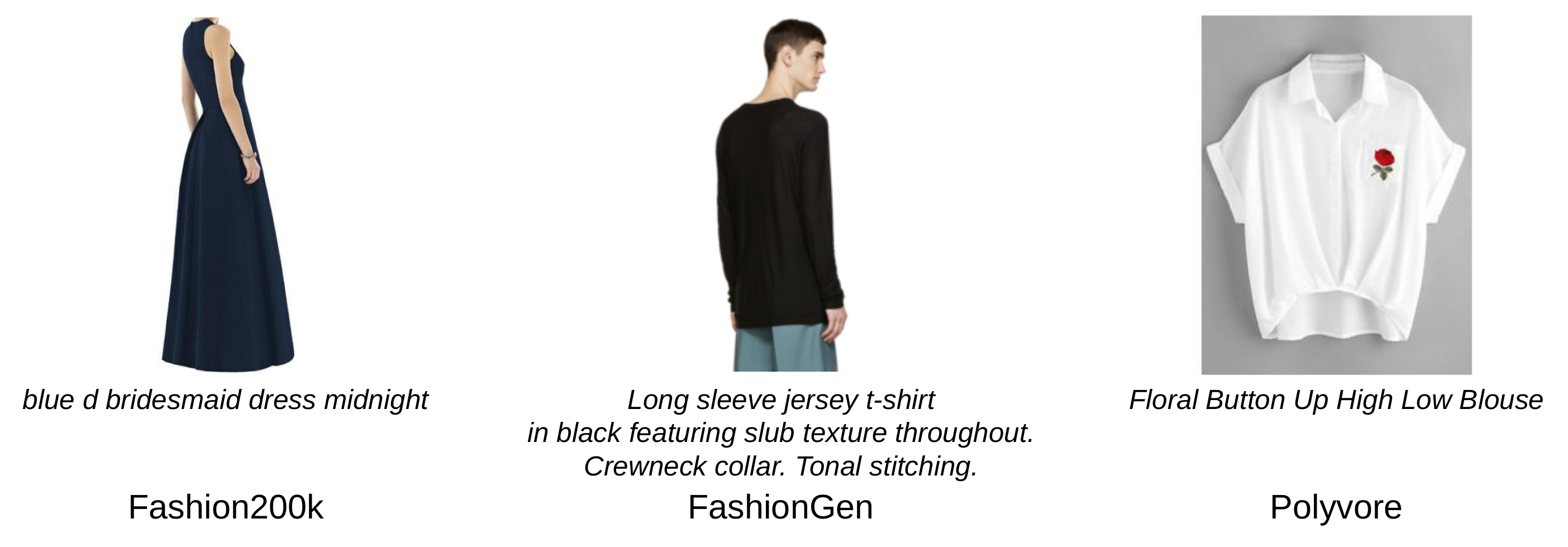}
    \caption{Examples of image-text pairs from the Fashion datasets.}
    \label{fig:dataex_fashion}
\end{figure}

\paragraph{Fashion.}
We aggregate Fashion200k~\citep{han2017automatic}, FashionGen~\citep{rostamzadeh2018fashion}, and Polyvore Outfits~\citep{vasileva2018learning} to form the primary training set for the fashion domain. To handle instances where a single item is shown from multiple visual perspectives, we stochastically select one image view per item during each training epoch. Regarding textual data, we merge product titles with their corresponding descriptions for the FashionGen and Polyvore subsets, while Fashion200k relies on its original sentence-based captions. Statistical details of these datasets are provided in Table~\ref{tab:stat_fashion}, with visual examples showcased in Figure~\ref{fig:dataex_fashion}.

\begin{table}[!ht]
\centering\small
\caption{Statistics of image-text paired datasets in the fashion domain.}
\label{tab:stat_fashion}
\setlength{\tabcolsep}{15pt}
\begin{tabular}{lccc}
\toprule
Dataset & Fashion200k & FashionGen & Polyvore Outfits \\
\cmidrule(lr){2-4}
\# of Pairs & 61,759 & 60,147 & 71,967 \\
\bottomrule
\end{tabular}
\end{table}

Zero-shot performance is assessed using the validation sets of the aforementioned fashion datasets, as outlined in Table~\ref{tab:stat_fashion_zeroshot}. For Fashion200k and FashionGen, we perform evaluations at both the super-class (coarse) and sub-class (fine-grained) levels. Category labels in the Polyvore dataset are standardized to match the super-class level of other benchmarks, ensuring a consistent evaluation scale across all fashion datasets.

\begin{table}[!ht]
\centering\small
\caption{Statistics of zero-shot classification benchmarks in the fashion domain.}
\label{tab:stat_fashion_zeroshot}
\setlength{\tabcolsep}{8pt}
\begin{tabular}{l cc cc c} 
\toprule
& \multicolumn{2}{c}{Fashion200k} & \multicolumn{2}{c}{FashionGen} & Polyvore Outfits \\
& Super-class & Sub-class & Super-class & Sub-class & Class \\
\cmidrule(lr){2-3} \cmidrule(lr){4-5} \cmidrule(lr){6-6}
\# of Images & \multicolumn{2}{c}{29,789} & \multicolumn{2}{c}{32,528} & 15,288 \\
\# of Classes & 5 & 31 & 48 & 121 & 11 \\
\bottomrule
\end{tabular}
\end{table}

\begin{figure}[!ht]
    \centering
    \includegraphics[width=0.5\linewidth]{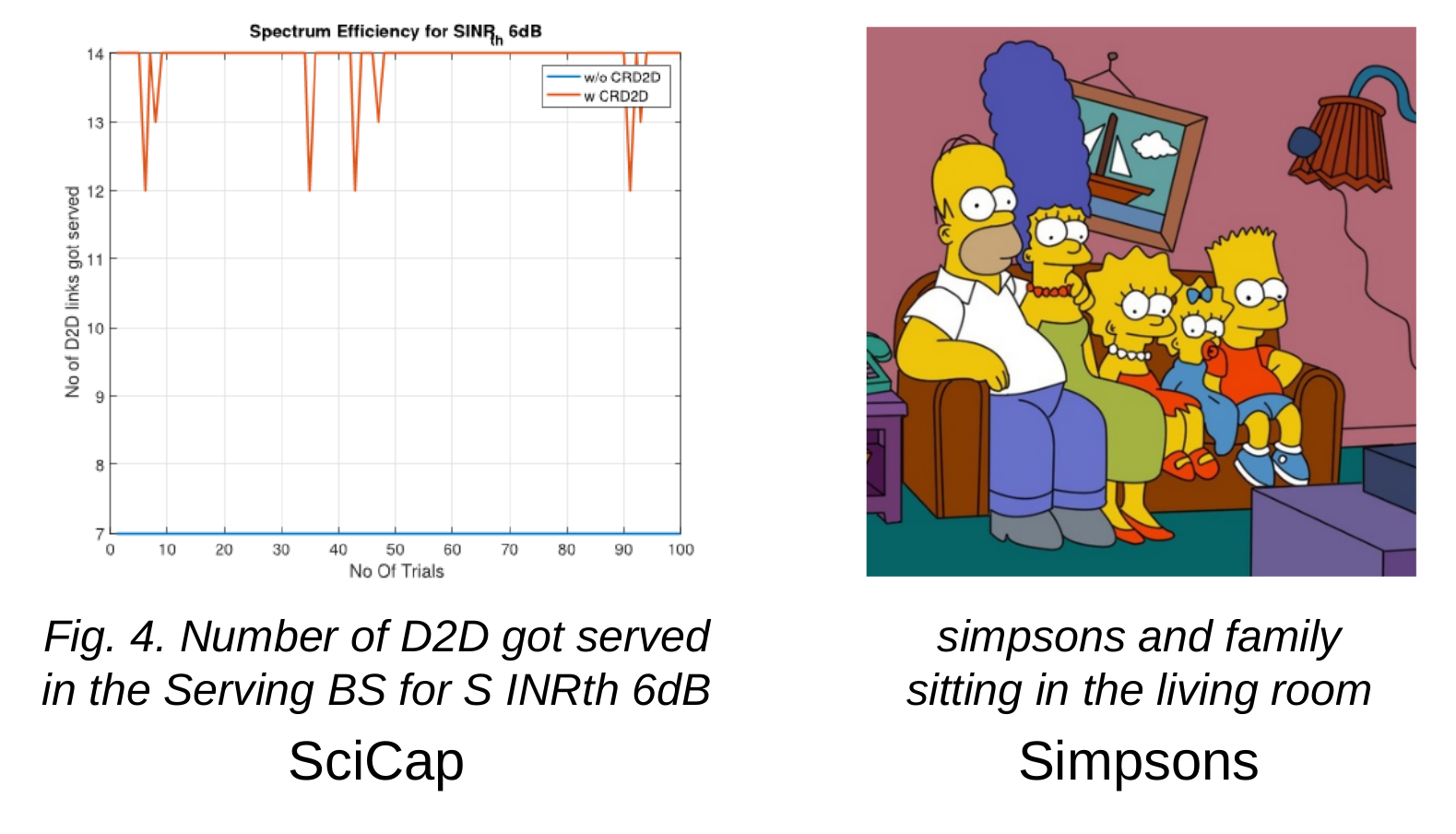}
    \caption{Examples of image-text pairs from the SciCap and Simpsons datasets.}
    \label{fig:dataex_more}
\end{figure}

\paragraph{SciCap.}
We incorporate the SciCap dataset, which comprises scientific figures such as charts, graphs, and mathematical notations. For our experiments, we utilize the SciCap-No-Subfig-Img subset, which excludes subfigures to ensure data consistency. This subset contains 106,834 image-text pairs. Sample images and their corresponding captions from this domain are presented in Figure~\ref{fig:dataex_more}.

\paragraph{Simpsons.}
For the Simpsons dataset, we employ 679 image-text pairs from the simpsons-blip-captions collection. Due to the limited scale of the labeled data, we integrate an additional image pool from the Simpsons Characters Data as an unlabeled reservoir. Samples from this domain are showcased in Figure~\ref{fig:dataex_more}.

\subsection{Implementation Details}
\label{app:implementation_details}

\paragraph{Setup.}
We adopt CLIP~\citep{radford2021learning} as the pretrained vision-language backbone for all experiments. 
By default, ViT-B/16 is used as the vision encoder and the experiments with other vision encoders are in Appendix~\ref{app:model_variants}. 
Unless otherwise noted, all experiments are performed on a single NVIDIA A100 GPU with a batch size of 128 for remote sensing and 256 for fashion datasets, where each mini-batch consists of an equal number of paired image-text samples and unpaired images.
As \sname is built on SemiCLIP~\citep{gan2025semi}, our training procedure follows the same two-stage paradigm and the model is trained for 25 epochs in stage 1 and 15 epochs in stage 2. 
We also follow SemiCLIP for the optimizer, cosine learning rate scheduler, learning rate, and the remaining training configurations.
The top-$k$ and top-$P$ are set to 4 and 30, respectively, following SemiCLIP.
The coefficient $c$ in the loss function $\mathcal{L}^{(s)}$ is set to $0.5$ for remote sensing and $0.1$  for fashion datasets.
The results of the hyperparameter search on the loss coefficients are shown in Appendix~\ref{app:loss_co}.
For the computation of PH, we follow the method proposed by ToMCLIP~\citep{you2026topological}.
We conduct experiments with two settings: \snameall (ours), which applies $\mathcal{L}_{\text{\sname}}$ to each batch without considering domain information, and \snamedom (ours), which applies it separately to each domain.
The $H_0$-death and $H_1$-birth edge sets are recomputed from the current embeddings at every mini-batch and treated as non-differentiable index selections. These indices determine which relations enter $\mathcal{L}_{\mathrm{ToMA}}$, while the directional alignment loss is computed from the current embeddings, allowing gradients to propagate normally through the encoders.

\paragraph{Baselines.}
We compare \sname against several competitive baselines. 
CLIP (original) refers to the pretrained model without further adaptation, and CLIP (fine-tuned) denotes the model fine-tuned using only labeled image-text pairs. 
We also consider CLIP (oracle), which is fine-tuned using both labeled and unlabeled image-text pairs with access to the ground-truth labels of the unlabeled data, serving as an upper-bound reference. 
In addition, we include Hard-PL~\citep{lee2013pseudo}, Soft-PL~\citep{assran2021semi}, and S-CLIP, all of which were introduced in S-CLIP~\citep{mo2023s}. 
SemiCLIP~\citep{gan2025semi} is also adopted as a baseline, since \sname is built on it.
The scores of CLIP, Hard-PL, Soft-PL and S-CLIP are imported from~\citep{gan2025semi}. 

The models are evaluated on zero-shot classification and image-text retrieval, where performance is measured by Top-1 classification accuracy ($\%$) and Recall@K (R@K), respectively. 
Results are reported as the mean and standard deviation over three runs with fixed random seeds $\{0,1,2\}$ for reproducibility. 
For fair comparison, SemiCLIP~\citep{gan2025semi} is retrained using the same set of random seeds.

\section{Preliminaries of Semi-Supervised Vision-Language Alignment}
\label{app:preliminary_semisup}

Semi-supervised vision-language alignment addresses the setting where paired image-text data $\mathcal{D}^l = \{(I_i, T_i)\}_{i=1}^N$ are scarce, while unlabeled images $\mathcal{D}^u = \{I_j\}_{j=1}^M$ are abundant with $M \gg N$. Its goal is to leverage unlabeled images to improve the alignment of the shared embedding space. SemiCLIP~\citep{gan2025semi} approaches this problem with a two-stage framework.

\textit{Stage 1. Semantic Concepts Mining.}
During supervised pre-training, semantic concepts are extracted as nouns from the captions. Their CLIP text embeddings, obtained via $f_T(\cdot)$, initialize a linear classifier $\psi$, which is trained to predict image-level concepts using the soft cross-entropy loss
\(
\mathcal{L}_{\text{SCM}} = - \sum_{i=1}^{N} \frac{y_{I_i}}{|y_{I_i}|} \log \psi(I^e_i)
\)
where $y_{I_i} \in \{0, 1\}^V$ denotes the multi-hot concept label. 
The training objective is $\mathcal{L}_{\text{Semi}}^{(1)} = \mathcal{L}_{\text{CLIP}} + \mathcal{L}_{\text{SCM}}$. The trained model then assigns pseudo-labels $\hat{y}_{I_j}$ to unlabeled images by selecting the top-$k$ confident concepts~\citep{sohn2020fixmatch}.

\textit{Stage 2. Semi-supervised Fine-tuning.}
In the second stage, the model is fine-tuned to improve alignment between unlabeled images and pseudo-labels. This is achieved with two losses. The first, concept-level semantic consistency, trains the model to preserve the pseudo-labels of strongly augmented unlabeled images:
\(
\mathcal{L}_{\text{SCM}}^u = - \sum_{j=1}^{|D^u|} \frac{\hat{y}_{I_j}}{|\hat{y}_{I_j}|} \log \psi(f_I(\Omega(I_j))).
\)
The second, caption-level trapezoidal consistency, constructs a surrogate caption $\hat{T}_j$ for each unlabeled image by concatenating its top-$k$ concepts, and then enforces structural consistency over $\tilde{\mathcal{D}} = \mathcal{D}^l \cup \mathcal{D}^p = \{(\tilde{I_i}, \tilde{T_i})\}_{i=1}^{N+M_P}$:
$$
\mathcal{L}_{\text{Trap}} = \underbrace{\mathcal{L}_{\text{CLIP}}}_{\text{upper base}} + \frac{1}{|\tilde{D}|^2} \sum_{i=1}^{|\tilde{D}|} \sum_{j=1}^{|\tilde{D}|} \left( \underbrace{(\langle \tilde{I}^e_i, \tilde{T}^e_j \rangle - \langle \tilde{I}^e_j, \tilde{T}^e_i \rangle)^2}_{\text{diagonals}} + \underbrace{(\langle \tilde{I}^e_i, \tilde{I}^e_j \rangle - \langle \tilde{T}^e_j, \tilde{T}^e_i \rangle)^2}_{\text{legs}} \right),
$$
where $\mathcal{D}^p \subset \mathcal{D}^u$.
The overall objective is $\mathcal{L}_{\text{Semi}}^{(2)} = \mathcal{L}_{\text{SCM}}^u + \mathcal{L}_{\text{Trap}}$.

\section{Reported and Retrained SemiCLIP Results}
\label{app:semiclip}

As noted in the main paper, we retrain SemiCLIP under the same evaluation pipeline and random seeds as our method, and use these retrained results for the main comparison. Our intention is not to replace the originally reported SemiCLIP results in~\citep{gan2025semi}, but to provide a controlled baseline under the same experimental protocol. Because the reported results and our method come from different training runs, direct comparison to the originally reported numbers may reflect not only method differences but also run-level variation. We therefore report both the published and retrained SemiCLIP scores in this appendix for transparency.

\begin{table*}[!ht]
\centering\small
\caption{Comparison between the originally reported SemiCLIP results~\citep{gan2025semi} and our retrained SemiCLIP results on zero-shot classification for remote sensing datasets. Parentheses denote performance gains over supervised CLIP, and green values indicate gains larger than one.}
\label{tab:rs-zeroshot-semi}
\begin{adjustbox}{width=\linewidth}
\begin{tabular}{@{}lccccccc@{}}
\toprule
Method & Data & RSICD-CLS & UCM-CLS & WHU-RS19 & RSSCN7 & AID\\
\midrule
CLIP (fine-tuned) & L &
75.7\stdv{2.2}\blankup & 80.1\stdv{6.8}\blankup & 92.2\stdv{1.5}\blankup & 70.7\stdv{5.5}\blankup & 79.8\stdv{3.8}\blankup\\
\midrule
Reported SemiCLIP~\citep{gan2025semi} & \multirow{2}{*}{L$=$U} &
{83.8}\stdv{1.0}\colorup{8.1} & {85.6}\stdv{3.4}\colorup{5.5} & {96.6}\stdv{1.2}\colorup{4.4} & {75.3}\stdv{1.9}\colorup{4.6} & {87.4}\stdv{0.5}\colorup{7.6}\\
Retrained SemiCLIP~\citep{gan2025semi} & &
81.4\stdv{1.1}\colorup{5.7} & 84.3\stdv{3.2}\colorup{4.2} & 95.6\stdv{0.3}\colorup{3.4} & {76.2}\stdv{2.7}\colorup{5.5} & 84.0\stdv{1.2}\colorup{4.2}\\
\midrule
Reported SemiCLIP~\citep{gan2025semi} & \multirow{2}{*}{L$\neq$U} &
{84.1}\stdv{0.5}\colorup{8.4} & {85.4}\stdv{2.8}\colorup{5.3} & {96.5}\stdv{1.7}\colorup{4.3} & 72.9\stdv{3.5}\colorup{2.2} & {86.2}\stdv{1.0}\colorup{6.4}\\
Retrained SemiCLIP~\citep{gan2025semi} & &
80.1\stdv{1.4}\colorup{4.4} & 82.8\stdv{1.0}\colorup{2.7} & 95.1\stdv{0.1}\colorup{2.9} & 75.0\stdv{0.8}\colorup{4.3} & 82.5\stdv{1.9}\colorup{2.7}\\
\bottomrule
\end{tabular}
\end{adjustbox}
\end{table*}

\begin{table*}[!ht]
\centering
\caption{Comparison between the originally reported SemiCLIP results~\citep{gan2025semi} and our retrained SemiCLIP results on image-text retrieval (R@5) for remote sensing datasets.}
\label{tab:rs-retrieval@r5-semi}
\begin{adjustbox}{width=0.9\linewidth}
\begin{tabular}{@{}lccccccc@{}}
\toprule
& & \multicolumn{3}{c}{Image$\to$Text R@5} & \multicolumn{3}{c}{Text$\to$Image R@5} \\
\cmidrule(lr){3-5}\cmidrule(lr){6-8}
Method & Data & RSICD & UCM & Sydney & RSICD & UCM & Sydney \\
\midrule
CLIP (fine-tuned) & L &
25.7\stdv{2.3} & 55.2\stdv{0.8} & 46.6\stdv{3.0} & 24.9\stdv{0.6} & 56.3\stdv{1.6} & 49.4\stdv{2.0} \\
\midrule
Reported SemiCLIP~\citep{gan2025semi} &  \multirow{2}{*}{L$=$U} &
{27.6}\stdv{1.6} & {59.2}\stdv{2.2} & {53.4}\stdv{4.6} & {26.4}\stdv{0.4} & {58.4}\stdv{2.4} & {58.0}\stdv{2.0} \\
Retrained SemiCLIP~\citep{gan2025semi} & &
27.9\stdv{0.7} & 57.3\stdv{2.6} & 52.3\stdv{4.3} & 25.8\stdv{0.2} & 59.4\stdv{1.2} & 58.6\stdv{2.4}\\
\midrule
Reported SemiCLIP~\citep{gan2025semi} & \multirow{2}{*}{L$\neq$U} &
{28.5}\stdv{0.2} & {59.5}\stdv{3.6} & {54.6}\stdv{3.6} & {26.3}\stdv{0.4} & {59.5}\stdv{1.3} & {58.6}\stdv{4.6} \\
Retrained SemiCLIP~\citep{gan2025semi} & &
26.1\stdv{0.9} & 57.9\stdv{1.0} & 54.6\stdv{0.8} & 25.1\stdv{0.2} & 59.4\stdv{1.5} & 59.2\stdv{1.6}\\
\bottomrule
\end{tabular}
\end{adjustbox}
\end{table*}

\begin{table*}[!ht]
\centering
\caption{Comparison between the originally reported SemiCLIP results~\citep{gan2025semi} and our retrained SemiCLIP results on zero-shot classification for fashion datasets. Parentheses denote performance gains over supervised CLIP, and green values indicate gains larger than one.}
\label{tab:fashion-zeroshot-semi}
\begin{adjustbox}{width=\linewidth}
\begin{tabular}{@{}lccccc@{}}
\toprule
& \multicolumn{2}{c}{Fashion200k} & \multicolumn{2}{c}{FashionGen} & Polyvore \\
\cmidrule(lr){2-3}\cmidrule(lr){4-5}\cmidrule(lr){6-6}
Method & Super-class & Sub-class & Super-class & Sub-class & Class \\
\midrule
CLIP (fine-tuned) &
79.0\stdv{3.5}\blankup & 35.1\stdv{0.7}\blankup & 35.4\stdv{8.1}\blankup & 24.5\stdv{2.4}\blankup & 60.4\stdv{2.3}\blankup \\
\midrule
Reported SemiCLIP~\citep{gan2025semi} &
{85.8}\stdv{0.4}\colorup{6.8} & {44.7}\stdv{1.0}\colorup{9.6} & \,\>{51.4}\stdv{0.6}\colorup{16.0} & \,\>{47.6}\stdv{1.7}\colorup{23.1} & {74.4}\stdv{0.5}\colorup{14.0} \\
Retrained SemiCLIP~\citep{gan2025semi} &
84.5\stdv{1.8}\colorup{5.5} & 43.7\stdv{1.2}\colorup{8.6} & 43.4\stdv{3.9}\colorup{8.0} & \,\>43.0\stdv{2.6}\colorup{18.5} & 73.0\stdv{2.1}\colorup{12.6}\\
\bottomrule
\end{tabular}
\end{adjustbox}
\end{table*}

\begin{table*}[!ht]
\centering
\caption{Comparison between the originally reported SemiCLIP results~\citep{gan2025semi} and our retrained SemiCLIP results on image-text retrieval (R@5) for fashion datasets.}
\label{tab:fashion-retrieval@r5-semi}
\begin{adjustbox}{width=\linewidth}
\begin{tabular}{@{}lcccccc@{}}
\toprule
& \multicolumn{3}{c}{Image$\to$Text R@5} & \multicolumn{3}{c}{Text$\to$Image R@5} \\
\cmidrule(lr){2-4}\cmidrule(lr){5-7}
Method & Fashion200k & FashionGen & Polyvore & Fashion200k & FashionGen & Polyvore \\
\midrule
CLIP (fine-tuned) &
13.7\stdv{0.4} & 32.1\stdv{0.2} & 16.3\stdv{0.5} & 13.5\stdv{0.2} & 31.9\stdv{0.2} & 16.2\stdv{0.3} \\
\midrule
Reported SemiCLIP~\citep{gan2025semi} &
{22.1}\stdv{0.4} & {48.4}\stdv{0.7} & {27.7}\stdv{0.3} & {22.5}\stdv{0.2} & {49.7}\stdv{0.3} & {27.6}\stdv{0.2} \\
Retrained SemiCLIP~\citep{gan2025semi} &
23.3\stdv{0.5} & 48.2\stdv{0.8} & 28.2\stdv{0.1} & {23.4}\stdv{0.2} & 49.5\stdv{0.5} & 28.6\stdv{0.1}\\
\bottomrule
\end{tabular}
\end{adjustbox}
\end{table*}

Tables~\ref{tab:rs-zeroshot-semi}-\ref{tab:fashion-retrieval@r5-semi} compare the originally reported SemiCLIP results and our retrained results on the remote sensing and fashion benchmarks. Overall, the retrained model exhibits the same qualitative behavior as the reported one, while showing some differences in absolute scores across settings. On remote sensing zero-shot classification (Table~\ref{tab:rs-zeroshot-semi}), the retrained results are generally lower on RSICD-CLS, UCM-CLS, WHU-RS19, and AID, but slightly higher on RSSCN7 under both L$=$U and L$\neq$U. On remote sensing retrieval (Table~\ref{tab:rs-retrieval@r5-semi}), the differences are smaller and more mixed, with several metrics remaining comparable and some text-to-image results on UCM and Sydney slightly improving after retraining.

A similar pattern appears on the fashion benchmarks. In zero-shot classification (Table~\ref{tab:fashion-zeroshot-semi}), the retrained SemiCLIP results are consistently below the reported scores, although the overall relative pattern across datasets is preserved. In image-text retrieval (Table~\ref{tab:fashion-retrieval@r5-semi}), however, the retrained model remains highly competitive and slightly exceeds the reported results on several metrics, including Fashion200k and Polyvore.

Taken together, these comparisons suggest that SemiCLIP is reasonably sensitive to the specific training run, especially for zero-shot classification. For this reason, we use the retrained SemiCLIP results in the main paper, as they provide a more directly matched and controlled baseline for evaluating \sname, while the originally reported results are included here to make the comparison fully transparent.

\section{Additional Experimental Results}
\label{app:add_exp}

\subsection{Image-text Retrieval Results on Fashion Datasets}
\label{app:fashion_retrival}

\begin{table*}[!ht]
\centering
\caption{%
Image-text retrieval results on fashion datasets using the same setup as in Table~\ref{tab:fashion-zeroshot}.
}\label{tab:fashion-retrieval@r5}
\vspace{-0.08in}
\begin{adjustbox}{width=0.88\linewidth}
\begin{tabular}{@{}lcccccc@{}}
\toprule
& \multicolumn{3}{c}{Image$\to$Text R@5} & \multicolumn{3}{c}{Text$\to$Image R@5} \\
\cmidrule(lr){2-4}\cmidrule(lr){5-7}
Method & Fashion200k & FashionGen & Polyvore & Fashion200k & FashionGen & Polyvore \\
\midrule
CLIP (original) &
\phantom{0}12.5\blankstdv & 23.5\blankstdv & 24.3\blankstdv & 12.3\blankstdv & 27.7\blankstdv & 27.4\blankstdv \\
\midrule
CLIP (fine-tuned) &
13.7\stdv{0.4} & 32.1\stdv{0.2} & 16.3\stdv{0.5} & 13.5\stdv{0.2} & 31.9\stdv{0.2} & 16.2\stdv{0.3} \\
CLIP (oracle) &
34.9\stdv{0.7} & 63.4\stdv{0.1} & 48.5\stdv{0.8} & 34.9\stdv{0.9} & 35.3\stdv{0.9} & 49.3\stdv{0.5} \\
\midrule
Hard-PL~\citep{lee2013pseudo} &
12.6\stdv{0.7} & 26.8\stdv{0.7} & 14.9\stdv{0.8} & 14.5\stdv{0.8} & 30.2\stdv{0.4} & 17.3\stdv{0.9} \\
Soft-PL~\citep{assran2021semi} &
15.2\stdv{0.8} & 35.0\stdv{0.3} & 18.8\stdv{0.3} & 15.9\stdv{0.7} & 37.4\stdv{0.7} & 21.9\stdv{0.2} \\
S-CLIP~\citep{mo2023s} &
16.9\stdv{0.4} & 37.9\stdv{0.4} & 22.6\stdv{0.4} & 17.4\stdv{0.3} & 40.9\stdv{0.4} & 24.2\stdv{0.5} \\
SemiCLIP~\citep{gan2025semi} &
23.3\stdv{0.5} & 48.2\stdv{0.8} & 28.2\stdv{0.1} & \textbf{23.4}\stdv{0.2} & 49.5\stdv{0.5} & 28.6\stdv{0.1}\\
\rowcolor{gray!15}
\snameall (ours) & 
23.5\stdv{0.2} & 48.6\stdv{0.3} & 28.5\stdv{0.3} & 23.2\stdv{0.3} & 49.5\stdv{0.1} & 28.5\stdv{0.3}\\
\rowcolor{gray!15}
\snamedom (ours) & 
\textbf{23.8}\stdv{0.4} & \textbf{48.7}\stdv{0.3} & \textbf{28.6}\stdv{0.1} & {23.3}\stdv{0.4} & \textbf{49.8}\stdv{0.3} & \textbf{28.7}\stdv{0.2}\\
\bottomrule
\end{tabular}
\end{adjustbox}
\end{table*}

Table~\ref{tab:fashion-retrieval@r5} reports the R@5 results on the
fashion datasets. Compared with SemiCLIP, \snamedom improves five of the six
retrieval metrics, including all three image-to-text settings, while remaining
within 0.1 percentage points on Fashion200k text-to-image retrieval. The
aggregate improvements are nevertheless modest, particularly compared with the
classification gains in Table~\ref{tab:fashion-zeroshot}. We therefore conduct a
query-level diagnostic analysis in Appendix~\ref{app:fashion_retrieval_analysis}
to examine how topology-aware alignment affects different retrieval queries and
semantic neighborhood structure.

\subsection{Diagnostic Analysis of Fashion Retrieval}
\label{app:fashion_retrieval_analysis}

Although the aggregate changes in exact-pair retrieval are modest, they may mask heterogeneous effects across queries and semantic neighborhood structure.
We therefore conduct a controlled query-level analysis on FashionGen to better understand how topology-aware alignment changes retrieval behavior. We use 7,519 products from the validation set, selecting one image-caption pair per product.
This controlled one-pair-per-product evaluation is used only for query-level diagnostics and is therefore not directly comparable to the standard FashionGen retrieval scores in Table~\ref{tab:fashion-retrieval@r5}.

\paragraph{Cross-modal topology alignment.}
We first examine whether \sname improves the agreement between the image and text
representation structures. We construct the $H_0$-death edge
graphs in batches of 256 over five shuffled passes. For each query, neighbor Jaccard measures the overlap between its image- and text-graph neighbors, while direction agreement measures the absolute cosine agreement between corresponding edge directions in the two modalities.

\begin{table}[!ht]
    \centering
    \caption{
    Cross-modal topology alignment on the controlled FashionGen validation set.
    Improvements are reported in percentage points (p.p.).
    }
    \label{tab:fashion_topology_alignment}
    \begin{tabular}{lccc}
        \toprule
        Metric & SemiCLIP & \sname & Improvements \\
        \midrule
        $H_0$ neighbor Jaccard
        & 22.57\% & 24.45\% & $+1.88$ p.p. \\
        $H_0$ direction agreement
        & 33.21\% & 42.77\% & $+9.56$ p.p. \\
        \bottomrule
    \end{tabular}
\end{table}

As shown in Table~\ref{tab:fashion_topology_alignment}, \sname improves both
neighborhood overlap and edge-direction agreement for the $H_0$-derived relations. Thus, the modest aggregate exact-pair retrieval gains occur despite clear improvements in the measured cross-modal topology alignment.

\paragraph{Behavior across query difficulty.}
We define query difficulty using the bidirectional retrieval margin of SemiCLIP. For each product, we compute the matched-pair similarity minus the highest non-matching similarity separately for image-to-text and text-to-image retrieval, and average the two margins. We then divide queries into equally
sized low- (hard), medium-, and high-margin (easy) thirds.
A smaller margin indicates that the ground-truth pair is less clearly separated from competing candidates. Table~\ref{tab:fashion_query_difficulty} reports the results for the hardest and easiest thirds.

\begin{table}[!ht]
    \centering
    \caption{
    Retrieval performance for the hardest and easiest FashionGen queries.
    Query difficulty is determined using the SemiCLIP positive-negative
    retrieval margin.
    }
    \label{tab:fashion_query_difficulty}
    \resizebox{0.9\linewidth}{!}{
    \begin{tabular}{llcccc}
        \toprule
        Difficulty & Direction
        & SemiCLIP (R@1) & \sname (R@1)
        & SemiCLIP (R@5) & \sname (R@5) \\
        \midrule
        Hard
        & I$\rightarrow$T
        & 0.12 & \textbf{3.95}
        & 8.14 & \textbf{19.94} \\
        Hard
        & T$\rightarrow$I
        & 0.12 & \textbf{3.55}
        & 7.66 & \textbf{20.10} \\
        \midrule
        Easy
        & I$\rightarrow$T
        & \textbf{77.45} & 64.57
        & \textbf{99.36} & 92.58 \\
        Easy
        & T$\rightarrow$I
        & \textbf{78.33} & 63.17
        & \textbf{99.32} & 91.06 \\
        \bottomrule
    \end{tabular}
    }
\end{table}

\sname substantially improves retrieval for the low-margin queries. In both retrieval directions, it improves the ground-truth rank for approximately 61\% of these queries. In contrast, performance decreases for some high-margin queries whose exact pairs are already well separated by SemiCLIP. This pattern suggests that \sname can improve cases that are poorly
resolved by the baseline while perturbing the fine-grained ordering of already well-separated pairs. These opposing effects partially cancel in aggregate Recall.

\paragraph{Semantic coherence of retrieved neighborhoods.}
Exact-pair Recall evaluates whether the single paired item is retrieved but does
not measure whether the remaining retrieved items are semantically related to
the query. We therefore additionally evaluate subcategory and coarse-category
P@5, excluding the exact paired item. We also report semantic nDCG@5, which
assigns graded relevance to the exact pair, items from the same subcategory, and
items from the same coarse category.
For semantic nDCG@5, relevance scores of 3, 2, 1, and 0 are assigned to the exact pair, an item from the same subcategory, an item from the same coarse category, and all remaining items, respectively.

\begin{table}[!ht]
    \centering
    \caption{
    Changes from SemiCLIP to \sname in exact-pair retrieval and semantic
    neighborhood coherence on the controlled FashionGen validation set.
    }
    \label{tab:fashion_semantic_coherence}
    \begin{tabular}{lcc}
        \toprule
        Metric
        & I$\rightarrow$T change
        & T$\rightarrow$I change \\
        \midrule
        Exact R@5
        & $+0.24$ p.p.
        & $-0.82$ p.p. \\
        Subcategory P@5
        & {$+1.29$ p.p.}
        & {$+1.27$ p.p.} \\
        Coarse-category P@5
        & {$+1.06$ p.p.}
        & {$+1.28$ p.p.} \\
        Semantic nDCG@5
        & $+0.86$ p.p.
        & $+0.39$ p.p. \\
        \bottomrule
    \end{tabular}
\end{table}

Table~\ref{tab:fashion_semantic_coherence} shows that \sname consistently
improves the semantic coherence of the retrieved neighborhoods, even when the
change in exact-pair Recall is small. Together with the query-difficulty
analysis, this result indicates that topology-aware alignment is particularly
helpful for ambiguous queries and neighborhood-level semantic organization.

Fashion retrieval requires precise instance-level ordering among many closely related products. A topology-aware structural objective can improve cross-modal relational agreement and semantic neighborhood coherence without necessarily preserving the exact ordering of pairs that are already well separated. Taken together, these results indicate that the effect of \sname on fashion retrieval is task dependent: it improves cross-modal structural agreement and the retrieval of low-margin queries, while its effect on aggregate exact-pair Recall remains modest.

\subsection{Additional Specialist-Domain Evaluation}
\label{app:more_datasets}

We further evaluate \sname on two additional specialist domains: science figures and comics. Specifically, we use the SciCap~\citep{hsu2021scicap} and Simpsons~\citep{adler2023simpsons} datasets. Each model is trained separately on the corresponding dataset and evaluated for image-text retrieval on its validation split.
Table~\ref{tab:other-retrieval} reports the results.

\begin{table*}[!ht]
\centering\small
\caption{%
Image-text retrieval results on SciCap (scientific figures) and Simpsons (comics) datasets. We train each model separately on each dataset and evaluate it on the corresponding validation set. Bold denotes the best result among semi-supervised methods for each metric.
}\label{tab:other-retrieval}
\begin{adjustbox}{width=\linewidth}
\begin{tabular}{@{}lcccccccc@{}}
\toprule
& \multicolumn{4}{c}{SciCap} & \multicolumn{4}{c}{Simpsons} \\
\cmidrule(lr){2-5}\cmidrule(lr){6-9}
\vspace{0.03in}
&
\multicolumn{2}{c}{Image$\to$Text} & \multicolumn{2}{c}{Text$\to$Image} &
\multicolumn{2}{c}{Image$\to$Text} & \multicolumn{2}{c}{Text$\to$Image} \\
Method & R@1 & R@5 & R@1 & R@5 & R@1 & R@5 & R@1 & R@5 \\
\midrule
CLIP (original) &
\phantom{0}8.3\blankstdv & 13.8\blankstdv & \phantom{0}8.8\blankstdv & 14.2\blankstdv &
13.2\blankstdv & 35.5\blankstdv & 10.5\blankstdv & 32.9\blankstdv \\
\midrule
CLIP (fine-tuned) &
11.0\stdv{0.6} & 19.0\stdv{0.8} & 11.1\stdv{0.9} & 18.1\stdv{0.0} &
19.3\stdv{2.0} & 48.2\stdv{6.0} & 14.5\stdv{0.0} & 45.6\stdv{2.0} \\
CLIP (oracle) &
14.8\stdv{0.9} & 25.5\stdv{1.4} & 14.8\stdv{0.5} & 25.7\stdv{0.3} &
28.3\stdv{3.7} & 52.3\stdv{5.6} & 22.3\stdv{3.1} & 55.8\stdv{4.1} \\
\midrule
Hard-PL~\citep{lee2013pseudo} &
11.2\stdv{0.2} & 18.9\stdv{0.2} & 12.2\stdv{0.1} & 20.1\stdv{0.1} &
16.7\stdv{2.7} & 42.1\stdv{5.3} & 15.4\stdv{4.2} & 43.9\stdv{3.3} \\
Soft-PL~\citep{assran2021semi} &
11.5\stdv{0.2} & 19.4\stdv{0.3} & 12.2\stdv{0.2} & 20.5\stdv{0.2} &
18.4\stdv{1.3} & 40.4\stdv{4.0} & 15.8\stdv{1.3} & 43.4\stdv{3.5} \\
S-CLIP~\citep{mo2023s} &
11.6\stdv{0.5} & 20.6\stdv{0.5} & 12.6\stdv{0.4} & 21.1\stdv{0.4} &
15.4\stdv{2.0} & 43.9\stdv{5.5} & 17.5\stdv{1.5} & 47.4\stdv{6.8} \\
SemiCLIP~\citep{gan2025semi} &
13.1\stdv{0.0} & 21.5\stdv{0.1} & 13.6\stdv{0.3} & 21.9\stdv{0.1} &
23.7\stdv{0.0} & 54.4\stdv{2.0} & \textbf{19.7}\stdv{1.3} & 54.4\stdv{3.3} \\
\rowcolor{gray!15}
\snameall (ours) & 
\textbf{13.2}\stdv{0.0} & \textbf{21.7}\stdv{0.1} & \textbf{13.7}\stdv{0.2} & \textbf{22.0}\stdv{0.2} & \textbf{27.2}\stdv{4.0} & \textbf{55.7}\stdv{1.5} & 18.9\stdv{3.3} & \textbf{55.7}\stdv{1.5}
\\
\bottomrule
\end{tabular}
\end{adjustbox}
\end{table*}

\paragraph{SciCap.}
On SciCap, \sname performs comparably to SemiCLIP across all retrieval metrics. It slightly improves image-to-text retrieval from 13.1 to 13.2 in R@1 and from 21.5 to 21.7 in R@5, and text-to-image retrieval from 13.6 to 13.7 in R@1 and from 21.9 to 22.0 in R@5. Although the gains are marginal, these results show that the topology-aware alignment signal does not degrade performance on the science-figure domain and remains stable beyond the main benchmarks.

\paragraph{Simpsons.}
On Simpsons, \sname yields more visible improvements for image-to-text retrieval, improving over SemiCLIP by 3.5 percentage points in R@1 and 1.3 percentage points in R@5. For text-to-image retrieval, \sname improves R@5 by 1.3 percentage points, while its R@1 is slightly lower than SemiCLIP. Overall, these results suggest that \sname can provide additional benefits on comics-style data, particularly for image-to-text retrieval, while maintaining comparable performance in the reverse retrieval direction.

\subsection{Seed-wise Experimental Results}
\label{app:exp_seed}

\begin{table*}[!ht]
\centering\small
\caption{Seed-wise averaged results over each dataset group. ZS denotes zero-shot classification accuracy, and I2T/T2I denote image-to-text and text-to-image retrieval performance, respectively.}
\label{tab:seed_wise_stability}
\resizebox{\textwidth}{!}{
\begin{tabular}{@{}llccccccccc@{}}
\toprule
\multirow{2}{*}{Dataset} & \multirow{2}{*}{Metric}
& \multicolumn{3}{c}{Seed 0}
& \multicolumn{3}{c}{Seed 1}
& \multicolumn{3}{c}{Seed 2} \\
\cmidrule(lr){3-5} \cmidrule(lr){6-8} \cmidrule(lr){9-11}
& & SemiCLIP & \snameall & \snamedom
& SemiCLIP & \snameall & \snamedom
& SemiCLIP & \snameall & \snamedom \\
\midrule
\multirow{3}{*}{RS (L$=$U)}
& ZS  & 82.5 & 85.9 & \textbf{87.7} & 84.4 & \textbf{84.6} & 83.3 & 85.9 & \textbf{87.8} & 86.9 \\
& I2T & 45.1 & 45.4 & \textbf{47.2} & 46.2 & 45.7 & \textbf{46.9} & 46.2 & \textbf{48.2} & 47.8 \\
& T2I & 46.9 & 46.0 & \textbf{51.9} & 47.2 & 49.0 & \textbf{51.1} & 49.7 & \textbf{49.8} & 49.7 \\
\midrule
\multirow{3}{*}{RS (L$\neq$U)}
& ZS  & 83.6 &\textbf{85.1} & 84.8 & 82.1 & 83.7 & \textbf{84.8} & 83.5 & 86.9 & \textbf{87.0} \\
& I2T & 45.7 & 46.9 & \textbf{47.3} & 45.8 & \textbf{47.6} & 47.1 & 47.2 & \textbf{48.7} & 47.7 \\
& T2I & 46.8 & 48.2 & \textbf{48.9} & 47.8 & 48.1 & \textbf{50.0} & 49.0 & 47.9 & \textbf{49.7} \\
\midrule
\multirow{3}{*}{Fashion}
& ZS  & 58.8 & 58.7 & \textbf{61.2} & 58.5 & \textbf{60.6} & 60.2 & 55.3 & \textbf{57.8} & \textbf{57.8} \\
& I2T & 33.3 & \textbf{33.8} & 33.7 & 33.2 & 33.4 & \textbf{33.7} & 33.3 & 33.4 & \textbf{33.7} \\
& T2I & 33.8 & \textbf{33.9} & 33.6 & 33.7 & 33.5 & \textbf{34.0} & 34.0 & 33.8 & \textbf{34.1} \\
\bottomrule
\end{tabular}
}
\end{table*}

Table~\ref{tab:seed_wise_stability} reports seed-wise averaged results over each dataset group. 
This analysis is intended to verify whether the improvements of our method are consistent across random seeds rather than driven by a single favorable run. 
Across the three dataset settings, three metrics, and three random seeds, \snamedom matches or improves upon SemiCLIP in 25 out of 27 seed-level aggregate comparisons, with strict improvements in 24 cases. 
The gains are particularly consistent in the remote sensing settings, including the distribution-shift case. 
On fashion datasets, the improvements are more modest, especially for retrieval, but the results remain comparable or slightly better in most seed-level comparisons. 
These results support the stability of the proposed topology-aware alignment signal.

\subsection{Different Neural Network Architectures}
\label{app:model_variants}

Our main experiments use ViT-B/16 as the default vision encoder. To examine whether the effect of \sname depends on this particular backbone choice, we additionally evaluate three different neural architectures on the remote sensing benchmarks: ResNet-50, ViT-B/32, and ViT-B/16. For each backbone, we follow the same semi-supervised training protocol as in Section~\ref{sec:rs} and report zero-shot classification and image-text retrieval results in Tables~\ref{tab:rs-model-zeroshot} and~\ref{tab:rs-model-retrieval}, respectively.

\begin{table*}[ht!]
\centering\small
\caption{%
Zero-shot classification results on remote sensing datasets using different neural architectures. Parentheses denote performance gains over supervised CLIP, and {\color{cadmiumgreen}green} values indicate gains larger than one. Bold denotes the best semi-supervised result in each setting.
}\label{tab:rs-model-zeroshot}
\begin{adjustbox}{width=\linewidth}
\begin{tabular}{@{}llcccccc@{}}
\toprule
Model & Method & RSICD-CLS & UCM-CLS & WHU-RS19 & RSSCN7 & AID\\
\midrule
\multirow{5}{*}{ResNet-50} & CLIP (original) &
45.3\blankstdv\blankup & 50.5\blankstdv\blankup & 65.5\blankstdv\blankup & 58.9\blankstdv\blankup & 47.8\blankstdv\blankup\\
& CLIP (fine-tuned) &
58.3\stdv{0.3}\blankup & 63.5\stdv{3.4}\blankup & 76.5\stdv{3.2}\blankup & 61.9\stdv{1.2}\blankup & 63.1\stdv{1.3}\blankup\\
& S-CLIP &
{66.9}\stdv{1.7}\colorup{8.6} & {66.7}\stdv{1.6}\colorup{3.2} & \,\>{86.9}\stdv{2.0}\colorup{10.4} & {66.2}\stdv{1.1}\colorup{4.3} & {73.0}\stdv{0.3}\colorup{9.9}\\
& SemiCLIP & 
\,\>69.2\stdv{1.8}\colorup{10.9} & \,\>73.9\stdv{0.9}\colorup{10.4} & 83.8\stdv{1.9}\colorup{7.3} & \textbf{68.2}\stdv{1.3}\colorup{6.4} & \,\>75.4\stdv{2.1}\colorup{12.3}
\\
& \sname (ours) & \,\>\textbf{72.0}\stdv{2.1}\colorup{13.7} & \,\>\textbf{79.5}\stdv{0.8}\colorup{16.0} & \,\>\textbf{89.7}\stdv{1.8}\colorup{13.2} & 67.5\stdv{0.9}\colorup{5.6} & \,\>\textbf{78.2}\stdv{2.4}\colorup{15.1}\\ 
\midrule
\multirow{5}{*}{ViT-B/32} & CLIP (original) &
56.2\blankstdv\blankup & 58.5\blankstdv\blankup & 76.3\blankstdv\blankup & 62.3\blankstdv\blankup & 55.6\blankstdv\blankup\\
& CLIP (fine-tuned) &
73.6\stdv{3.1}\blankup & 79.0\stdv{4.4}\blankup & 91.1\stdv{1.3}\blankup & 70.4\stdv{3.1}\blankup & 79.6\stdv{2.5}\blankup\\
& S-CLIP &
{78.5}\stdv{0.5}\colorup{4.9} & {77.4}\stdv{4.4}\down{1.6} & {94.9}\stdv{0.5}\colorup{3.8} & {71.1}\stdv{2.6}\up{0.7} & {84.2}\stdv{1.6}\colorup{4.6}\\
& SemiCLIP & 
80.7\stdv{1.4}\colorup{7.1} & 83.8\stdv{3.3}\colorup{4.8} & 94.6\stdv{0.8}\colorup{3.5} & \textbf{75.0}\stdv{1.8}\colorup{4.6} & 83.0\stdv{2.0}\colorup{3.4} \\
& \sname (ours) & 
\textbf{82.0}\stdv{1.1}\colorup{8.4} & \textbf{87.1}\stdv{3.3}\colorup{8.1} & \textbf{97.0}\stdv{0.4}\colorup{6.0} & 72.1\stdv{1.7}\colorup{1.7} & \textbf{84.9}\stdv{1.4}\colorup{5.3}\\
\midrule
\multirow{5}{*}{ViT-B/16} & CLIP (original) &
59.2\blankstdv\blankup & 60.2\blankstdv\blankup & 81.2\blankstdv\blankup & 69.0\blankstdv\blankup & 59.6\blankstdv\blankup\\
& CLIP (fine-tuned) &
75.7\stdv{2.2}\blankup & 80.1\stdv{6.8}\blankup & 92.2\stdv{1.5}\blankup & 70.7\stdv{5.5}\blankup & 79.8\stdv{3.8}\blankup\\
& S-CLIP &
81.4\stdv{1.8}\colorup{5.7} & 81.3\stdv{3.4}\colorup{1.2} & 95.9\stdv{1.8}\colorup{3.7} & 75.1\stdv{2.0}\colorup{4.4} & 86.4\stdv{1.7}\colorup{6.6}\\
& SemiCLIP & 
81.4\stdv{1.1}\colorup{5.7} & 84.3\stdv{3.2}\colorup{4.2} & 95.6\stdv{0.3}\colorup{3.4} & \textbf{76.2}\stdv{2.7}\colorup{5.5} & 84.0\stdv{1.2}\colorup{4.2}
\\
& \sname (ours) & \textbf{84.8}\stdv{1.3}\colorup{9.1} & \textbf{86.1}\stdv{3.4}\colorup{6.0} & \textbf{97.9}\stdv{0.4}\colorup{5.7} & 74.2\stdv{1.7}\colorup{3.5} & \textbf{87.3}\stdv{2.2}\colorup{7.5}\\ 
\bottomrule
\end{tabular}
\end{adjustbox}
\end{table*}

\begin{table*}[ht!]
\centering\small
\caption{%
Image-text retrieval results on remote sensing datasets using different neural architectures. Bold denotes the best semi-supervised result in each setting.
}\label{tab:rs-model-retrieval}
\begin{adjustbox}{width=0.9\linewidth}
\begin{tabular}{@{}llcccccc@{}}
\toprule
& & \multicolumn{3}{c}{Image$\to$Text R@5} & \multicolumn{3}{c}{Text$\to$Image R@5} \\
\cmidrule(lr){3-5}\cmidrule(lr){6-8}
Model & Method & RSICD & UCM & Sydney & RSICD & UCM & Sydney \\
\midrule
\multirow{5}{*}{ResNet-50} & CLIP (original) &
9.4\blankstdv & 34.3\blankstdv & 36.2\blankstdv & 10.1\blankstdv & 24.8\blankstdv & 51.7\blankstdv\\
& CLIP (fine-tuned) &
15.4\stdv{1.7} & 41.3\stdv{1.8} & 47.1\stdv{6.5} & 15.1\stdv{1.0} & 40.9\stdv{1.6} & 56.1\stdv{2.4} \\
& S-CLIP &
18.4\stdv{0.6} & 45.7\stdv{1.4} & 50.0\stdv{3.0} & 16.8\stdv{1.2} & 43.5\stdv{1.5} & 55.1\stdv{2.0} \\
& SemiCLIP & 
20.0\stdv{0.9} & 45.2\stdv{3.1} & \textbf{49.4}\stdv{3.2} & 17.4\stdv{0.5} & 45.4\stdv{3.5} & 54.6\stdv{3.5}
\\
& \sname (ours) & \textbf{21.1}\stdv{1.1} & \textbf{47.5}\stdv{1.1} & 47.1\stdv{2.1} & \textbf{18.6}\stdv{0.5} & \textbf{47.1}\stdv{1.0} & \textbf{56.9}\stdv{4.2}\\ 
\midrule
\multirow{5}{*}{ViT-B/32} & CLIP (original) &
12.1\blankstdv & 37.6\blankstdv & 41.4\blankstdv & 12.9\blankstdv & 35.2\blankstdv & 37.9\blankstdv\\
& CLIP (fine-tuned) &
24.5\stdv{0.6} & 56.3\stdv{1.5} & 47.1\stdv{5.0} & 24.6\stdv{1.0} & 57.0\stdv{1.1} & 50.0\stdv{4.6} \\
& S-CLIP &
25.0\stdv{0.8} & 55.7\stdv{2.5} & 43.1\stdv{3.0} & 23.9\stdv{0.5} & 59.2\stdv{1.8} & 51.7\stdv{1.7} \\
& SemiCLIP & 
26.3\stdv{1.6} & \textbf{60.5}\stdv{2.3} & \textbf{52.3}\stdv{3.5} & 24.0\stdv{1.4} & 60.5\stdv{1.4} & 55.8\stdv{2.1}
\\
& \sname (ours) & \textbf{27.8}\stdv{0.7} & 60.2\stdv{1.5} & 50.6\stdv{3.2} & \textbf{24.8}\stdv{0.5} & \textbf{61.3}\stdv{0.8} & \textbf{56.3}\stdv{2.9}\\ 
\midrule
\multirow{5}{*}{ViT-B/16} & CLIP (original) &
\phantom{0}12.6\blankstdv & 46.7\blankstdv & 44.8\blankstdv & 13.9\blankstdv & 39.5\blankstdv & 44.8\blankstdv \\
& CLIP (fine-tuned) &
25.7\stdv{2.3} & 55.2\stdv{0.8} & 46.6\stdv{3.0} & 24.9\stdv{0.6} & 56.3\stdv{1.6} & 49.4\stdv{2.0}\\
& S-CLIP &
27.5\stdv{1.1} & 57.0\stdv{1.5} & 51.1\stdv{4.3} & 25.6\stdv{0.6} & 57.3\stdv{4.1} & 50.6\stdv{2.6}\\
& SemiCLIP & 
27.9\stdv{0.7} & 57.3\stdv{2.6} & \textbf{52.3}\stdv{4.3} & 25.8\stdv{0.2} & 59.4\stdv{1.2} & \textbf{58.6}\stdv{2.4}
\\
& \sname (ours) & \textbf{29.0}\stdv{1.5} & \textbf{61.4}\stdv{4.6} & 48.9\stdv{2.1} & \textbf{27.6}\stdv{1.2} & \textbf{60.8}\stdv{3.0} & 56.3\stdv{3.5}\\ 
\bottomrule
\end{tabular}
\end{adjustbox}
\end{table*}

Overall, the results show that \sname generalizes well across different backbone architectures. In zero-shot classification (Table~\ref{tab:rs-model-zeroshot}), \sname achieves the best semi-supervised performance in most settings and consistently improves over supervised CLIP by clear margins. These gains are particularly pronounced on RSICD-CLS, UCM-CLS, WHU-RS19, and AID, indicating that the proposed topology-aware alignment is not tied to a specific encoder family. While the improvement on RSSCN7 is relatively smaller and the ViT-B/16 backbone remains slightly below SemiCLIP on this dataset, the overall trend remains consistently favorable to \sname.

A similar pattern is observed for image-text retrieval (Table~\ref{tab:rs-model-retrieval}). Across different backbones, \sname improves over SemiCLIP on most retrieval settings, with especially consistent gains on RSICD and UCM for both image-to-text and text-to-image retrieval. The results on Sydney are comparatively mixed, suggesting that the benefit of topology-aware alignment may vary depending on the dataset characteristics and retrieval direction. Nevertheless, the overall results confirm that \sname provides a complementary structural signal beyond the baseline semi-supervised objective.

These observations suggest that the effectiveness of \sname does not rely on the default ViT-B/16 configuration. Instead, the proposed topology-aware regularization transfers across both convolutional and transformer-based encoders, supporting its usefulness as a backbone-agnostic approach for semi-supervised vision-language adaptation.

\subsection{Loss Coefficient and Homology-Dimension Ablations}
\label{app:loss_co}

We conduct ablation studies to analyze the role of the topology-aware alignment in 
\begin{equation}
\mathcal{L}^{(s)}
=
\mathcal{L}_{\text{Semi}}^{(s)}
+
\frac{c}{2}
\left(
\mathcal{L}_{\text{\sname}}^{\text{0-death}}
+
\mathcal{L}_{\text{\sname}}^{\text{1-birth}}
\right).
\label{eq:tomar_stage}
\end{equation}
Specifically, we vary the coefficient $c \in \{0.1, 0.5, 1, 2\}$ and compare three variants: using only the $H_0$-death term (+ $(c/2) \mathcal{L}_{\text{\sname}}^{\text{0-death}}$), using only the $H_1$-birth term (+ $(c/2) \mathcal{L}_{\text{\sname}}^{\text{1-birth}}$), and using both terms together (+ $c/2 (\mathcal{L}_{\text{\sname}}^{\text{0-death}}+\mathcal{L}_{\text{\sname}}^{\text{1-birth}})$). 
This experiment is designed to examine two questions: how sensitive \sname is to the weighting coefficient $c$, and whether the 1-dimensional topological signal provides useful information beyond the 0-dimensional one. 
Notably, our formulation uses only the birth edges of $H_1$ under the 1-skeleton filtration, which allows us to incorporate 1-dimensional information in a lightweight form rather than relying on full higher-dimensional simplicial computations. 

\begin{table*}[!ht]
\centering\small
\caption{
Ablation on the loss coefficient $c$ and homology components on remote sensing dataset. We compare \sname with only $H_0$-death, only $H_1$-birth, and both jointly under different values of $c$.
}
\label{tab:coefficient-rsicd-cls}
\begin{adjustbox}{width=\linewidth}
\begin{tabular}{@{}lcccccccccccc@{}}
\toprule
& \multicolumn{4}{c}{Zero-Shot} & \multicolumn{4}{c}{Image$\to$Text} & \multicolumn{4}{c}{Text$\to$Image} \\
\cmidrule(lr){2-5}\cmidrule(lr){6-9}\cmidrule(lr){10-13}
 & $c=0.1$ & $c=0.5$ & $c=1$ & $c=2$ & $c=0.1$ & $c=0.5$ & $c=1$ & $c=2$ & $c=0.1$ & $c=0.5$ & $c=1$ & $c=2$ \\
\midrule
CLIP (fine-tuned) & \multicolumn{4}{c}{79.7} & \multicolumn{4}{c}{42.5} & \multicolumn{4}{c}{43.5}
\\
%
\midrule
SemiCLIP~\citep{gan2025semi}  & \multicolumn{4}{c}{84.3} & \multicolumn{4}{c}{45.8} & \multicolumn{4}{c}{47.9} \\
+ $(c/2) \mathcal{L}_{\mathrm{\sname}}^{\text{0-death}}$ &  84.7 & 85.0 & 85.5 & 85.8 & 46.7 & 47.8 & 48.7 & 46.5 & 49.7 & 49.9 & 49.4 & 47.7 
\\
+ $(c/2) \mathcal{L}_{\mathrm{\sname}}^{\text{1-birth}}$ & 85.5 & 86.3 & 85.8 & 85.7 & 46.0 & 47.3 & 47.8 & 45.6 & 48.8 & 49.1 & 50.0 & 48.7
\\
+ $c/2 (\mathcal{L}_{\mathrm{\sname}}^{\text{0-death}}+\mathcal{L}_{\mathrm{\sname}}^{\text{1-birth}})$ & 84.6 & 86.0 & 85.9 & 85.4 & 46.8 & 46.2 & 46.9 & 46.1 & 49.6 & 48.7 & 50.1 & 47.9 
\\
%
%
\bottomrule
\end{tabular}
\end{adjustbox}
\end{table*}

As shown in Table~\ref{tab:coefficient-rsicd-cls}, \sname is reasonably robust to the choice of the loss coefficient $c$, especially in the range $c \in \{0.1, 0.5, 1\}$. Across all three variants, adding topology-aware alignment consistently improves the zero-shot average over SemiCLIP, confirming that both $H_0$- and $H_1$-based signals provide useful structural supervision. Among the single-component variants, the $H_1$-birth-only term yields the strongest zero-shot result, reaching 86.3 at $c=0.5$, which suggests that even the lightweight 1-dimensional signal carries informative structure beyond the connectivity information captured by $H_0$. For retrieval, the pattern is slightly different: the $H_0$-death-only variant gives the best image-to-text score (48.7 at $c=1$), while the joint variant achieves the best text-to-image score (50.1 at $c=1$). These results indicate that the two homology components play complementary roles, with $H_1$ being particularly helpful for classification and $H_0$ remaining beneficial for retrieval-oriented local structure.

At the same time, the results also show that excessively large topology weights can be harmful. In particular, when $c=2$, performance tends to drop for retrieval across several variants, suggesting that overly strong topological regularization may begin to interfere with the original semi-supervised alignment objective. This trend supports the use of a moderate coefficient in practice. 

\begin{table}[!ht]
\centering\small
\caption{Results for different weights for $\mathcal{L}^{0\text{-death}}_{\text{\sname}}$ and $\mathcal{L}^{1\text{-birth}}_{\text{\sname}}$.}
\label{tab:coefficient-rsicd-cls2}
\small
\setlength{\tabcolsep}{4pt}
\begin{adjustbox}{width=0.8\linewidth}
\begin{tabular}{@{}ccccccccccccc@{}}
\toprule
& \multicolumn{4}{c}{Zero-Shot} & \multicolumn{4}{c}{Image$\to$Text} & \multicolumn{4}{c}{Text$\to$Image} \\
\cmidrule(lr){2-5}\cmidrule(lr){6-9}\cmidrule(lr){10-13}
\diagbox{$c_2$}{$c_1$} & $0.1$ & $0.5$ & $1$ & $2$ & $0.1$ & $0.5$ & $1$ & $2$ & $0.1$ & $0.5$ & $1$ & $2$ \\
\midrule
$0.5$ & 84.9 & 85.7 & 85.8 & 85.5 & 46.5 & 45.5 & 46.4 & 47.3 & 49.6 & 49.5 & 49.7 & 48.3 \\
$1$ & 84.6 & 86.0 & 85.9 & 85.4 & 46.8 & 46.2 & 46.9 & 46.1 & 49.6 & 48.7 & 50.1 & 47.9  \\
$2$ & 84.7 & 85.8 & 85.4 & 84.0 & 47.8 & 45.6 & 45.7 & 44.3 & 49.2 & 49.1 & 47.3 & 46.4  \\
\bottomrule
\end{tabular}
\end{adjustbox}
\end{table}

Table~\ref{tab:coefficient-rsicd-cls2} further analyzes the relative contribution of the two topological terms. In this experiment, we use
\[
\mathcal{L}^{(s)}
=
\mathcal{L}_{\text{Semi}}^{(s)}
+
\frac{c_1}{2}
\left(
\mathcal{L}_{\text{\sname}}^{\text{0-death}}
+
c_2 \mathcal{L}_{\text{\sname}}^{\text{1-birth}}
\right).
\]
Here, $c_1$ controls the overall magnitude of the topology-aware regularizer, while $c_2$ reweights the $H_1$-birth term relative to the $H_0$-death term. Overall, the results suggest that performance is more sensitive to the global scale $c_1$ than to the relative weighting $c_2$. Moderate values of $c_1$ yield the most stable results, whereas larger values often degrade retrieval performance. The best zero-shot classification result is obtained at $(c_1,c_2)=(0.5,1)$, and the best text-to-image retrieval result at $(1,1)$, indicating that comparable weighting of the two components already works well in practice. Although image-to-text retrieval shows a slight preference for a larger $H_1$ weight, no single asymmetric configuration consistently outperforms the balanced setting across all metrics. Taken together, these results suggest that the $H_1$-birth term contributes useful complementary information, but that the benefit of \sname mainly comes from combining both terms under a moderate overall regularization strength.

\subsection{Results for R@1}

We additionally report image-text retrieval results in terms of Recall@1 (R@1), which provides a stricter evaluation than Recall@5 (R@5). Tables~\ref{tab:rs-retrieval@r1} and~\ref{tab:fashion-retrieval@r1} summarize the results on remote sensing and fashion datasets, respectively.

\begin{table*}[!ht]
\centering
\caption{%
Image-text retrieval results of R@1 on remote sensing datasets using the same setup as in Table~\ref{tab:rs-zeroshot}. Bold denotes the best results within the same setups.
}\label{tab:rs-retrieval@r1}
\begin{adjustbox}{width=0.9\linewidth}
\begin{tabular}{@{}lccccccc@{}}
\toprule
& & \multicolumn{3}{c}{Image$\to$Text R@1} & \multicolumn{3}{c}{Text$\to$Image R@1} \\
\cmidrule(lr){3-5}\cmidrule(lr){6-8}
Method & Data & RSICD & UCM & Sydney & RSICD & UCM & Sydney \\
\midrule
CLIP (original) & - &
\phantom{0}4.5\blankstdv & 17.1\blankstdv & 10.3\blankstdv & 3.6\blankstdv & 9.0\blankstdv & 12.1\blankstdv \\
\midrule
CLIP (fine-tuned) & L &
10.6\stdv{1.2} & 18.9\stdv{2.7} & 20.7\stdv{0.0} & 7.3\stdv{0.7} & 14.9\stdv{0.5} & 12.6\stdv{1.0} \\
\midrule
Hard-PL~\citep{lee2013pseudo} & \multirow{6}{*}{L$=$U} &
11.4\stdv{1.7} & 17.0\stdv{2.9} & 17.9\stdv{2.8} & 6.8\stdv{0.5} & 14.1\stdv{2.3} & \textbf{20.7}\stdv{3.0} \\
Soft-PL~\citep{assran2021semi} & &
11.6\stdv{1.2} & 19.2\stdv{1.1} & 20.7\stdv{3.0} & 7.1\stdv{0.4} & 13.7\stdv{0.6} & 20.3\stdv{5.6} \\
S-CLIP~\citep{mo2023s} & &
11.4\stdv{0.3} & 19.5\stdv{1.3} & 17.8\stdv{2.6} & 7.1\stdv{0.8} & 14.0\stdv{1.4} & 16.7\stdv{1.0} \\
SemiCLIP~\citep{gan2025semi} & &
11.0\stdv{0.7} & 19.4\stdv{0.6} & 18.4\stdv{4.5} & 6.8\stdv{0.5} & 15.2\stdv{0.8} & 18.4\stdv{0.8}\\
\rowcolor{gray!15}
\snameall (ours) & &  
\textbf{12.4}\stdv{1.0} & 19.4\stdv{2.8} & 20.7\stdv{1.4} & \textbf{8.1}\stdv{0.7} & \textbf{17.1}\stdv{0.8} & 20.1\stdv{1.6}\\
\rowcolor{gray!15}
\snamedom (ours) & &  
12.3\stdv{0.8} & \textbf{20.0}\stdv{2.8} & \textbf{21.3}\stdv{3.5} & 7.8\stdv{0.2} & 16.2\stdv{0.7} & \textbf{20.7}\stdv{2.8}\\
\midrule
Hard-PL~\citep{lee2013pseudo} & \multirow{6}{*}{L$\neq$U} &
11.1\stdv{1.5} & 17.1\stdv{2.4} & 19.0\stdv{3.4} & 6.1\stdv{0.2} & 16.0\stdv{1.5} & 16.7\stdv{5.3} \\
Soft-PL~\citep{assran2021semi} & &
11.2\stdv{0.4} & 16.0\stdv{1.5} & 16.7\stdv{3.6} & 6.5\stdv{0.2} & 15.1\stdv{2.2} & 18.4\stdv{1.0} \\
S-CLIP~\citep{mo2023s} & &
10.9\stdv{0.6} & 18.9\stdv{1.5} & \textbf{20.4}\stdv{2.2} & 7.2\stdv{0.6} & \textbf{16.5}\stdv{1.2} & 14.9\stdv{2.6} \\
SemiCLIP~\citep{gan2025semi} & &
10.5\stdv{0.7} & 20.6\stdv{2.0} & 20.1\stdv{0.8} & 7.2\stdv{0.7} & 15.7\stdv{0.8} & 20.7\stdv{0.0}\\
\rowcolor{gray!15}
\snameall (ours) & &  
13.2\stdv{0.9} & \textbf{21.1}\stdv{2.1} & 18.4\stdv{3.5} & 8.0\stdv{0.6} & 15.4\stdv{1.8} & 21.8\stdv{2.9}\\
\rowcolor{gray!15}
\snamedom (ours) & &  
\textbf{13.3}\stdv{0.5} & 19.7\stdv{1.6} & 19.5\stdv{1.6} & \textbf{8.1}\stdv{0.2} & 14.8\stdv{2.2} & \textbf{24.7}\stdv{3.2}\\
\bottomrule
\end{tabular}
\end{adjustbox}
\end{table*}

\begin{table*}[!ht]
\centering
\caption{%
Image-text retrieval results of R@1 on fashion datasets using the same setup as in Table~\ref{tab:fashion-zeroshot}. Bolds denotes the best results within the same setups.
}\label{tab:fashion-retrieval@r1}
\begin{adjustbox}{width=\linewidth}
\begin{tabular}{@{}lcccccc@{}}
\toprule
& \multicolumn{3}{c}{Image$\to$Text R@1} & \multicolumn{3}{c}{Text$\to$Image R@1} \\
\cmidrule(lr){2-4}\cmidrule(lr){5-7}
Method & Fashion200k & FashionGen & Polyvore & Fashion200k & FashionGen & Polyvore \\
\midrule
CLIP (original) &
\phantom{0}4.7\blankstdv & 10.9\blankstdv & 11.4\blankstdv & 4.5\blankstdv & 12.5\blankstdv & 13.0\blankstdv \\
\midrule
CLIP (fine-tuned) &
4.8\stdv{0.3} & 12.9\stdv{0.4} & 6.4\stdv{0.2} & 4.2\stdv{0.0} & 13.0\stdv{0.0} & 6.2\stdv{0.1} \\
\midrule
Hard-PL~\citep{lee2013pseudo} &
4.8\stdv{0.2} & 11.1\stdv{0.2} & 6.5\stdv{0.3} & 5.1\stdv{0.5} & 13.7\stdv{0.4} & 7.8\stdv{0.3} \\
Soft-PL~\citep{assran2021semi} &
5.6\stdv{0.3} & 14.7\stdv{0.4} & 8.2\stdv{0.2} & 5.3\stdv{0.4} & 15.4\stdv{0.5} & 9.8\stdv{0.0} \\
S-CLIP~\citep{mo2023s} &
6.6\stdv{0.3} & 16.7\stdv{0.4} & 10.4\stdv{0.2} & 6.1\stdv{0.2} & 18.9\stdv{0.2} & 10.9\stdv{0.4} \\
SemiCLIP~\citep{gan2025semi} &
\textbf{10.1}\stdv{0.2} & 23.8\stdv{0.2} & 14.0\stdv{0.1} & \textbf{9.1}\stdv{0.1} & 24.8\stdv{0.2} & 13.6\stdv{0.2}\\
\rowcolor{gray!15}
\snameall (ours) & 
10.0\stdv{0.2} & 24.2\stdv{0.2} & \textbf{14.2}\stdv{0.3} & 9.0\stdv{0.1} & 25.0\stdv{0.1} & 13.8\stdv{0.1}\\
\rowcolor{gray!15}
\snamedom (ours) & 
9.9\stdv{0.2} & \textbf{24.5}\stdv{0.1} & 14.1\stdv{0.1} & 8.9\stdv{0.1} & \textbf{25.1}\stdv{0.3} & \textbf{13.8}\stdv{0.1}\\
\bottomrule
\end{tabular}
\end{adjustbox}
\end{table*}

Overall, the R@1 results are consistent with the R@5 results. On remote sensing datasets, \sname and \sname-domain generally improve over SemiCLIP, with particularly clear gains on RSICD under both L$=$U and L$\neq$U settings. The two variants also achieve the best results in several settings on UCM and Sydney, showing that topology-aware alignment improves not only coarse retrieval performance but also top-ranked retrieval precision.

On fashion datasets, the improvements are more selective. \sname and \sname-domain outperform SemiCLIP on FashionGen and Polyvore in several retrieval settings, while SemiCLIP remains slightly stronger on Fashion200k. Overall, these R@1 results support the same conclusion as the R@5 evaluation: \sname provides a complementary structural signal that is particularly effective on remote sensing datasets and on more heterogeneous fashion datasets.

\subsection{Comparison with Topology-Free Graph Baselines}
\label{app:graph_baselines}

To separate the effect of introducing an additional relational
alignment objective from the effect of topology-informed edge
construction, we compare ToMA with two topology-free graph baselines:
a random graph and a $k$-nearest-neighbor ($k$-NN) graph.
The random baseline connects randomly selected pairs, whereas the
$k$-NN baseline constructs neighborhoods according to the current
representation geometry. To control for the number of aligned
relations, we approximately match the number of graph edges to that
of the $H_0$-death edges: for a batch of $N=128$ samples, the
$H_0$ minimum spanning tree contains $N-1=127$ edges, while the
random baseline samples 128 edges and the $k$-NN baseline uses
$k=1$, yielding one nearest-neighbor relation per sample.
These baselines replace the homology-based edge construction of ToMA
while retaining the same encoders, directional alignment objective,
training protocol, and evaluation settings. We evaluate both the
standard and domain-partitioned variants, denoted by the suffix ``-d.''

\begin{figure}[!ht]
    \centering
    \includegraphics[width=0.99\linewidth]{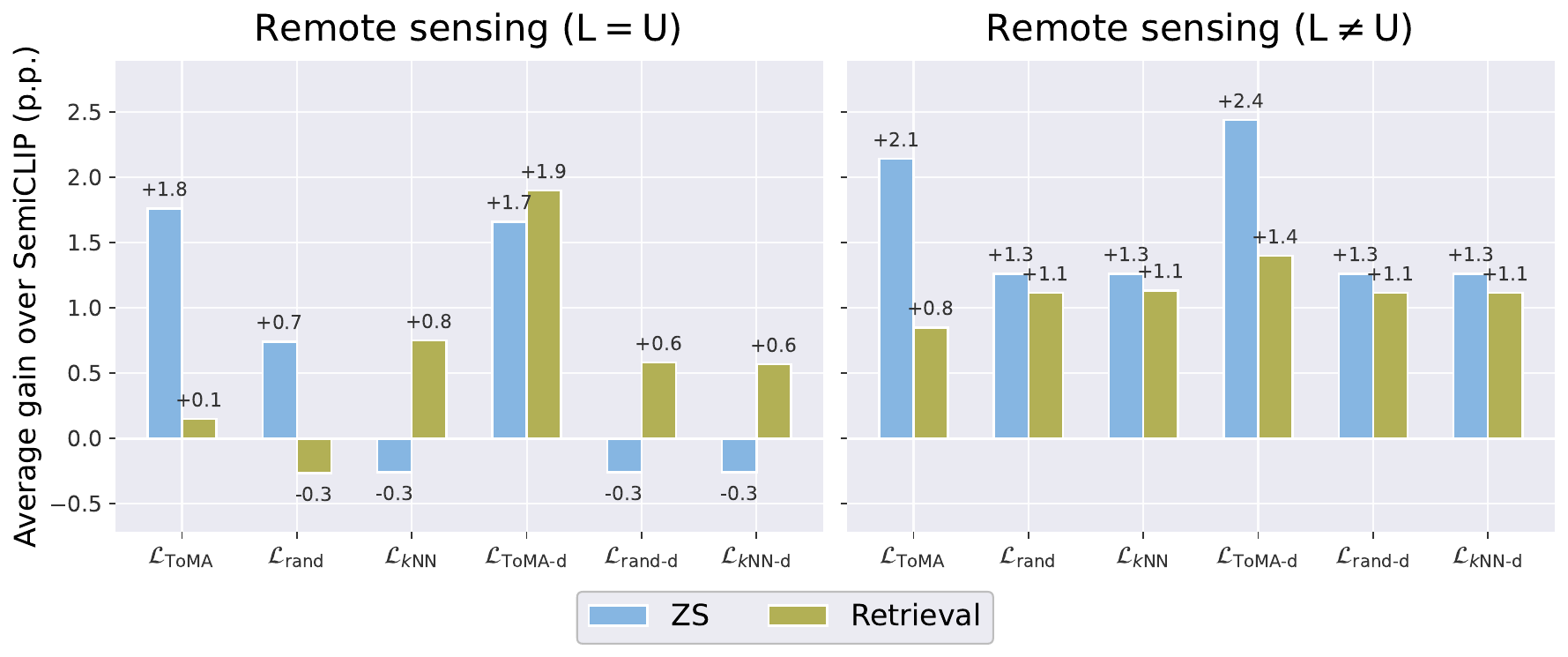}
    \caption{
    Comparison with topology-free graph baselines.
    Bars denote average gains over SemiCLIP for zero-shot classification
    (ZS) and image-text retrieval under L$=$U and L$\neq$U.
    Random and $k$-NN replace the homology-based edge construction with
    randomly sampled and nearest-neighbor graphs, respectively.
    The suffix ``-d'' denotes the corresponding domain-partitioned variant.
    }
    \label{fig:graph_baselines}
\end{figure}

Figure~\ref{fig:graph_baselines} reports the average gains over SemiCLIP
for zero-shot classification and image-text retrieval. When the labeled
and unlabeled data are drawn from the same domains (L$=$U), ToMA improves
zero-shot classification by $1.8$ percentage points, compared with
$0.7$ for the random graph and $-0.3$ for the $k$-NN graph. The
domain-partitioned ToMA variant achieves a gain of $1.7$ percentage
points, whereas both domain-partitioned graph baselines yield a decrease
of $0.3$ percentage points. For retrieval, the standard $k$-NN baseline
achieves a larger average gain than standard ToMA
($+0.8$ versus $+0.1$ percentage points). In contrast,
ToMA-d provides the largest retrieval improvement, reaching
$+1.9$ percentage points, compared with approximately $+0.6$ for the
domain-partitioned random and $k$-NN baselines.
For the L$=$U zero-shot comparison, we additionally perform a paired
analysis between ToMA and random edge alignment over the three matched
seeds. The macro-averaged ToMA$-$random difference across the five
evaluation datasets is $+1.11$ percentage points
(95\% CI $[-2.01,\,4.24]$, paired $t$-test $p=0.265$).
Thus, the observed advantage is positive for all three matched seeds
but is not statistically significant under the current three-seed evaluation.

Under the more challenging L$\neq$U setting, ToMA and ToMA-d improve
zero-shot classification by $2.1$ and $2.4$ percentage points,
respectively, exceeding the $1.3$ point gains obtained by the
corresponding random and $k$-NN baselines. Although their underlying
scores differ slightly, the random and $k$-NN controls yield nearly
identical aggregate gains at the reported precision. For retrieval, the
topology-free baselines remain competitive with standard ToMA:
the random and $k$-NN graphs improve the average score by approximately
$1.1$ percentage points, whereas ToMA yields a gain of $0.8$ percentage
points. Nevertheless, ToMA-d again obtains the strongest
domain-partitioned result, with an improvement of $1.4$ percentage
points.

Overall, the results show that generic graph regularization can itself
benefit representation alignment, particularly for retrieval, and that
the topology-aware construction is not uniformly superior under every
evaluation setting. However, ToMA provides consistently stronger
zero-shot classification performance, while its domain-partitioned
variant achieves the largest retrieval gains in both domain settings.
These findings suggest that the gains of ToMA are not attributable
solely to the use of an arbitrary graph regularizer, while domain-wise
topology construction can provide additional benefits for retrieval.

Together with the homology-dimension analysis in
Figure~\ref{fig:tda_ablation-2}, these results further suggest that the
benefit of ToMA does not simply arise from aligning a larger number of
pairwise relations. Under the 1-skeleton filtration, the $H_1$-birth
term already contains the dense set of all non-MST edges, whereas the
$H_0$-death term contains only the $N-1$ MST edges; nevertheless, the
$H_0$-only domain variant outperforms the $H_1$-only variant, and their
combination achieves the strongest performance. Since the two terms are
normalized separately before being combined, this pattern highlights
the importance of topology-induced edge decomposition and the separate
treatment of connectivity and cycle-closing relations, rather than
merely generic graph regularization or dense pairwise alignment.

\subsection{Results for Smaller Batch Size}
\label{app:bs64}

We additionally evaluate \sname with a smaller batch size of 64 on the remote sensing benchmarks. Since topology-aware alignment is computed within each mini-batch, this experiment tests whether the benefit of \sname remains effective even when fewer samples are available to estimate the batch-level structure.

\begin{table*}[ht!]
\centering\small
\caption{%
Zero-shot classification results on remote sensing datasets for batch size 64. Parentheses denote performance gains over supervised CLIP, and {\color{cadmiumgreen}green} values indicate gains larger than one. Bold denotes the best semi-supervised result in each setting.
}\label{tab:rs-bs}
\begin{adjustbox}{width=\linewidth}
\begin{tabular}{@{}lcccccc@{}}
\toprule
Method & RSICD-CLS & UCM-CLS & WHU-RS19 & RSSCN7 & AID\\
\midrule
CLIP (original) &
59.2\blankstdv\blankup & 60.2\blankstdv\blankup & 81.2\blankstdv\blankup & 69.0\blankstdv\blankup & 59.6\blankstdv\blankup\\
\midrule
CLIP (fine-tuned) &
75.3\stdv{2.7}\blankup & 77.2\stdv{7.6}\blankup & 92.3\stdv{2.9}\blankup & 72.4\stdv{3.2}\blankup & 79.7\stdv{3.5}\blankup\\
\midrule
Hard-PL &
{78.5}\stdv{3.3}\colorup{3.2} & {70.1}\stdv{4.7}\down{7.1} & \,\>{91.6}\stdv{0.4}\down{0.7} & \,\>{62.4}\stdv{4.8}\down{10.0} & {82.2}\stdv{2.9}\colorup{2.5}\\
Soft-PL &
{77.8}\stdv{3.4}\colorup{2.5} & {71.2}\stdv{3.7}\down{6.0} & \,\>{94.3}\stdv{1.0}\colorup{2.0} & {64.9}\stdv{5.5}\down{7.5} & {82.6}\stdv{1.6}\colorup{2.9}\\
S-CLIP &
{74.2}\stdv{1.6}\down{1.1} & {69.7}\stdv{3.0}\down{7.5} & \,\>{93.3}\stdv{1.5}\colorup{1.0} & {71.3}\stdv{1.7}\down{1.1} & {79.8}\stdv{1.6}\up{0.1}\\
SemiCLIP & 
83.8\stdv{0.4}\colorup{8.5} & \textbf{85.5}\stdv{5.0}\colorup{8.3} & \,\>94.9\stdv{0.8}\colorup{2.6} & 71.6\stdv{5.0}\down{0.8} & 87.0\stdv{1.1}\colorup{7.3}\\
\rowcolor{gray!15}
\sname (ours) & \,\>\textbf{85.3}\stdv{2.0}\colorup{10.0} & 84.3\stdv{5.0}\colorup{7.1} & \,\>\textbf{97.5}\stdv{0.6}\colorup{5.2} & \textbf{75.7}\stdv{2.9}\colorup{3.3} & \textbf{88.5}\stdv{3.2}\colorup{8.8}
\\ 
\bottomrule
\end{tabular}
\end{adjustbox}
\end{table*}

As shown in Table~\ref{tab:rs-bs}, \sname consistently achieves the best semi-supervised performance across four datasets. Compared with SemiCLIP, it improves zero-shot classification on RSICD-CLS, WHU-RS19, RSSCN7, and AID, while remaining only slightly below SemiCLIP on UCM-CLS. In particular, \sname reaches 85.3 on RSICD-CLS, 97.5 on WHU-RS19, 75.7 on RSSCN7, and 88.5 on AID, showing clear gains over both supervised CLIP and the semi-supervised baselines.


These results show that \sname remains effective with a batch size of 64, 
indicating that it does not require the default batch size of 128 to provide 
a useful structural signal. We next examine a broader range of batch sizes 
to characterize the batch-size sensitivity of \sname more systematically.

\subsection{Batch-Size Sensitivity}
\label{app:bs}

\begin{figure}[!ht]
    \centering
    \includegraphics[width=0.99\linewidth]{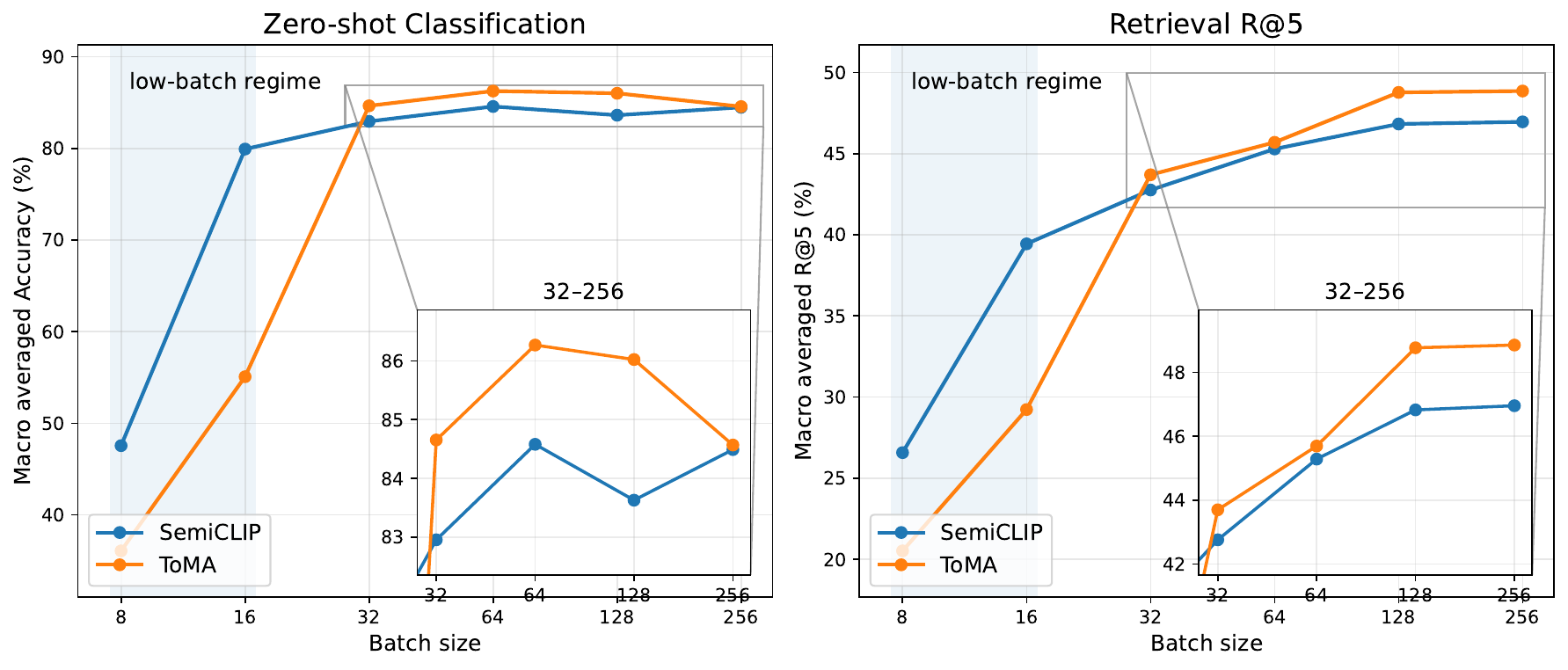}
    \caption{Effect of batch size on remote sensing datasets (L$=$U). We report macro-averaged zero-shot classification accuracy across five benchmarks and macro-averaged retrieval R@5 across three datasets and both retrieval directions. The main panels show the full range of batch sizes from 8 to 256, and the shaded regions indicate the low-batch regime. The insets enlarge the range from 32 to 256. Both methods degrade at very small batch sizes, whereas \sname matches or outperforms SemiCLIP consistently once the batch size reaches 32.}
    \label{fig:batch_size}
\end{figure}

To further characterize the effect of batch size, we evaluate SemiCLIP and 
\sname using batch sizes of 8, 16, 32, 64, 128, and 256. Figure~\ref{fig:batch_size}
reports macro-averaged performance on the remote sensing datasets (L$=$U). For 
zero-shot classification, we average the accuracy across the five evaluation 
datasets. For retrieval, we average R@5 across the three datasets and both 
retrieval directions. The main panels show the full batch-size range, while 
the insets provide an enlarged view of the results between 32 and 256.

Both methods deteriorate when the batch size becomes very small, indicating 
that the semi-supervised alignment objectives generally benefit from a 
sufficient number of samples within each mini-batch. The degradation is more 
pronounced for \sname at batch sizes of 8 and 16. Since \sname estimates 
topologically salient relationships within each mini-batch, an extremely small 
batch provides only a limited sample of the representation manifold. The 
resulting connectivity and cycle structure can therefore become more sensitive 
to the particular composition of the batch, reducing the reliability of the 
topological alignment signal.

Once the batch size reaches 32, \sname recovers sharply and matches or 
outperforms SemiCLIP across all evaluated batch sizes. Between batch sizes 32 
and 256, \sname improves the macro-averaged classification accuracy by 
approximately 0.1--2.4 percentage points and retrieval R@5 by approximately 
0.4--1.9 percentage points. The classification performance remains broadly 
stable throughout this range, while retrieval performance continues to improve 
up to batch sizes of 128 and 256. These results indicate that \sname does not 
require a very large batch, but it benefits from having enough samples to 
estimate meaningful batch-level topology. 
Notably, this transition at a batch size of 32 is consistent with the 
batch-size sensitivity reported by the TopKD paper~\citep{kim2024topological}, 
who found that topology-informed knowledge distillation became consistently 
beneficial once the batch size exceeded 16, as a sufficient number of points 
was required to characterize the underlying manifold topology reliably.

\subsection{Stability of MST Edges under Batch Perturbations}
\label{app:mst_stability}

Because \sname extracts topologically salient edges within each mini-batch, 
we examine whether these edges capture reproducible representation structure 
or are dominated by the particular composition of a batch. We conduct a batch 
perturbation analysis on RS-ALL using a batch size of 128. For each original 
batch, we randomly replace 13 samples, corresponding to approximately 10\% of 
the batch, while keeping the remaining image-text pairs fixed. We then 
recompute the image and text MSTs for the perturbed batch. The reported results 
are averaged over 500 batch perturbations for each of three training seeds.

Edges incident to a replaced sample necessarily disappear from the perturbed 
batch and therefore cannot be used to assess structural stability. We thus 
measure the conditional retention rate only over original MST edges whose two 
endpoints remain in the perturbed batch. Let \(E\) and \(\widetilde{E}\) denote 
the MST edge sets obtained from the original and perturbed batches, respectively, 
and let \(E_{\mathrm{surv}}\) denote the edges whose end points (samples) are retained after perturbation. 
We define edge retention as

\begin{equation}
\operatorname{Ret}(E,\widetilde{E})
=
\frac{
\left|
E \cap \widetilde{E}
\cap
E_{\mathrm{surv}}
\right|
}{
\left|
E
\cap
E_{\mathrm{surv}}
\right|
}.
\label{eq:edge_retention}
\end{equation}
We compute this quantity separately for the image MST, the text MST, and the 
edges shared by both modalities.

\begin{table}[!ht]
\centering
\caption{
Conditional retention rates of MST edges under batch perturbations on RS-ALL.
For each batch of 128 samples, 13 samples are randomly replaced, and edge
retention is evaluated only among original edges whose two endpoints remain in
the perturbed batch. Results are averaged over 500 perturbations for each of
three training seeds.
}
\label{tab:mst_stability}
\resizebox{0.7\linewidth}{!}{
\begin{tabular}{lccc}
\toprule
Method
& Image MST
& Text MST
& Cross-modal common edges \\
\midrule
CLIP
& 91.31\%
& 91.15\%
& 89.70\% \\

SemiCLIP
& 91.42\%
& 91.36\%
& 88.78\% \\

ToMA
& 91.53\%
& 91.40\%
& 88.74\% \\

ToMA-domain
& 91.47\%
& 91.42\%
& 88.77\% \\
\bottomrule
\end{tabular}
}
\end{table}

As shown in Table~\ref{tab:mst_stability}, the unimodal MSTs exhibit high 
retention across all methods. Image and text MST retention remains approximately 
91\%, while the retention of cross-modal common edges remains approximately 
89\%. In particular, SemiCLIP and \sname achieve nearly identical retention rates 
for cross-modal common edges, with 88.78\% and 88.74\%, respectively. These 
results indicate that \sname does not artificially increase the generic stability 
of an MST relative to the underlying baseline.

The main effect of \sname instead lies in the amount of cross-modal structure 
shared before perturbation. \sname increases the average number of image-text 
common MST edges from 40.32 for SemiCLIP to 43.50, while preserving essentially 
the same retention rate under batch perturbation. \sname produces a larger set of 
image-text connectivity relations that remains reproducible under nearby 
perturbations of the mini-batch.

\subsection{Domain Partitioning as a Control}
\label{app:domain_partition_control}

To disentangle the effect of domain partitioning from that of the \sname objective, we introduce a domain-partitioned SemiCLIP control. ToMA-domain assumes that the source-domain identity of each unlabeled sample is available as metadata. This setting is applicable when samples are collected from known datasets, acquisition sites, sensors, or data processing pipelines; standard ToMA does not require such information.

In our original implementation, both \sname and \sname-domain use the same domain-agnostic SemiCLIP objective. The only difference lies in the construction of the \sname loss. \sname identifies and aligns topological relations over the full mini-batch, whereas \sname-domain partitions the mini-batch by source domain before constructing and applying the \sname loss. Therefore, the comparison between \sname and \sname-domain isolates the effect of domain-wise topology construction.

We additionally evaluate \textbf{SemiCLIP-domain}, which applies the consistency losses of SemiCLIP separately within each source-domain subset but does not use the \sname objective. The remaining training components, including supervised contrastive learning and pseudo-label generation, remain unchanged.

\begin{table}[!ht]
    \centering
    \caption{
    Zero-shot classification results for the domain-partitioned controls.
    The average is the macro-average across the five evaluation datasets.
    }
    \label{tab:domain_partition_control}
    \resizebox{\linewidth}{!}{
    \begin{tabular}{lcccccc}
        \toprule
        Method
        & RSICD-CLS
        & UCM-CLS
        & WHU-RS19
        & RSSCN7
        & AID
        & Average \\
        \midrule
        SemiCLIP
        & 81.4
        & 84.3
        & 95.6
        & \textbf{76.2}
        & 84.0
        & 84.30 \\
        SemiCLIP-domain
        & 81.9
        & 85.3
        & 95.6
        & 75.5
        & 84.1
        & 84.48 \\
        \sname
        & \textbf{84.8}
        & 86.1
        & \textbf{97.9}
        & 74.2
        & \textbf{87.3}
        & \textbf{86.06} \\
        \sname-domain
        & 83.8
        & \textbf{87.5}
        & 97.8
        & 73.9
        & 86.8
        & 85.96 \\
        \bottomrule
    \end{tabular}
    }
\end{table}

As shown in Table~\ref{tab:domain_partition_control}, SemiCLIP-domain improves the average accuracy only marginally, from 84.30 to 84.48, and its effects vary across datasets. Likewise, \sname-domain does not improve the aggregate performance over standard \sname, achieving 85.96 compared with 86.06.

In contrast, both \sname variants substantially outperform their corresponding SemiCLIP baselines on average. \sname improves over SemiCLIP by 1.76 percentage points, while \sname-domain improves over SemiCLIP-domain by 1.48 percentage points. These results indicate that the improvement of \sname-domain cannot be explained by domain partitioning alone.

We therefore regard domain-wise topology construction as an optional, dataset-dependent extension. It can redistribute performance across individual datasets, such as improving UCM-CLS, but does not provide a consistent aggregate advantage. Overall, the results are more consistent with the \sname objective, rather than domain separation itself, accounting for the principal improvement.

\subsection{Different Labeled Image-Text Pair Ratios}
\label{app:ssl_ratio}

We further evaluate \sname under different labeled image-text pair ratios on the remote sensing datasets. In addition to the default 10\% setting, we consider 20\% and 30\% labeled pairs, while treating the remaining samples as unlabeled. Table~\ref{tab:rs-ratio} reports the zero-shot classification results.

\begin{table*}[ht!]
\centering\small
\caption{%
Zero-shot classification results on remote sensing datasets using different image-text pair ratios. Parentheses indicate the performance gap from the supervised CLIP, and bold denotes the best results within each pair ratio.}\label{tab:rs-ratio}
\begin{adjustbox}{width=\linewidth}
\begin{tabular}{@{}lccccccc@{}}
\toprule
Method & Ratio & RSICD-CLS & UCM-CLS & WHU-RS19 & RSSCN7 & AID\\
\midrule
CLIP (original) & 0\% & 
59.2\blankstdv\blankup & 60.2\blankstdv\blankup & 81.2\blankstdv\blankup & 69.0\blankstdv\blankup & 59.6\blankstdv\blankup\\
\midrule
CLIP (fine-tuned) & \multirow{3}{*}{10\%} & 
75.7\stdv{2.2}\blankup & 80.1\stdv{6.8}\blankup & 92.2\stdv{1.5}\blankup & 70.7\stdv{5.5}\blankup & 79.8\stdv{3.8}\blankup\\
SemiCLIP & &
81.4\stdv{1.1}\colorup{5.7} & 84.3\stdv{3.2}\colorup{4.2} & 95.6\stdv{0.3}\colorup{3.4} & \textbf{76.2}\stdv{2.7}\colorup{5.5} & 84.0\stdv{1.2}\colorup{4.2}\\
\snameall (ours) & &  
\textbf{84.8}\stdv{1.3}\colorup{9.1} & \textbf{86.1}\stdv{3.4}\colorup{6.0} & \textbf{97.9}\stdv{0.4}\colorup{5.7} & 74.2\stdv{1.7}\colorup{3.5} & \textbf{87.3}\stdv{2.2}\colorup{7.5}\\
\midrule
CLIP (fine-tuned) & \multirow{3}{*}{20\%} &
78.8\stdv{1.4}\colorup{3.1} & 80.8\stdv{1.6}\up{0.7} & 91.9\stdv{0.4}\down{0.3} & 74.4\stdv{3.9}\colorup{3.7} & 82.7\stdv{0.5}\colorup{2.9}\\
SemiCLIP & &
85.5\stdv{1.8}\colorup{9.8} & \textbf{89.1}\stdv{0.4}\colorup{9.0} & 95.5\stdv{1.1}\colorup{3.3} & 67.4\stdv{2.6}\down{3.3} & 88.6\stdv{1.4}\colorup{8.8}\\
\snameall (ours) & &  
\,\>\textbf{87.0}\stdv{0.9}\colorup{11.3} & 84.9\stdv{1.0}\colorup{4.8} & \textbf{97.2}\stdv{0.9}\colorup{5.0} & \textbf{77.9}\stdv{1.9}\colorup{7.2} & \textbf{90.4}\stdv{2.0}\colorup{10.6}\\
\midrule
CLIP (fine-tuned) & \multirow{3}{*}{30\%} &
81.4\stdv{1.9}\colorup{5.7} & 83.8\stdv{2.8}\colorup{3.7} & 93.6\stdv{1.2}\colorup{1.4} & \textbf{72.9}\stdv{3.3}\colorup{2.2} & 85.2\stdv{1.6}\colorup{5.4}\\
SemiCLIP & &
\,\>85.9\stdv{0.5}\colorup{10.2} & \textbf{88.5}\stdv{0.9}\colorup{8.5} & 96.3\stdv{0.5}\colorup{4.1} & 68.9\stdv{3.4}\down{1.8} & 89.7\stdv{0.2}\colorup{9.9}\\
\snameall (ours) & &  
\,\>\textbf{87.4}\stdv{0.8}\colorup{11.7} & 84.5\stdv{3.3}\colorup{4.4} & \textbf{97.8}\stdv{0.7}\colorup{5.6} & {72.4}\stdv{4.2}\colorup{1.7} & \,\>\textbf{90.2}\stdv{2.2}\colorup{10.4}\\
\midrule
CLIP (fine-tuned) & 100\% &
87.5\stdv{1.8}\blankup & 67.9\stdv{2.3}\blankup & 94.2\stdv{1.5}\blankup & 75.7\stdv{2.9}\blankup & 89.3\stdv{1.2}\blankup\\
\bottomrule
\end{tabular}
\end{adjustbox}
\end{table*}

Overall, \sname remains effective across different labeled pair ratios. At the default 10\% setting, \sname achieves the best results on RSICD-CLS, UCM-CLS, WHU-RS19, and AID, while SemiCLIP performs better on RSSCN7. When the amount of labeled supervision increases to 20\% and 30\%, \sname continues to outperform SemiCLIP on RSICD-CLS, WHU-RS19, RSSCN7, and AID, although SemiCLIP remains stronger on UCM-CLS. These results suggest that the benefit of topology-aware alignment is not limited to the low-label regime, but continues to provide a useful complementary structural signal even when more labeled pairs are available.

In particular, the gains of \sname remain consistently strong on RSICD-CLS, WHU-RS19, and AID across all three ratios. This indicates that \sname improves the transferability of the learned representation in a robust manner, rather than relying on a specific labeled-data regime. On the other hand, the trends on UCM-CLS and RSSCN7 are less uniform. Importantly, this behavior is not unique to \sname: the supervised CLIP baseline itself does not improve monotonically as the labeled ratio increases on these datasets. This suggests that the effect of additional labeled supervision is dataset-dependent, while \sname still provides a robust complementary structural signal across different label regimes.

\subsection{Effect of Directional Invariance in Equation~(4)}
\label{app:direction}

In \eqref{eq:tomar_loss}, we use the absolute value of the directional consistency term. 
This design is not intended to resolve the arbitrary orientation of an unordered edge. 
Indeed, reversing the order of the two endpoints reverses both the source-space and mapped target-space difference vectors, leaving their cosine similarity unchanged. 
Instead, the absolute value makes the objective invariant to whether the mapped edge direction is parallel or anti-parallel to the source edge. 
Thus, \sname aligns the underlying one-dimensional relational subspace induced by a topologically selected edge, rather than enforcing a fixed global orientation across modalities.

\begin{table*}[ht!]
\centering\small
\caption{Ablation on the edge-direction term in Eq.~\ref{eq:tomar_loss}. \sname-direction removes the absolute value from the directional consistency term. Parentheses denote performance gains over supervised CLIP, and green values indicate gains larger than one. Bold denotes the best semi-supervised result in each setting.}
\label{tab:rs-direction}
\begin{adjustbox}{width=\linewidth}
\begin{tabular}{@{}lcccccc@{}}
\toprule
Method & RSICD-CLS & UCM-CLS & WHU-RS19 & RSSCN7 & AID\\
\midrule
CLIP (fine-tuned) &
75.7\stdv{2.2}\blankup & 80.1\stdv{6.8}\blankup & 92.2\stdv{1.5}\blankup & 70.7\stdv{5.5}\blankup & 79.8\stdv{3.8}\blankup\\
\midrule
\sname & \textbf{84.8}\stdv{1.3}\colorup{9.1} & \textbf{86.1}\stdv{3.4}\colorup{6.0} & \,\>\textbf{97.9}\stdv{0.4}\colorup{5.7} & \textbf{74.2}\stdv{1.7}\colorup{3.5} & \textbf{87.3}\stdv{2.2}\colorup{7.5}\\
\sname-direction & 77.4\stdv{1.0}\colorup{1.7} & 72.3\stdv{4.7}\down{7.8} & \,\>94.2\stdv{2.4}\colorup{2.0} & \,\>57.1\stdv{3.1}\down{13.6} & 81.1\stdv{1.2}\colorup{1.3}
\\ 
\bottomrule
\end{tabular}
\end{adjustbox}
\end{table*}

To verify this design choice, we compare \sname with a variant, denoted as \textit{\sname-direction}, that removes the absolute value and directly uses the raw cosine similarity of edge directions. 
This variant penalizes anti-parallel mapped directions even when they preserve the same one-dimensional relational structure. 
The results are reported in Table~\ref{tab:rs-direction}. 
Without the absolute value, performance drops substantially across most datasets. 
In particular, \sname-direction falls far behind the full \sname on UCM-CLS and RSSCN7, and also shows weaker gains on RSICD-CLS, WHU-RS19, and AID.

These results indicate that enforcing a fixed orientation is unnecessarily restrictive for topology-aware alignment. 
The absolute value in \eqref{eq:tomar_loss} is therefore not a minor implementation detail, but a design choice that allows \sname to focus on structural consistency of corresponding topological relations while remaining invariant to parallel and anti-parallel realizations of the same edge-induced relational subspace.

\section{Computational Time of \sname}
\label{app:run_time}

We also compare the run time of CLIP fine-tuned, SemiCLIP, and \sname in Table~\ref{tab:run_time}. Since CLIP fine-tuned is trained in a single stage for 25 epochs, whereas SemiCLIP and \sname use a two-stage schedule with 25 epochs in stage 1 and 15 epochs in stage 2, we report both per-epoch and total training time. 
For this run-time comparison only, we unify the batch size to 256 for all methods, although the default batch size of \sname in the main experiments is 128. Since both SemiCLIP and \sname rely on batchwise pairwise computations, changing the batch size affects the run time of both methods. The reported values should therefore be interpreted as a controlled comparison under matched conditions.

\begin{table}[!ht]
\centering\small
\caption{Run time comparison between CLIP fine-tuned, SemiCLIP, and \sname. Run time is measured under the default remote sensing setting on a single NVIDIA A100 GPU. For fair comparison, the batch size is unified to 256 only for this run-time experiment, although the default batch size of \sname in the main experiments is 128. CLIP fine-tuned is trained for 25 epochs in a single stage, whereas SemiCLIP and \sname are trained for 25 epochs in Stage 1 and 15 epochs in Stage 2.}
\label{tab:run_time}
\begin{tabular}{@{}lccc@{}}
\toprule
 & CLIP fine-tuned & SemiCLIP & \sname \\
\midrule
Stage 1 run time per epoch (s) & \multirow{2}{*}{7.4} & 19.2 & 19.3 \\
Stage 2 run time per epoch (s) &  & 21.3 & 21.3 \\
Increase over SemiCLIP (stage 1) & -- & -- & +0.5\% \\
Increase over SemiCLIP (stage 2) & -- & -- & +0.0\% \\
Total training time (s) & 185.0 & 799.5 & 802.0 \\
Increase over SemiCLIP (total) & -- & -- & +0.3\% \\
\bottomrule
\end{tabular}
\end{table}

Under this controlled setting, \sname introduces almost no additional computational overhead over SemiCLIP: the stage-1 run time increases only from 19.2s to 19.3s per epoch, while the stage-2 run time remains unchanged at 21.3s. The total training time also increases only marginally, from 799.5s to 802.0s. This small overhead results from the lightweight PH computation, which extracts $H_0$-death and $H_1$-birth relations from the 1-skeleton without constructing 2-simplices. Overall, these results suggest that the proposed topology-aware alignment is computationally lightweight in practice.

\section{Compute Resources}
\label{app:compute_resoure}

All experiments were conducted on a local server equipped with two AMD EPYC 7513 CPUs
(2.5\,GHz, 32 cores each; 64 cores / 128 threads in total), 512\,GB RAM, and two NVIDIA A100
GPUs with 80\,GB memory each. The server also provides 512\,GB of working storage and 3.5\,TB
of data storage. Unless otherwise noted, each individual experimental run used a single NVIDIA
A100 80GB GPU. The run-time comparison reported in Appendix~\ref{app:run_time} was also measured on a single
A100 GPU.


\end{document}